\documentclass[11pt]{article}
\usepackage{enumitem}
\usepackage{amsmath}
\usepackage{subcaption}
\usepackage{authblk}
\usepackage{caption}
\usepackage{graphicx}

\usepackage[preprint]{acl}
\usepackage{booktabs,threeparttable,tabularx,makecell}

\usepackage{times}
\usepackage{latexsym}
\usepackage{xurl}
\usepackage{siunitx}

\usepackage[T1]{fontenc}

\usepackage[utf8]{inputenc}

\usepackage{microtype}

\usepackage{inconsolata}

\usepackage{booktabs} 
\usepackage{multirow}
\usepackage{amssymb}  
\usepackage{bm}       
\usepackage{booktabs}
\usepackage{graphicx}
\usepackage[table]{xcolor}
\usepackage{makecell}
\usepackage{xcolor}

\usepackage{graphicx}
\definecolor{zccolor}{RGB}{0,0,0}

\title{Omni-I2C: A Holistic Benchmark for High-Fidelity Image-to-Code Generation}

\author{
\small Jiawei Zhou$^{1,*}$ \quad
Chi Zhang$^{1,*}$ \quad
Xiang Feng$^{1}$ \quad
Qiming Zhang$^{2}$ \quad
Haibo Qiu$^{2}$ \\
\small Lihuo He$^{3,\dagger}$ \quad
Dengpan Ye$^{4,\dagger}$ \quad
Xinbo Gao$^{3}$ \quad
Jing Zhang$^{1,\dagger}$ \\
\makebox[0.33\textwidth][c]{\small $^{1}$School of Computer Science, Wuhan University, Wuhan, China} \\
\makebox[0.33\textwidth][c]{\small $^{2}$Independent Researcher, China} 
\makebox[0.33\textwidth][c]{\small $^{3}$Xidian University, Xi'an, China} 
\makebox[0.33\textwidth][c]{\small $^{4}$Guangzhou University, Guangzhou, China} \\
\small $^{\dagger}$Corresponding author: \texttt{jingzhang.cv@gmail.com}, \texttt{lhhe@mail.xidian.edu.cn}, \texttt{yedp@gzhu.edu.cn}
}

\begin{document}
\maketitle

\begingroup
\renewcommand{\thefootnote}{\fnsymbol{footnote}}
\footnotetext[1]{Equal contribution.}
\endgroup

\begin{abstract}
We present Omni-I2C, a comprehensive benchmark designed to evaluate the capability of Large Multimodal Models (LMMs) in converting complex, structured digital graphics into executable code. We argue that this task represents a non-trivial challenge for the current generation of LMMs: it demands an unprecedented synergy between high-fidelity visual perception—to parse intricate spatial hierarchies and symbolic details—and precise generative expression—to synthesize syntactically sound and logically consistent code. Unlike traditional descriptive tasks, Omni-I2C requires a holistic understanding where any minor perceptual hallucination or coding error leads to a complete failure in visual reconstruction. Omni-I2C features 1080 meticulously curated samples, defined by its breadth across subjects, image modalities, and programming languages. By incorporating authentic user-sourced cases, the benchmark spans a vast spectrum of digital content—from scientific visualizations to complex symbolic notations—each paired with executable reference code. To complement this diversity, our evaluation framework provides necessary depth; by decoupling performance into perceptual fidelity and symbolic precision, it transcends surface-level accuracy to expose the granular structural failures and reasoning bottlenecks of current LMMs. Our evaluation reveals a substantial performance gap among leading LMMs; even state-of-the-art models struggle to preserve structural integrity in complex scenarios, underscoring that multimodal code generation remains a formidable challenge. Data and code are available at https://github.com/MiliLab/Omni-I2C.
\end{abstract}

\section{Introduction}

\begin{figure}[t!]
  \centering
  \includegraphics[width=0.45\textwidth]{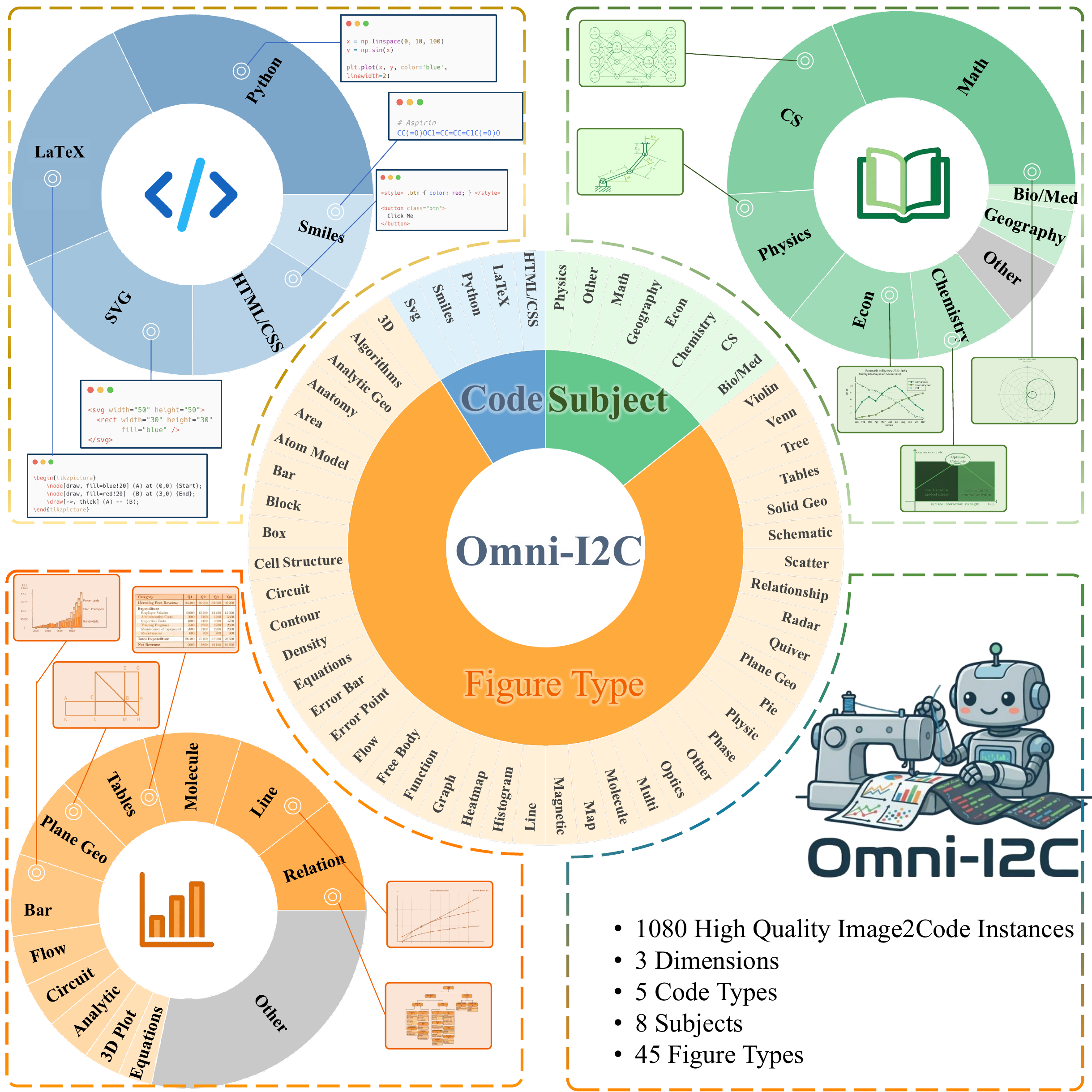}
  \caption{Taxonomy overview of the Omni-I2C benchmark. The dataset features a diverse distribution across 5 code types, 8 major subjects, and 45 distinct figure types.}
\label{fig:overview}
\end{figure}

\begin{table*}[t!]
\centering
\caption{Compared with prior benchmarks, Omni-I2C is more diverse in terms of subjects, figure types and output modalities.}
\label{tab:benchmarks comparison}

\scriptsize

\begin{tabular}{l c c c c c}
\toprule
\textbf{Benchmark} & \textbf{Domain} & \textbf{Samples} & \textbf{Subjects} & \textbf{Figure Types} & \textbf{Output Modalities} \\
\midrule

ChartMimic~\citep{yang2024chartmimic} & Chart & 4,800 & 8 & 22 & Python \\ 
Chart2Code~\citep{tang2025charts} & Chart & 2,023 & 4 & 22 & Python \\ 
Plot2Code~\citep{wu2025plot2code} & Plot & 368 & - & 6 & Python, R \\ 
Design2Code~\citep{si2025design2code} & Web & 484 & - & - & HTML \\ 
UniSVG~\citep{li2025unisvg} & Icon & 2,850 & - & - & SVG \\ 
Image2Struct~\citep{roberts2024image2struct} & Doc/Music & 2,400 & 3 & 6 & LaTeX, HTML \\  
\midrule

\rowcolor{gray!15} 
\textbf{Omni-I2C} & \textbf{Generalist} & \textbf{1,080} & \textbf{8} & \textbf{45} & Python, SVG, HTML, LaTeX, SMILES \\ 
\bottomrule
\end{tabular}%

\vspace{-3mm}
\end{table*}

In professional workflows, programming is rarely a requirement isolated from visual cues. Whether it is an engineer re-implementing complex data interfaces from legacy archives, a developer transforming high-fidelity visual mockups into structured layouts, or a researcher converting technical diagrams and notations into executable logic, the demand for Image-to-Code (I2C) conversion is both ubiquitous and valuable~\citep{yang2024chartmimic,si2025design2code,wu2025plot2code,li2024mmcode}. This ``pixel-to-program'' synthesis represents an advanced application of human-AI collaboration, where LMMs act as intelligent assistants that bridge the gap between visual information and executable functional logic.

Beyond its practical utility, I2C generation offers a unique paradigm for evaluating high-fidelity perception compared to conventional tasks such as Visual Question Answering (VQA)~\citep{yue2024mmmu,yue2025mmmu,lu2023mathvista,antol2015vqa}. While VQA evaluates a model's ability to extract specific information through targeted queries, I2C generation demands a holistic, structural reconstruction of the entire visual input. It is intrinsically less tolerant of perceptual discrepancies; even a minute error—such as misidentifying a LaTeX subscript or a specific coordinate—will cause the rendered output to deviate, fundamentally altering the semantic meaning of the result. This execution-based verifiability enables a more comprehensive, fine-grained, and objective evaluation of the perceptual capabilities of a model.

Recognizing this potential, recent benchmarking efforts begin to explore I2C generation within specialized domains. For instance, ChartMimic~\citep{yang2024chartmimic} and Plot2Code~\citep{wu2025plot2code} focus on synthesizing plotting scripts from scientific charts, while Design2Code~\citep{si2025design2code} evaluates the transformation of web screenshots into front-end implementations. Although these works have established a vital foundation, current evaluations remain largely restricted in image variety, programming languages, and task scenarios. To comprehensively assess the limits of perception, coding and reasoning, a more diverse and general-purpose testbed is essential to evaluate whether LMMs can generalize across varied technical and professional domains.

\begin{figure*}
\vspace{-4mm}
    \centering
    \includegraphics[width=1.0\linewidth]{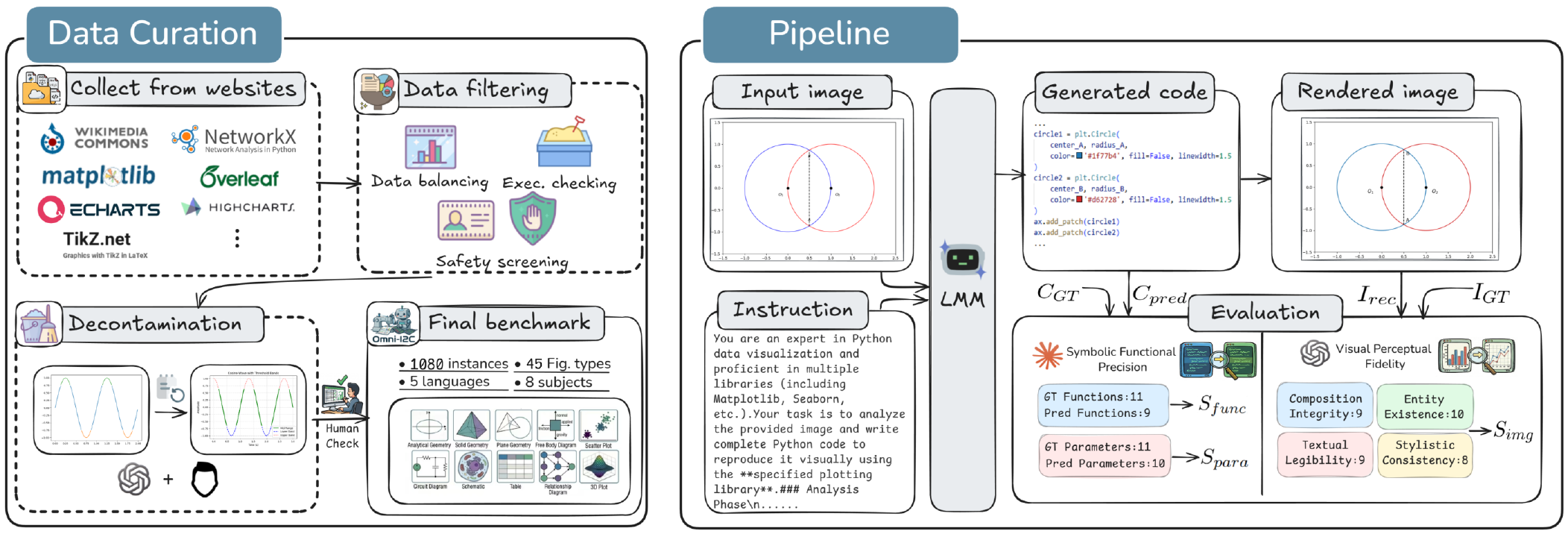}
    \caption{The Data Curation and pipeline of Omni-I2C.    \textbf{Data Curation} (Left): From raw web collection to a refined benchmark with diverse themes and languages, ensured by strict filtering and human verification. \textbf{Pipeline} (Right): The proposed workflow for image-to-code generation and automatic evaluation.}
    \label{fig:overall pipeline}
\vspace{-4mm}
\end{figure*}

To bridge this gap, we introduce Omni-I2C, a comprehensive benchmark for stress-testing the limits of Image-to-Code (I2C) generation. The dataset contains 1.1k meticulously curated samples drawn from real-world user applications, covering a broad spectrum of technical graphics such as scientific visualizations, molecular structures, and complex symbolic notations, and supporting diverse programming languages. Furthermore, we propose a multi-level evaluation framework that decomposes model performance into fine-grained attributes, enabling a more nuanced understanding of model strengths and weaknesses across different dimensions. Unlike previous metrics restricted to specific domains or coding formats, our evaluation logic offers comprehensive coverage and demonstrates superior alignment with human judgment through extensive empirical validation.

Through an extensive evaluation of 13 proprietary and open-weight LMMs, we reveal a profound performance gap in high-fidelity image-to-code generation. Even leading frontier models, such as Gemini 3 Pro and GPT-5.1, frequently falter in the challenging scenarios presented by our benchmark. These results highlight substantial room for improvement and position Omni-I2C as a challenging benchmark for advancing LMMs. Our contributions are summarized as follows:
\begin{itemize}
    \item We present Omni-I2C, a meticulously curated dataset of 1.1k items, including 5 programming languages,  8 major subjects, and 45 distinct figure types. It serves as a rigorous testbed for evaluating the perception and coding capabilities of LMMs.
    \item We propose an evaluation framework that assesses code-level integrity and image-level perceptual consistency, enabling more diagnostic and attributable analyses of model behavior than traditional heuristic metrics.
    \item Our comprehensive analysis of SOTA LMMs exposes a significant performance gap in high-fidelity reconstruction, identifying critical failure modes and charting a path toward more precise, trusted multimodal agents.
\end{itemize}

\section{Related Work}
\subsection{Multimodal Code Benchmarks}
The rapid advancement of Large Language Models (LLMs) has reshaped automated code generation, from general-purpose foundation models~\citep{OpenAI2025gpt5,GOOGLE2025gemini3pro,ANTHROPIC2025claudesonnet4_5} to specialized architectures~\citep{guo2024deepseek,hui2024qwen2} that excel at complex programming tasks~\citep{jimenez2023swe,zhou2023webarena,yao2024tau}.  Model evaluation, however, has largely remained single-modal, relying on text-only benchmarks such as HumanEval~\citep{chen2021evaluating}, MBPP~\citep{austin2021program}, and DS-1000~\citep{lai2023ds}. As models become multimodal, evaluation must expand to real-world settings where visual perception is integral.

Recent multimodal benchmarks—MMCode~\citep{li2024mmcode}, Design2Code~\citep{si2025design2code}, Plot2Code~\citep{wu2025plot2code}, and ChartMimic~\citep{yang2024chartmimic}—take steps in this direction by testing HTML generation and chart-to-code synthesis. Yet they are typically domain-bound, with limited diversity in subjects, visual modalities, and programming languages. To address these limitations, we introduce Omni-I2C, a comprehensive benchmark for high-fidelity visual understanding and image-to-code generation. See Tab.~\ref{tab:benchmarks comparison} for detailed comparisons with prior benchmarks.

\subsection{Visual Perception in LMMs}
The evolution of Large Multimodal Models has transitioned from coarse recognition to complex, fine-grained visual reasoning, bolstered by high-quality data and test-time scaling~\citep{alayrac2022flamingo, liu2023visual, chen2024sharegpt4v,wang2024measuring,lu2023mathvista,yue2024mmmu,qiao2025we,zhang2025thyme,zhang2025deepsketcher,zheng2025deepeyes,meng2025mm}. However, existing VQA benchmarks—often relying on multiple-choice or short-form answers—suffer from sparse or ambiguous supervision~\citep{zhang2025perceptual}. Such formats frequently allow models to exploit language priors or ``shortcuts'' without true visual grounding, making failures difficult to diagnose. Image-to-code generation mitigates this by introducing dense, execution-based supervision through “pixel-to-program” mapping—where perceptual mistakes translate into functional reconstruction errors—allowing Omni-I2C to assess perceptual fidelity more granularly than traditional VQA. 

\section{The Omni-I2C Benchmark}
\subsection{Task Definition}
The Omni-I2C benchmark probes cross-modal intelligence by requiring models to synthesize executable code that reconstructs complex digital graphics. Formally, given an input image $I$ and an instruction $T$ specifying the target programming language and reconstruction goal, an LMM $f_\theta$ generates code $C_{pred}$:$$C_{pred} = f_{\theta}(I, T).$$Executing $C_{pred}$ produces a reconstructed image $I_{\mathrm{rec}}$. Omni-I2C then evaluates the fidelity of $I_{\mathrm{rec}}$ to $I$ along multiple dimensions, including semantic consistency and structural alignment.

\subsection{Benchmark Coverage Analysis}
\label{sec:benchmark_taxonomy}
The Omni-I2C benchmark comprises 1,080 high-quality evaluation items curated to span a broad range of programming languages, academic subjects, and visual representations. This coverage provides substantially greater breadth and depth than existing image-to-code benchmarks.

\noindent\textbf{Code Types.}
Omni-I2C includes five languages: Python, LaTeX, HTML, SVG, and SMILES. They are selected to cover complementary real-world workflows: Python and LaTeX represent widely used academic pipelines for visualization and typesetting, HTML and SVG target broad web and engineering settings for structured layouts and vector graphics, and SMILES brings domain-specific structure from cheminformatics.

\noindent\textbf{Subjects.}
To reflect practical use cases, Omni-I2C covers over eight subjects, spanning foundational sciences (e.g., Mathematics, Physics, Chemistry) and applied domains (e.g., Computer Science, Biomedicine, Economics). This diversity pushes models beyond pattern matching toward domain-aware interpretation of field-specific conventions and visual languages.

\noindent\textbf{Image Types.}
Our benchmark features 45 distinct plot types, ranging from standard statistical charts to complex biological diagrams and geographical maps. This wide coverage ensures a comprehensive evaluation of a model's visual perception abilities.

By jointly varying code types, domains, and visual formats, Omni-I2C evaluates integrated perception-and-generation ability rather than isolated skills, reducing the impact of single-domain expertise. Fig.~\ref{fig:overview} summarizes the data distribution; additional taxonomy details and qualitative examples are provided in the App.~\ref{sec:appendix_data}.

\subsection{Data Curation Process}
\label{sec:data_curation}
Omni-I2C is built through a three-stage pipeline consisting of data collection, filtering, and decontamination, as demonstrated in Fig.~\ref{fig:overall pipeline}.

\noindent\textbf{Data Collection.} We adopt a taxonomy-driven strategy, synthesizing a schema from benchmarks~\citep{yang2024chartmimic,wang2023mathcoder} and authoritative references (App.~\ref{sec:appendix_data}) to guide retrieval. High-quality acquisition is ensured through three criteria: (i) \textit{Fidelity}, requiring a strict structural alignment between the code and its rendered output; (ii) \textit{Complexity}, prioritizing samples with non-trivial logical depth or intricate spatial arrangements; and (iii) \textit{Authenticity}, focusing on idiomatic code that reflects real-world programming practices. We harvest image-code pairs from official galleries (e.g., \textit{Matplotlib}, \textit{TikZ.net}) and refine subsets from existing datasets~\citep{yang2024chartmimic,yang2025scaling} to address under-represented categories in the Omni-I2C taxonomy.

\noindent\textbf{Data filtering.} To ensure the integrity of the evaluation set, we implement a multi-stage filtering pipeline. First, technical executability is guaranteed via isolated sandbox execution, where any code with compilation or runtime errors is discarded. Second, a manual audit sanitizes the collection by removing private data (PII) and social biases. Third, we employ an LLM-driven semantic refinement process to develop a granular taxonomy, followed by an expert review to rectify any misalignments. Finally, the dataset is balanced by pruning over-represented classes and excluding instances with extreme sequence lengths. This rigorous curation ensures a high-quality, robust collection of evaluation samples.

\noindent\textbf{Data decontamination.} As the raw data originates from public web sources, we develop a refactoring strategy to mitigate data leakage and prevent shortcut learning through memorization. We employ an LLM to systematically refactor the code’s stylistic and aesthetic attributes, such as fonts and layouts. For textual elements within images, we perform semantic sanitization by erasing original text and injecting entirely new, synthetic strings. Furthermore, we manually modify key attributes, including colors, label coordinates, and textual content. To further decouple visual perception from linguistic priors, we occasionally introduce counter-factual ``pseudo-labels'' by substituting labels or changing the content to nonsense. This design choice ensures that models cannot rely on common-sense guesses or memorized patterns, forcing them to anchor their outputs strictly on the provided visual evidence. Representative visual comparisons of the original and decontaminated samples are provided in App.~\ref{sec:appendix_data}.

\subsection{Evaluation Metrics}
\label{sec:metrics}
Current I2C evaluation paradigms often fail to reconcile perceptual intuition with objective rigor~\citep{li2024mmcode,si2025design2code,wu2025plot2code,yang2024chartmimic}. Methods relying solely on programmatic metrics tend to be library-dependent, while those based purely on LMM-judges may introduce subjective variance. To provide a more comprehensive and robust assessment, Omni-I2C introduces a synergetic perceptual-symbolic evaluation framework that harmonizes qualitative visual perception with quantitative algorithmic rigor.

Under this paradigm, each evaluation instance is characterized by a cross-comparison between the predicted pair $(I_{rec}, C_{pred})$ and the ground-truth reference $(I_{GT}, C_{GT})$. Specifically, this framework bifurcates the assessment into two complementary tracks: Visual Perceptual Fidelity and Symbolic Functional Precision. 

\noindent\textbf{Visual Perceptual Fidelity.} This track utilizes a frontier LMM to provide a human-aligned score $S_{img}$ of the rendered output $I_{rec}$. Rather than requiring the model to perform unreliable fine-grained parsing, this track prioritizes global structural coherence and stylistic essence. Specifically, $S_{img}$ is derived from a multi-dimensional assessment across four axes: \textit{Style}, \textit{Layout}, \textit{Elements}, and \textit{Text}. While the textual axis ensures content accuracy, the former three dimensions are designed to capture holistic, global attributes—such as color harmony and spatial distribution. This serves as a qualitative proxy for  human perception, ensuring that the synthesized result is visually plausible and stylistically consistent with the original input.

\noindent\textbf{Symbolic Functional Precision.} To complement visual intuition with objective grounding, this track evaluates the generated code against deterministic, pre-defined checklists. The reference functional logic blocks ($\mathcal{F}_{GT}$) and fine-grained parameters ($\mathcal{P}_{GT}$) for each instance are pre-extracted via a powerful LLM and manually verified. To avoid fuzzy code-to-code comparisons, the LLM evaluator acts solely as a structured parser, scrutinizing the predicted code $C_{pred}$ to verify the presence of these fixed elements ($\mathcal{F}_{match}$, $\mathcal{P}_{match}$). We quantify the model's performance via Functional Coverage ($S_{func}$) and Parametric Accuracy ($S_{para}$), as:

$$S_{func} = \frac{|\mathcal{F}_{match}|}{|\mathcal{F}_{GT}|}, \quad S_{para} = \frac{|\mathcal{P}_{match}|}{|\mathcal{P}_{GT}|}.$$

Further details regarding the implementation of these metrics are provided in the App.~\ref{evaluation setup}.

While this pipeline is applied to HTML, LaTeX, Python, and SVG tasks, we adopt a domain-specific approach for SMILES data. We assess reconstruction fidelity by calculating the Tanimoto Similarity~\citep{bajusz2015tanimoto} between the Morgan Fingerprints~\citep{rogers2010extended} of the ground-truth and generated code, which serves as a proxy for structural similarity. We provide examples in App.~\ref{appendix:smiles_metric} for illustration.

\begin{table*}[t]
\vspace{-2mm}
\centering
\tiny
\setlength{\tabcolsep}{2pt} 
\renewcommand{\arraystretch}{1.1} 
\caption{Main results on Omni-I2C. Exec. denotes Execution Rate. For \textbf{SMILES}, we report Exec. and Tanimoto Similarity (T.S.) score. Best results in each category are \textbf{bold}, second best are \underline{underlined} for proprietary models and open-weight models.}
\label{tab:reorganized_breakdown_final_updated}

\resizebox{\linewidth}{!}{
\begin{tabular}{l | ccc | c | ccc | c | ccc | c | ccc | c | ccc | c | c | c}
\toprule
\multirow{2}{*}{\textbf{Model}} & \multicolumn{4}{c|}{\textbf{Overall(1080)}} & \multicolumn{4}{c|}{\textbf{HTML(166)}} & \multicolumn{4}{c|}{\textbf{LaTeX(265)}} & \multicolumn{4}{c|}{\textbf{Python(357)}} & \multicolumn{4}{c|}{\textbf{SVG(192)}} & \multicolumn{2}{c}{\textbf{SMILES(100)}} \\
\cmidrule(lr){2-5} \cmidrule(lr){6-9} \cmidrule(lr){10-13} \cmidrule(lr){14-17} \cmidrule(lr){18-21} \cmidrule(lr){22-23}
& $S_{\text{func}}$ & $S_{\text{para}}$ & $S_{\text{img}}$ & Exec & $S_{\text{func}}$ & $S_{\text{para}}$ & $S_{\text{img}}$ & Exec & $S_{\text{func}}$ & $S_{\text{para}}$ & $S_{\text{img}}$ & Exec & $S_{\text{func}}$ & $S_{\text{para}}$ & $S_{\text{img}}$ & Exec & $S_{\text{func}}$ & $S_{\text{para}}$ & $S_{\text{img}}$ & Exec & T.S. & Exec \\
\midrule
\rowcolor{gray!10}
\multicolumn{23}{c}{\textit{Proprietary Models}} \\
Claude Sonnet 4.5 & 60.4 & 40.8 & \textbf{76.7} & \textbf{97.2} & 67.2 & \underline{55.3} & \underline{84.2} & \textbf{99.4} & \textbf{47.7} & \textbf{29.4} & \textbf{66.8} & \textbf{92.8} & 62.5 & 41.0 & 81.0 & \textbf{98.9} & 68.1 & \underline{44.0} & 75.8 & \textbf{100.0} & \underline{59.4} & \textbf{94.0} \\
GPT 5.1 & \underline{60.7} & \underline{40.9} & 74.1 & 92.7 & \underline{67.4} & 55.2 & 84.2 & \underline{99.4} & 46.9 & \underline{28.0} & 55.2 & \underline{84.5} & \textbf{63.3} & \textbf{42.9} & \textbf{81.9} & 94.7 & \underline{69.1} & 42.5 & \underline{76.9} & \underline{100.0} & 49.0 & 82.0 \\
Gemini 3 Pro & \textbf{62.2} & \textbf{41.4} & \underline{76.1} & \underline{92.8} & \textbf{68.9} & \textbf{56.0} & \textbf{86.2} & 98.2 & \underline{47.2} & 27.2 & \underline{56.5} & 83.4 & \underline{62.6} & \underline{42.2} & \underline{81.9} & \underline{95.0} & \textbf{76.3} & \textbf{47.0} & \textbf{83.7} & 100.0 & 48.0 & \underline{87.0} \\
Gemini 2.5 Pro & 58.4 & 39.6 & 71.2 & 89.3 & 61.5 & 54.9 & 80.8 & 98.8 & 45.0 & 26.9 & 52.5 & 72.5 & 62.0 & 40.9 & 80.7 & 93.3 & 67.6 & 41.6 & 71.1 & 99.0 & \textbf{75.6} & 85.0 \\
\textit{Average} & 60.4 & 40.7 & 74.5 & 93.0 & 66.2 & 55.4 & 83.8 & 99.0 & 46.7 & 27.9 & 57.8 & 83.3 & 62.6 & 41.8 & 81.4 & 95.5 & 70.3 & 43.8 & 76.9 & 99.8 & 58.0 & 87.0 \\
\midrule
\rowcolor{gray!10}
\multicolumn{23}{c}{\textit{Open-Weight Models}} \\
InternVL3\_5\_241B\_A28B-Instruct & 50.8 & 36.2 & \underline{66.5} & \textbf{95.3} & 55.6 & 51.9 & \underline{76.6} & 98.8 & 36.2 & 25.5 & \underline{54.5} & \underline{90.6} & \underline{55.9} & 36.3 & \textbf{73.8} & \textbf{94.4} & 57.3 & 37.5 & 61.1 & \textbf{100.0} & \textbf{68.9} & \textbf{96.0} \\
InternVL3\_5-38B-Instruct & 37.5 & 30.2 & 52.2 & 90.2 & 40.6 & 46.6 & 61.8 & 91.0 & 22.3 & 21.0 & 38.6 & 87.2 & 44.0 & 27.7 & 59.2 & 87.1 & 43.6 & 33.5 & 49.7 & 99.0 & 58.8 &  \underline{94.0} \\
InternVL3\_5-8B-Instruct & 28.6 & 24.1 & 42.3 & 88.4 & 36.7 & 40.2 & 56.5 & 97.6 & 14.1 & 16.8 & 34.6 & 89.1 & 34.1 & 20.6 & 45.8 & 77.9 & 31.2 & 26.8 & 34.2 & 99.0 & 48.1 & 87.0 \\
Qwen3-VL-235B-A22B-Instruct & \textbf{57.0} & \textbf{39.9} & \textbf{70.0} & \underline{93.8} & \textbf{65.4} & \textbf{54.3} & \textbf{79.3} & \textbf{100.0} & \textbf{46.2} & \textbf{30.1} & \textbf{60.6} & 90.2 & \textbf{57.1} & \textbf{39.4} & \underline{72.4} & \underline{90.5} & \textbf{64.6} & \underline{42.0} & \textbf{70.4} & 99.5 & \underline{66.9} & \underline{94.0} \\
Qwen2.5-VL-72B-Instruct & 46.0 & 34.2 & 59.8 & 92.7 & 52.0 & 51.4 & 72.3 & \underline{100.0} & 34.4 & \underline{25.6} & 53.3 & \textbf{92.8} & 49.4 & 32.2 & 64.0 & 88.5 & 50.4 & 34.8 & 50.2 & 99.5 & 29.8 & 82.0 \\
Qwen3-VL-32B-Instruct & \underline{53.4} & \underline{37.9} & 62.9 & 88.7 & \underline{62.0} & \underline{54.0} & 74.0 & 100.0 & \underline{40.6} & 24.9 & 51.0 & 81.1 & 53.1 & \underline{37.1} & 65.0 & 84.6 & \underline{64.2} & \textbf{43.4} & \underline{65.7} & \underline{100.0} & 51.6 & 83.0 \\
Qwen3-VL-8B-Instruct & 46.6 & 33.6 & 53.8 & 85.5 & 59.1 & 51.6 & 70.9 & 97.6 & 33.8 & 22.7 & 38.4 & 75.1 & 46.7 & 31.1 & 58.6 & 82.1 & 53.3 & 37.6 & 51.5 & 99.0 & 53.4 & 85.0 \\
Qwen2.5-VL-7B-Instruct & 30.4 & 25.9 & 41.5 & 78.7 & 39.0 & 45.2 & 59.9 & 94.0 & 17.0 & 14.8 & 29.6 & 65.7 & 34.7 & 22.5 & 46.0 & 78.7 & 33.7 & 30.6 & 33.8 & 99.5 & 15.3 & 51.0 \\
Gemma3-27B-Instruct & 33.5 & 26.5 & 45.1 & 78.3 & 39.8 & 45.0 & 62.1 & 99.4 & 12.0 & 13.7 & 21.8 & 52.8 & 40.9 & 24.5 & 52.3 & 82.6 & 44.1 & 31.7 & 49.1 & 99.5 & 23.1 & 55.0 \\
\textit{Average} & 42.6 & 32.1 & 54.9 & 88.0 & 50.0 & 48.9 & 68.2 & 97.6 & 28.5 & 21.7 & 42.5 & 80.5 & 46.2 & 30.2 & 59.7 & 85.2 & 49.2 & 35.3 & 51.7 & 99.4 & 46.2 & 80.8 \\
\bottomrule
\end{tabular}%
}
\vspace{-3mm}
\end{table*}

\begin{figure*}
    \centering
    \includegraphics[width=1\linewidth]{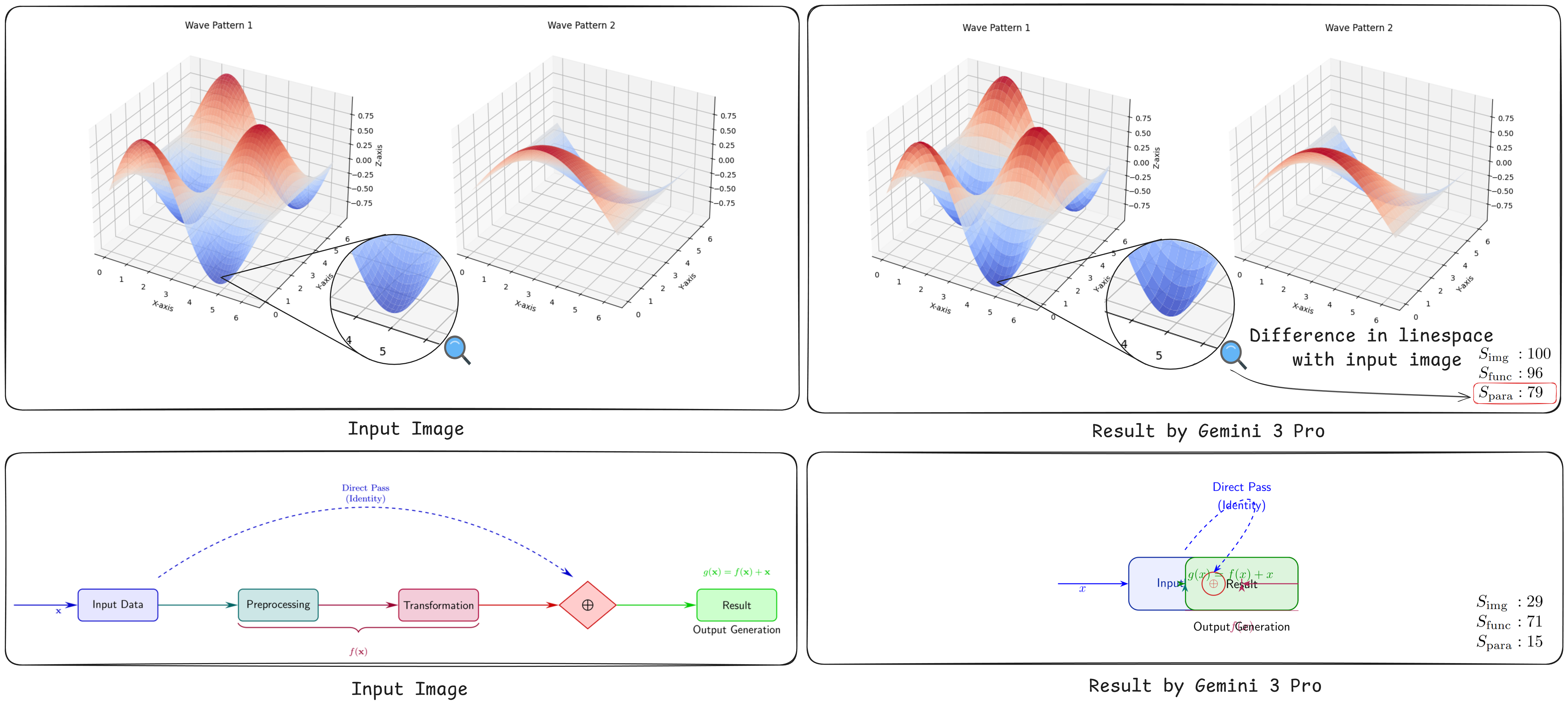 }
    \caption{Representative examples from Gemini 3 Pro.}
    \label{fig:case_study_main}
    \vspace{-4mm}
\end{figure*}

\section{Experiment}

\subsection{Baseline Setup}
We benchmark 13 LMMs on Omni-I2C, including proprietary frontiers—Claude Sonnet 4.5~\citep{ANTHROPIC2025claudesonnet4_5}, GPT 5.1~\citep{OPENAI2025gpt51}, Gemini 3 Pro~\citep{GOOGLE2025gemini3pro}, and Gemini 2.5 Pro~\citep{comanici2025gemini}—and a diverse open-weight spectrum (7B--241B) including InternVL 3.5~\citep{wang2025internvl3}, Qwen3-VL~\citep{Qwen3-VL}, Qwen2.5-VL~\citep{Qwen2.5-VL}, and Gemma3-27B~\citep{team2025gemma}. This setup spans major closed and open ecosystems to ensure a comprehensive comparison across varying model scales and regimes. We employ refined, language-specific prompts that enforce explicit structural constraints (see App.~\ref{sec:appendix_setups} for details). For the evaluation, we utilize GPT 4.1 to score perceptual fidelity and Claude Sonnet 4.5 to audit code-level functionality.

\subsection{Main Results}
Tab.~\ref{tab:reorganized_breakdown_final_updated} presents a comprehensive overview of the benchmarking results across 13 LMMs. Combined with the qualitative analysis in Fig.~\ref{fig:case_study_main} and App. \ref{case_study}, we derive several critical observations regarding the current landscape of image-to-code execution:

\noindent\textbf{The challenging nature of the Omni-I2C.} 
Omni-I2C reveals that translating complex visuals into executable code remains a significant barrier. Tab.~\ref{tab:reorganized_breakdown_final_updated} shows that high execution rates do not guarantee high-fidelity reconstruction. Even for frontier models, a performance ceiling exists in semantic and parametric precision, confirming that Omni-I2C provides a rigorous and unsaturated testbed for evaluating the next generation of LMMs.

\noindent\textbf{The Universal Gap Between Logic and Precision.} 
A consistent trend is that Parametric Accuracy ($S_{\text{para}}$) significantly lags behind Functional Coverage ($S_{\text{func}}$). This is because $S_{\text{para}}$ identifies fine-grained symbolic discrepancies—such as the subtle \texttt{linspace} deviations in Fig.~\ref{fig:case_study_main}—that are difficult to discern via human vision or perceptual metrics ($S_{\text{img}}$). Consequently, $S_{\text{para}}$ provides a high-precision audit that visual assessment alone lacks the resolution to capture.

\noindent\textbf{Divergent Expertise among Proprietary Frontiers.} While proprietary models maintain a systematic lead, their strengths are domain-specific. Claude Sonnet 4.5 achieves the highest execution rate (97.2\%), marking it the most reliable for syntactic validity and LaTeX tasks. Conversely, Gemini 3 Pro excels in code-level metrics ($S_{\text{func}}$, $S_{\text{para}}$) and visually-intensive domains (HTML/SVG). Meanwhile, GPT 5.1 maintains the lead in Python generation.

\noindent\textbf{Scaling Laws and Generational Leaps in Open-Weight Models.} Performance tracks with both scale and architecture. The InternVL 3.5 series exhibits clear scaling effects up to 241B. Meanwhile, architectural advancements can outweigh raw parameter counts; for instance, the 32B Qwen3-VL consistently surpasses the larger 72B Qwen2.5-VL, illustrating a distinct generational leap. Most notably, Qwen3-235B achieves state-of-the-art results in structured LaTeX tasks, rivaling proprietary frontiers and marking a significant milestone for open-source precision in specialized domains.

\noindent\textbf{LaTeX Complexity as a Rigorous Geometric Discriminator.} LaTeX stands out as the most formidable challenge in Omni-I2C, characterized by a substantial performance drop across all models compared to other categories. This task poses a significant barrier for small-scale models (7B--8B), which exhibit a marked deficiency in maintaining structural and syntactic integrity, often yielding the lowest scores across all metrics. The stark contrast between LaTeX and other visual-to-code tasks underscores that the precise translation of visual structures into dense, rule-heavy markup remains the primary differentiator for frontier LMMs.

\section{Discussion}

\subsection{Study on Evaluation Metrics}
\noindent\textbf{Image Level Evaluation.} Within the Visual Perceptual Fidelity track, we prompt an LMM judge to quantify the visual fidelity ($S_{\text{img}}$) of the reconstructed outputs. 
To validate the reliability of this metric, we conduct a human-preference alignment study involving three candidate models—Gemini 2.5 Pro, InternVL 3.5-8B-Instruct, and Qwen3-VL-235B-A22B-Instruct—evaluated by three distinct judges: GPT 4.1, Qwen-VL-Max, and Qwen3-VL-Plus. We benchmark our proposed $S_{\text{img}}$ metric against four existing image-level evaluation paradigms by measuring their respective alignment with human consensus. The prompt templates of existing image-level evaluation paradigms are shown in App.~\ref{Ablation study of image_level}. For the annotation process, 10 human participants are each assigned 50 randomly sampled instances. For each instance, annotators are presented with the ground-truth image ($I_{GT}$) alongside three reconstructions ($I_{rec}$) generated by the contestant models, which they subsequently ranked according to a standardized preference policy (further details are provided in the App.~\ref{human study}).

The results, summarized in Tab.~\ref{tab:image_level_final}, indicate that both the judge model and the prompting strategy significantly influence the reliability of perceptual assessment. Notably, the Omni-I2C prompt paired with GPT 4.1 achieves the strongest correlation with human preferences, yielding a Kendall’s $\tau$ of 0.8304 and the lowest RMSE of 0.4284. Even when deployed with alternative judges such as Qwen-VL-Max or Qwen3-VL-Plus, our metric consistently outperforms methods from prior work.

\begin{table}[t]
\vspace{-2mm}
    \centering
    \small
    \setlength{\tabcolsep}{3pt}
    \caption{
        \textbf{Image-level alignment analysis. }
        Gray shading highlights the best-performing setting (Omni-I2C with GPT 4.1). A detailed description of the evaluation metrics and the rationale behind their selection are provided in the App.~\ref{sec:appendix_judge}.
    }
    \label{tab:image_level_final}
    
    \resizebox{\columnwidth}{!}{
    \begin{tabular}{llccc}
        \toprule
        \textbf{Prompt} & \textbf{Judge} & \textbf{Ken. $\tau$ ($\uparrow$)} & \textbf{Spr. $\rho$ ($\uparrow$)} & \textbf{RMSE ($\downarrow$)} \\
        \midrule
        
        \multirow{3}{*}{Chart2Code} 
            & GPT 4.1       & \textbf{0.7254} & \textbf{0.7769} & \textbf{0.5474} \\
            & Qwen-VL-Max      & 0.5971 & 0.6402 & 0.6951 \\
            & Qwen3-VL-Plus    & 0.5736 & 0.6156 & 0.7293 \\
        \midrule
        
        \multirow{3}{*}{ChartMimic} 
            & GPT 4.1       & \textbf{0.7835} & \textbf{0.8252} & \textbf{0.4934} \\
            & Qwen-VL-Max      & 0.7241 & 0.7723 & 0.5575 \\
            & Qwen3-VL-Plus    & 0.7564 & 0.8004 & 0.5265 \\
        \midrule
        
        \multirow{3}{*}{Plot2Code} 
            & GPT 4.1       & \textbf{0.7918} & \textbf{0.8355} & \textbf{0.4780} \\
            & Qwen-VL-Max      & 0.7800 & 0.8240 & 0.4972 \\
            & Qwen3-VL-Plus    & 0.7623 & 0.7999 & 0.5229 \\
        \midrule
        
        \multirow{3}{*}{Self-Reflection} 
            & GPT 4.1       & \textbf{0.7979} & \textbf{0.8417} & \textbf{0.4661} \\
            & Qwen-VL-Max      & 0.7275 & 0.7712 & 0.5508 \\
            & Qwen3-VL-Plus    & 0.7214 & 0.7700 & 0.5575 \\
        \midrule
        
        \multirow{3}{*}{\textbf{Omni-I2C}} 
            & \cellcolor{gray!15} GPT 4.1       & \cellcolor{gray!15} \textbf{0.8304} & \cellcolor{gray!15} \textbf{0.8681} & \cellcolor{gray!15} \textbf{0.4284} \\
            & Qwen-VL-Max      & 0.7897 & 0.8323 & 0.4780 \\
            & Qwen3-VL-Plus    & 0.7931 & 0.8353 & 0.4780 \\
            
        \bottomrule
    \end{tabular}
    }
\vspace{-3mm}
\end{table}

\noindent\textbf{Code Level Evaluation.} 
Given the semantic complexity of code generation, we utilize an LLM-based auditor to compute Functional Coverage ($S_{func}$) and Parametric Accuracy ($S_{para}$). While this enables scalable and automatic assessment, a single LLM judge is prone to systematic biases and capability limitations. Aggregating multiple judges (e.g., via consensus-based voting~\citep{cobbe2021training}) yields a more stable evaluation signal by mitigating idiosyncratic errors. The prompt templates of code-level evaluation paradigms are shown in App~\ref{sec:appendix_setups}.

To examine judge reliability, we evaluate three frontier models—Claude Sonnet 4.5, Gemini 2.5 Pro, and GPT 5.3 codex—by eliciting rankings over 1,080 instances across three target models (InternVL 3.5-8B-Instruct, Qwen2.5-VL-72B-Instruct and Qwen3-VL-235B-A22B-Instruct). We then construct a Majority-Vote (MV) ranking as a silver standard for expert consensus. As shown in Tab.~\ref{tab:judge_code}, the judges exhibit high agreement overall (Kendall’s $W=0.76$, mean Kendall’s $\tau=0.63$, and mean Spearman’s $\rho=0.67$). Notably, the observed disagreements are small in magnitude, with a mean rank range of 0.58, suggesting that variances are largely restricted to local swaps rather than systematic inversions. We further evaluate each judge’s alignment with the MV: GPT 5.3 codex demonstrates the highest fidelity to the consensus, achieving an MV align score of 0.8514 ($\tau=0.8390, \rho=0.8643$). On a per-instance basis, GPT 5.3 codex is the judge most frequently closest to the MV ranking (36.58\% of cases), significantly outperforming its peers. 

Based on the results, we adopt GPT 5.3 codex as our final code judge. This choice provides a high-fidelity proxy for the consensus ensemble while optimizing for API cost. We note, however, that multi-judge aggregation remains a principled and more robust option for studies where computational budgets permit higher overhead.

\begin{table}[t]
\vspace{-2mm}
\centering
\small
\caption{\textbf{Judge agreement and alignment to majority vote (MV).} All values are computed over 774 indices and reported as mean$\pm$std unless otherwise noted. \emph{MV align} is the composite alignment to the MV ranking, defined as $\frac{1}{2}(\tau+\rho)$, where $\tau$ is Kendall's $\tau$ and $\rho$ is Spearman's $\rho$ against MV.}
\label{tab:judge_code}
\resizebox{\columnwidth}{!}{%
\begin{tabular}{@{}lcc@{}}
\toprule
\multicolumn{3}{@{}c@{}}{\textbf{Per-judge vs MV}} \\
\midrule
\textbf{Code Judge} & \textbf{MV align} $\uparrow$ & \textbf{Closest to MV (count/\%)} $\uparrow$ \\
\midrule
Claude Sonnet 4.5 & 0.8482 \; ($\tau$=0.8335, $\rho$=0.8629) & 276 / 35.62\% \\
Gemini 2.5 Pro    & 0.7556 \; ($\tau$=0.7368, $\rho$=0.7743) & 215 / 27.80\% \\
GPT 5.3 codex     & \textbf{0.8517} \; ($\tau$=0.8390, $\rho$=0.8643) & \textbf{283 / 36.58\%} \\
\midrule
\multicolumn{3}{@{}c@{}}{\textbf{Overall (all judges)}} \\
\midrule
Kendall's $W$ $\uparrow$       & \multicolumn{2}{c}{0.7618 $\pm$ 0.2482} \\
Kendall's $\tau$ $\uparrow$    & \multicolumn{2}{c}{0.6315 $\pm$ 0.3668} \\
Spearman's $\rho$ $\uparrow$   & \multicolumn{2}{c}{0.6781 $\pm$ 0.3660} \\
Rank range (mean) $\downarrow$ & \multicolumn{2}{c}{0.58} \\
\bottomrule
\end{tabular}%
}
\vspace{-3mm}
\end{table}

\begin{table}[t]
\centering
\tiny
\setlength{\tabcolsep}{4pt} 
\renewcommand{\arraystretch}{1.1} 
\caption{Studies on different inference prompting methods. The best results for each model are highlighted in \textbf{bold}. Detailed cross-lingual results are provided in App.~\ref{sec:appendix_prompt}.}
\label{tab:different_prompts}
\begin{tabular}{l | c | ccc | c}
\toprule
\textbf{Model} & \textbf{Method} & $S_{\text{func}}$ & $S_{\text{para}}$ & $S_{\text{img}}$ & \textbf{Exec} \\
\midrule
\multirow{5}*{Gemini 2.5 Pro} & Direct & 57.81 & 39.92 & 71.01 & 89.49 \\
& Few-Shot & 59.52 & 39.76 & 72.09 & 89.70 \\
& Scaffold & 51.31 & 35.87 & 56.92 & 84.28 \\
& Self-Commentary & \textbf{60.11} & \textbf{41.36} & \textbf{72.37} & \textbf{94.18} \\
& Hint-Enhanced & 58.42 & 39.62 & 71.21 & 89.72 \\
\hline
\multirow{5}*{Qwen3-VL-235B-A22B-Instruct} & Direct & 55.78 & \textbf{40.03} & 62.58 & 88.16 \\
& Few-Shot & 56.15 & 39.99 & 63.13 & 88.47 \\
& Scaffold & 50.01 & 35.71 & 52.21 & 84.29 \\
& Self-Commentary & 55.36 & 39.78 & 63.56 & 89.49 \\
& Hint-Enhanced & \textbf{57.03} & 39.92 & \textbf{69.99} & \textbf{93.79} \\
\hline
\multirow{5}*{InternVL3.5-8B-Instruct} & Direct & 26.77 & 23.14 & 31.79 & 72.04 \\
& Few-Shot & 27.27 & 22.53 & 33.71 & 75.00 \\
& Scaffold & 18.02 & 16.31 & 19.52 & 61.02 \\
& Self-Commentary & 26.41 & 22.94 & 32.84 & 75.51 \\
& Hint-Enhanced & \textbf{28.56} & \textbf{24.11} & \textbf{42.31} & \textbf{88.40} \\
\hline
\multirow{5}*{Average} & Direct & 46.79 & 34.36 & 55.13 & 83.23 \\
& Few-Shot & 47.65 & 34.09 & 56.31 & 84.39 \\
& Scaffold & 39.78 & 29.30 & 42.88 & 76.53 \\
& Self-Commentary & 47.29 &  \textbf{34.69} & 56.26 & 86.39 \\
& Hint-Enhanced &  \textbf{48.00} & 34.55 &  \textbf{61.17} &  \textbf{90.64} \\
\bottomrule
\end{tabular}%
\vspace{-3mm}
\end{table}

\begin{figure*}[t]
\vspace{-2mm}
  \centering
  \begin{subfigure}{0.6\textwidth}
    \includegraphics[width=\textwidth]{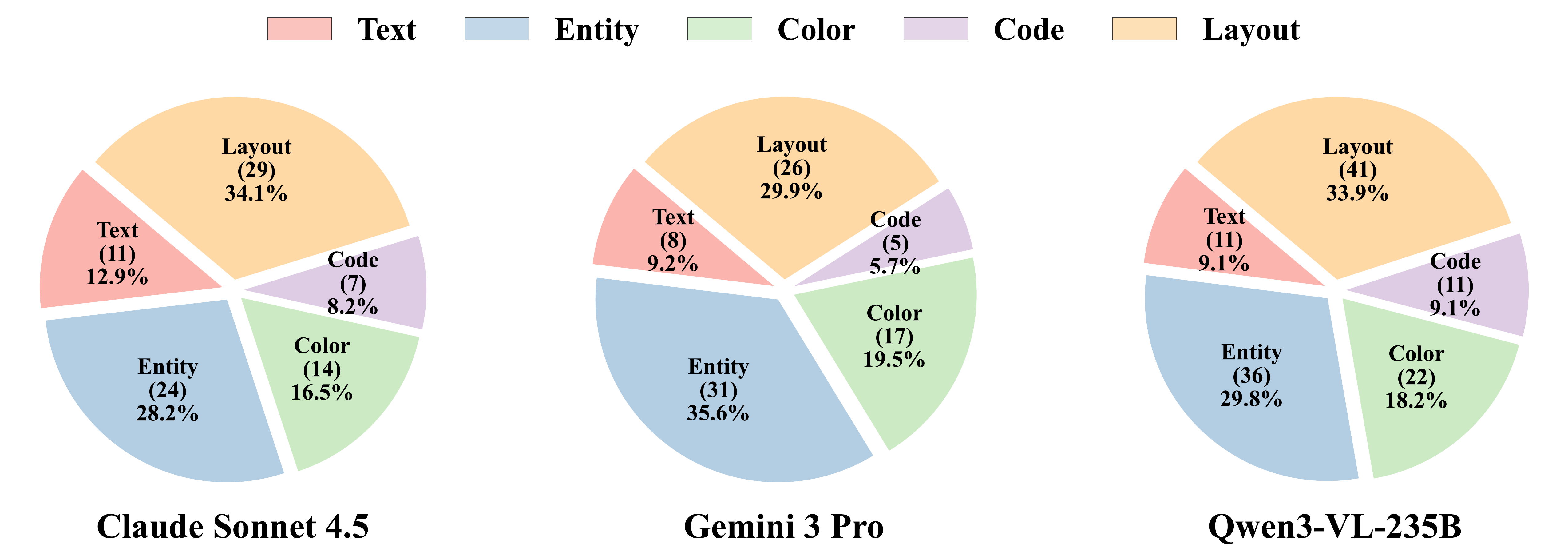}
    \caption{Distribution of Error Patterns}
  \end{subfigure}
  \begin{subfigure}{0.35\textwidth}
    \includegraphics[width=\textwidth]{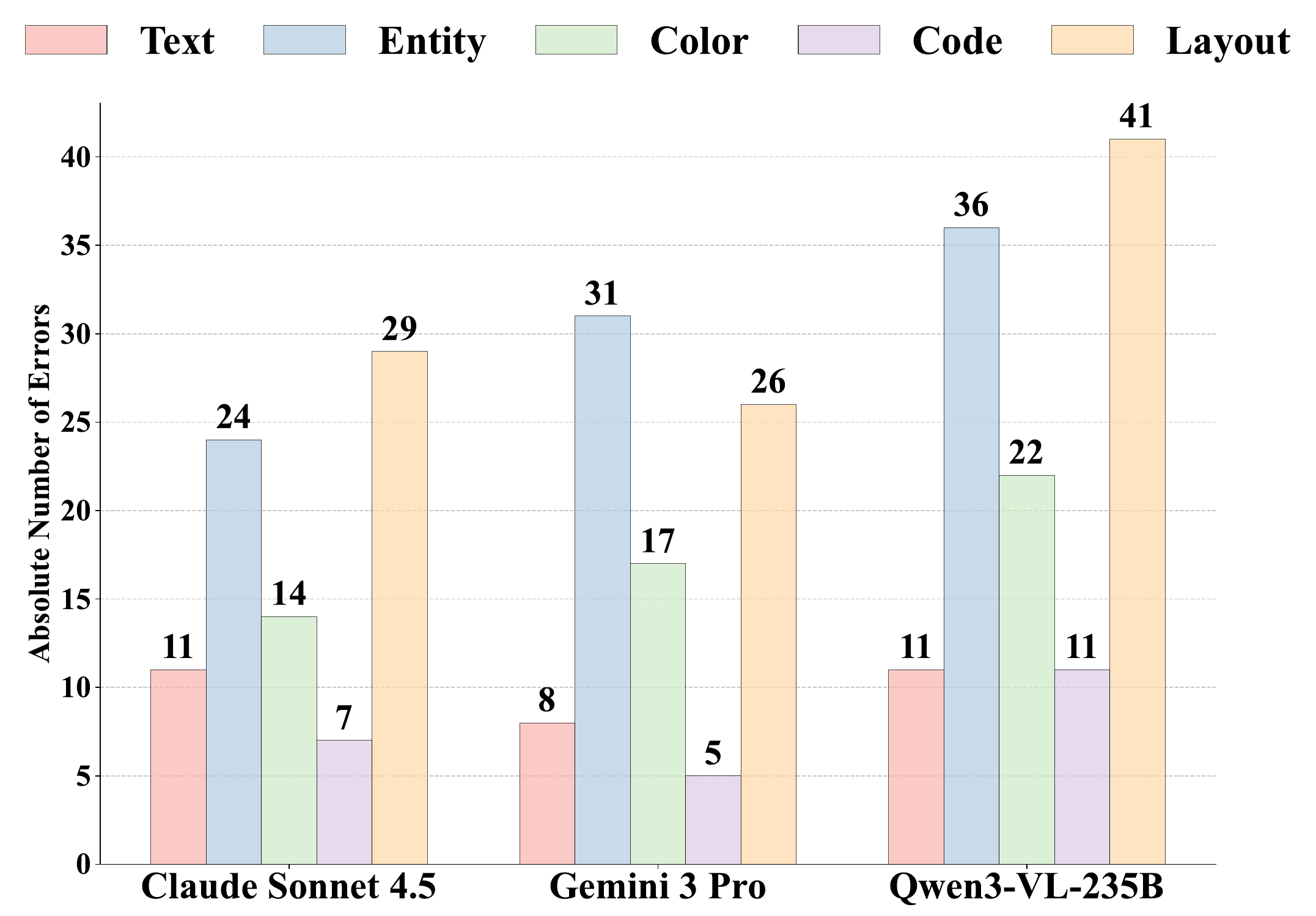}
    \caption{Frequency of Error Patterns}
  \end{subfigure}
  \caption{Comparative Analysis of Error Patterns in Claude Sonnet 4.5, Gemini 3 Pro, and Qwen3-VL-235B-A22B-Instruct. Case studies and evaluations across various languages are detailed in the App.~\ref{sec:appendix_err}.}
    \label{fig:error_analysis}
    \vspace{-2mm}
\end{figure*}

\subsection{Different Prompting Methods}
\label{sec:prompt_main}
We evaluate three representative models—Gemini 2.5 Pro, Qwen3-VL-235B-A22B-Instruct, and InternVL 3.5-8B-Instruct—under five distinct prompting strategies to assess their impact on Omni-I2C performance. These include: (i) Direct (zero-shot); (ii) Few-Shot (in-context examples); (iii) Hint-Enhanced, eliciting visual rationales before code generation; (iv) Scaffold, which overlays a coordinate grid for spatial grounding; and (v) Self-Commentary, which interleaves code with explanatory logic. Detailed implementations and full templates for each strategy are provided in App.~\ref{sec:appendix_prompt}. Our results in Tab.~\ref{tab:different_prompts} reveal a high sensitivity to prompting across all scales. For instance, when transitioning from Direct to Hint-Enhanced prompting, InternVL 3.5-8B-Instruct's overall execution rate (Exec) jumps by over 16\% (from 72.04\% to 88.40\%), alongside a notable 10.5\% increase in its visual reconstruction score ($S_{\text{img}}$, from 31.79\% to 42.31\%). Overall, reasoning-augmented strategies (\textit{Hint-Enhanced}, \textit{Self-Commentary}) consistently outperform baselines by externalizing the image-to-code translation process, achieving the highest average scores across most metrics. Conversely, \textit{Scaffold} prompting often degrades performance due to visual noise, while \textit{Few-Shot} remains a formidable baseline for syntactically rigid tasks like HTML. This high variance underscores that achieving ``plug-and-play'' stability remains an open challenge. An extended analysis of model behaviors is provided in App.~\ref{sec:appendix_prompt}.

\subsection{Error Analysis}
\label{sec:err}
To elucidate the fundamental bottlenecks in current LMMs, we conduct a manual audit of 240 instances (100 per model) on Claude Sonnet 4.5, Gemini 3 Pro, and Qwen3-VL-235B-A22B-Instruct. We define a five-tier taxonomy to categorize observed failure modes: (1) Code-level Logic, where syntactically valid code fails to map visual requirements to correct programmatic attributes; (2) Textual Precision, involving OCR failures in dense labels or complex symbolic notations; (3) Entity Integrity, characterized by the omission or misinterpretation of visual primitives like data points or legends; (4) Colorimetric Accuracy, where generated color codes deviate from the ground truth; and (5) Spatial Layout, involving distorted aspect ratios or erroneous coordinate transformations. 

As illustrated in Fig.~\ref{fig:error_analysis} (a), the error distributions exhibit a high degree of consistency across different models. Notably, Entity- and Layout-related errors are dominant, collectively accounting for over 60\% of the total. This indicates that current models continue to face significant challenges in recognizing visual entities and structures, as well as in implementing precise layout control through code. Furthermore, as depicted in Fig.~\ref{fig:error_analysis} (b), while the distributional patterns remain similar, the absolute volume of errors varies due to disparities in fundamental capabilities, such as visual perception and code generation. Consequently, Qwen3-VL-235B-A22B-Instruct exhibits a significantly higher error count compared to the other two models. A more granular analysis, including exhaustive definitions and qualitative case studies for each category, is provided in App.~\ref{sec:appendix_err}.

\section{Conclusion}
We introduce Omni-I2C, a benchmark for Image-to-Code generation characterized by its breadth across languages and subjects, and its depth in evaluation. By decoupling visual fidelity from symbolic precision, Omni-I2C exposes critical bottlenecks in I2C mapping. Our results reveal a significant performance gap even among frontier LMMs, establishing a substantial challenge to guide the development of more robust, visually-grounded multimodal agents.

\newpage
\section{Limitation}
\label{limitation}
Despite its multi-faceted evaluation design, Omni-I2C is subject to several limitations. Omni-I2C currently lacks natural scene images, a common gap in Image-to-Code research that hinders the evaluation of LMMs in generalizing to real-world visual complexities. Also, the dual-dimension evaluation relies on LLM/LMM-as-a-judge, which introduces non-negligible API costs and potential latency, limiting its scalability for rapid iterative testing. Future iterations of Omni-I2C will aim to bridge the ``natural image gap'' and investigate hybrid evaluation paradigms that balance semantic accuracy with resource efficiency, thereby setting more rigorous benchmarks for the community.

\section{Acknowledgments}
This work was supported in part by the New Generation Artificial Intelligence-National Science and Technology Major Project (No. 2025ZD0123602) and the National Natural Science Foundation of China (No. 62276203). The numerical calculations in this paper have been done in part on the supercomputing system in the Supercomputing Center of Wuhan University.
We also thank the contributors to the data collection and subsequent human alignment experiments. They include Chunqing Du, Qian Cai, Zhuo Song, Xiaoshu Yang, Yuhao Zhang, Junhan Liu, Zheyu Zheng, Jiahui Li, Jiaming Sun, Duojie Chen, Zihan Lou, Zhiwei Wang, Tong Liu, and Zhiyi Yang.

\bibliography{custom}
\clearpage
\appendix

\section{Dataset Details}
\label{sec:appendix_data}

This part provides exhaustive details regarding our data sourcing, the principles of curation, the human-centric annotation pipeline, figure type taxonomy and data rewriting details.

\subsection{Data Composition and Sourcing}
\label{sec:data_composition}
The construction of Omni-I2C follows a hybrid approach, integrating high-fidelity real-world samples with logically rigorous synthetic data to balance ecological validity and structural complexity. To ensure ethical and legal rigor, all real-world instances are exclusively curated from publicly accessible platforms under permissive licenses or copyright-free terms, ensuring full compliance for academic research and redistribution.

\paragraph{Real-world Data Sourcing} To reflect the diversity of practical applications, we systematically curated samples from 12 representative platforms. These sources span a wide range of subjects and visual modalities, providing a robust foundation for evaluating real-world performance. A comprehensive list of these platforms and their corresponding URLs is provided in Tab.~\ref{tab:data_sources}.

\paragraph{Synthetic Data Integration and Refinement} In addition to original collections, we incorporate selected data from existing benchmarks \cite{yang2024chartmimic, yang2025scaling} to further broaden our scope. Recognizing that raw synthetic data often lacks the necessary complexity or suffers from fidelity issues, we implemented a rigorous re-processing pipeline. This refinement procedure involves code refactoring and visual alignment.

\begin{table}[htbp]
\centering
\small 
\caption{Comprehensive list of real-world data sources.}
\label{tab:data_sources}
\begin{tabular}{lp{5.4cm}} 
\toprule
\textbf{Category} & \textbf{Source URL} \\
\midrule
\multirow{5}{*}{Python} & \url{https://matplotlib.org/stable/gallery/index.html} \\
                        & \url{https://networkx.org/documentation/stable/auto_examples/index.html} \\
                        & \url{https://plotly.com/python/} \\
                        & \url{https://altair-viz.github.io/gallery/index.html} \\
                        & \url{https://python-graph-gallery.com/} \\
\midrule
\multirow{4}{*}{LaTeX}  & \url{https://tikz.net/} \\
                        & \url{https://www.overleaf.com/gallery} \\
                        & \url{https://texample.net/} \\
                        & \url{https://ctan.mirrors.hoobly.com/graphics/pgf/contrib/pgfplots/doc/pgfplots.pdf} \\
\midrule
\multirow{2}{*}{HTML}   & \url{https://echarts.apache.org/examples/zh/index.html} \\
                        & \url{https://www.highcharts.com/demo} \\
\midrule
SVG                     & \url{https://commons.wikimedia.org/wiki/Main_Page} \\
\bottomrule
\end{tabular}
\end{table}

\subsection{Data Collection Principles}
To ensure the quality and diagnostic value of Omni-I2C, we adhere to three core curation principles during the real-world data collection process:

\noindent\textbf{Structural Correspondence.} We enforce strict congruence between code snippets and their rendered outputs. Every visual component must be directly traceable to specific logical blocks within the code. 

\noindent\textbf{Code Idiomaticity.} We prioritize implementations that utilize standard and idiomatic libraries (e.g., Matplotlib in Python or standard TikZ libraries). By avoiding obscure or idiosyncratic syntax, we ensure that the evaluation focuses on the model's fundamental perception, reasoning and coding  capabilities rather than its familiarity with ``edge-case'' language usage, thereby minimizing confounding variables during execution.

\noindent\textbf{Logical Non-triviality.} Our selection process favors visualizations with high structural and compositional complexity. We deliberately exclude rudimentary geometric primitives—such as isolated triangles or squares—in favor of multi-layered scientific plots, intricate symbolic notations, and nested UI components that require deep, multi-step reasoning to synthesize.

For synthetic data integration, our primary objective is to mitigate the data sparsity inherent in natural distributions. We strategically employ synthetic samples to supplement underrepresented or unseen categories identified in the real-world corpus, ensuring that the benchmark remains balanced across diverse subjects and programming paradigms.

\subsection{Human Annotation Pipeline}
The integrity of our 1,080 benchmark samples is guaranteed by a multi-stage human-in-the-loop process.

\paragraph{Annotator Background} 
We recruit 9 undergraduate students majoring in Computer Science or Artificial Intelligence. All participants had undergone basic research training and possessed prior experience in Python or web development, providing the necessary technical literacy for high-quality curation.

\paragraph{Training and Standardization} 
To ensure inter-annotator consistency, we conducted a rigorous training session prior to the mass annotation phase. Annotators are briefed on the ``Structural Correspondence'' requirement and trained on specific taxonomic guidelines to minimize ambiguity. A prime example of our classification protocol involves distinguishing between \textit{Analytic Geometry} and \textit{Function-related}. We enforced a strict mathematical definition: a \textit{Function-related} is defined by the property that every $x$ value corresponds to a unique $y$ value (i.e., the vertical line test), whereas \textit{Analytic Geometry} figures are not bound by this constraint. Cases with ambiguous classifications are adjudicated through a majority voting mechanism among senior annotators.

Following these protocols, annotators utilized a custom-built data collection platform that synchronizes code editing with real-time rendering to enforce a unified metadata schema. Upon uploading the source code and corresponding visual artifacts, annotators labeled each entry across three hierarchical dimensions: \textit{Subject}, \textit{Figure Type}, and \textit{Code Type}.

\begin{figure}[t]
    \centering
    \includegraphics[width=0.9\linewidth]{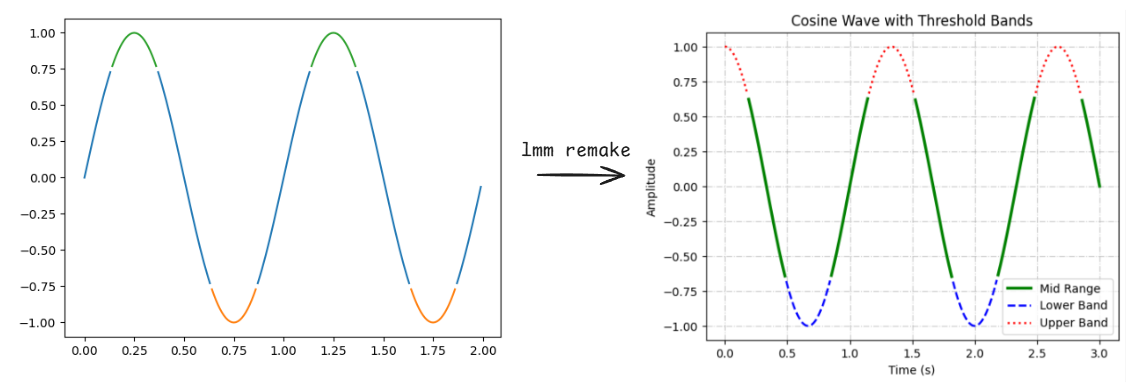}
    \caption{Examples of LLM-based code restructuring with foreground and numerical perturbations.}
    \label{fig:llm_rewrite}
\end{figure}
\begin{figure}[t]
    \centering
    \includegraphics[width=0.9\linewidth]{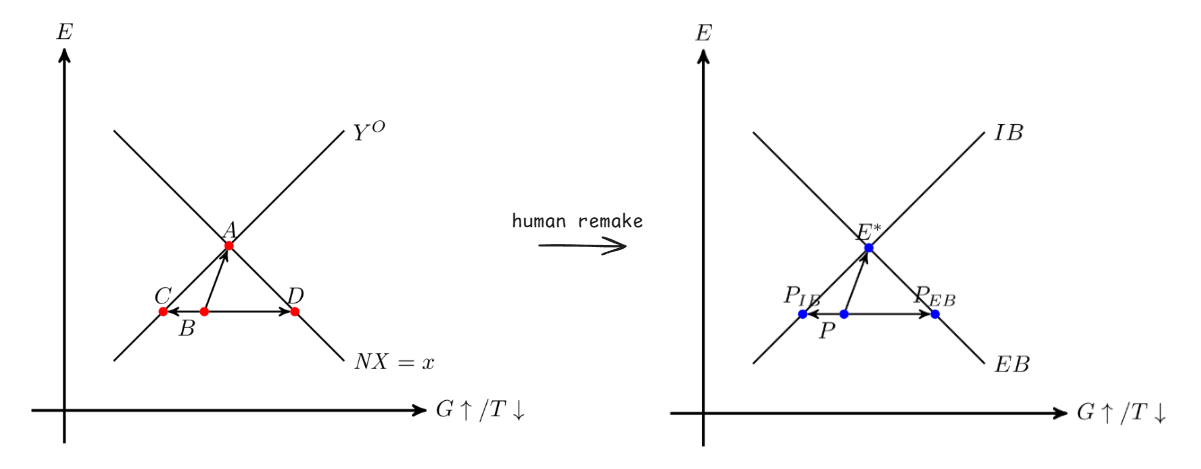}
    \caption{Comparison between original images and human-refined samples with modified key attributes.}
    \label{fig:human_rewrite}
\end{figure}
\begin{figure}[t]
    \centering
    \includegraphics[width=0.9\linewidth]{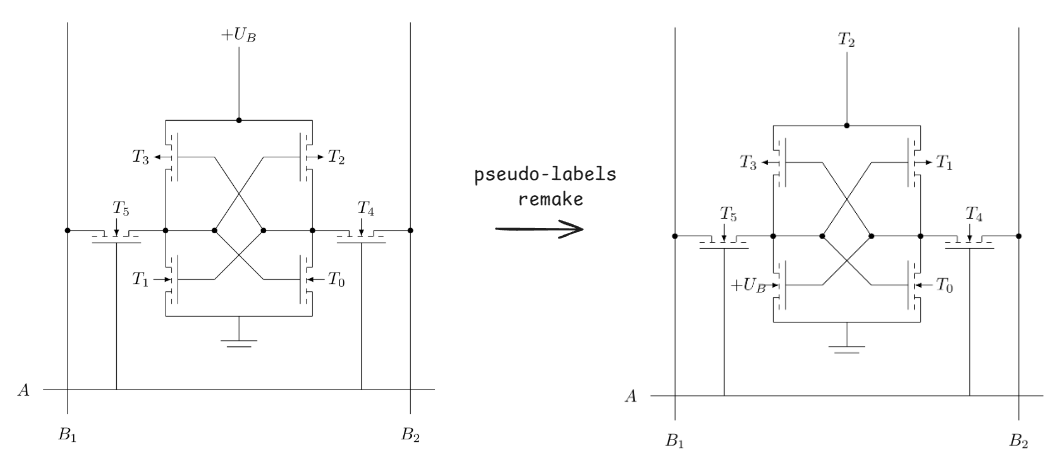}
    \caption{Visualization of counterfactual labels used to decouple perception from language priors.}
    \label{fig:counterfactual}
\end{figure}

\subsection{Figure Type Taxonomy}
\label{appendix:taxonomy}

To rigorously evaluate the model's code generation performance across diverse visual scenarios, Omni-I2C incorporates a granular classification system. Currently, Omni-I2C encompasses 45 figure types, covering core visual expressions in scientific research, engineering design, data analysis, and professional education. 

The specific categories include: 3d-plot, Algorithms, Area, Contour, Density, Graph, Histogram, Phase-Diagram, Quiver, Tree, UML-class-diagram, Violin, analytical-geometry, anatomy-diagram, atom-model, bar-chart, block-diagram, box-chart, cell-structure, circuit, equations-texts, error-bar, error-point, flow-chart, free-body-diagram, function-related, gauge-chart, heatmap, line-graph, magnetic-field-line, map, molecular-formula, multi-graph, optics-ray-diagram, other-figures, pie-chart, plane-geometry, radar-chart, relationship-diagram, scatter-plot, schematic, solid-geometry, tables, venn-diagram. Sparse or ambiguous categories are aggregated into a miscellaneous group to maintain statistical rigor.

\begin{figure*}[htbp]
    \centering
    \small
    \captionsetup[subfigure]{justification=raggedright, singlelinecheck=false}
    
    \begin{subfigure}[t]{0.24\textwidth}
        \centering
        \includegraphics[width=\linewidth]{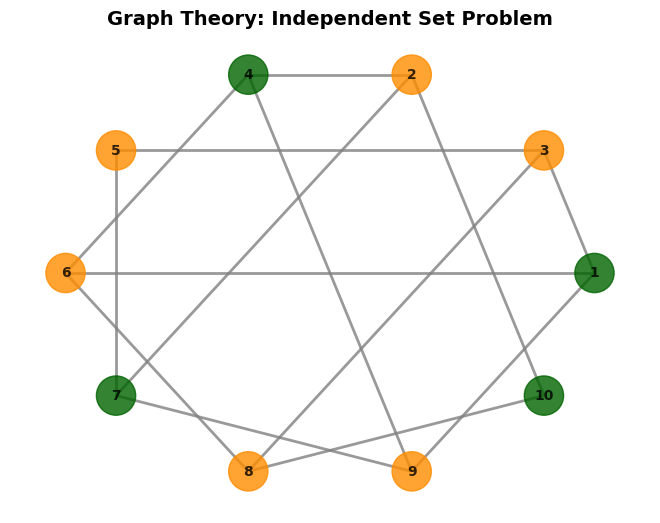}
        \caption{\textbf{Relationship-diagram}: Topological structures representing logical relationships between entities.}
    \end{subfigure}
    \hfill
    \begin{subfigure}[t]{0.24\textwidth}
        \centering
        \includegraphics[width=\linewidth]{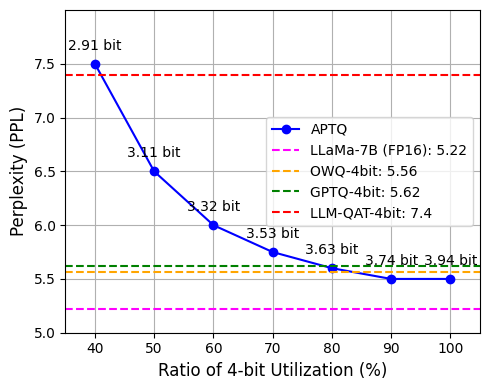}
        \caption{\textbf{Line-graph}: Curves depicting data trends over continuous variables.}
    \end{subfigure}
    \hfill
    \begin{subfigure}[t]{0.24\textwidth}
        \centering
        \includegraphics[width=\linewidth]{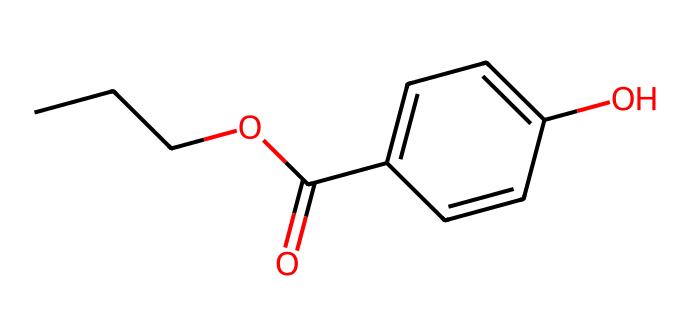}
        \caption{\textbf{Molecular-formula}: Chemical structures representing molecular compositions.}
    \end{subfigure}
    \hfill
    \begin{subfigure}[t]{0.24\textwidth}
        \centering
        \includegraphics[width=\linewidth]{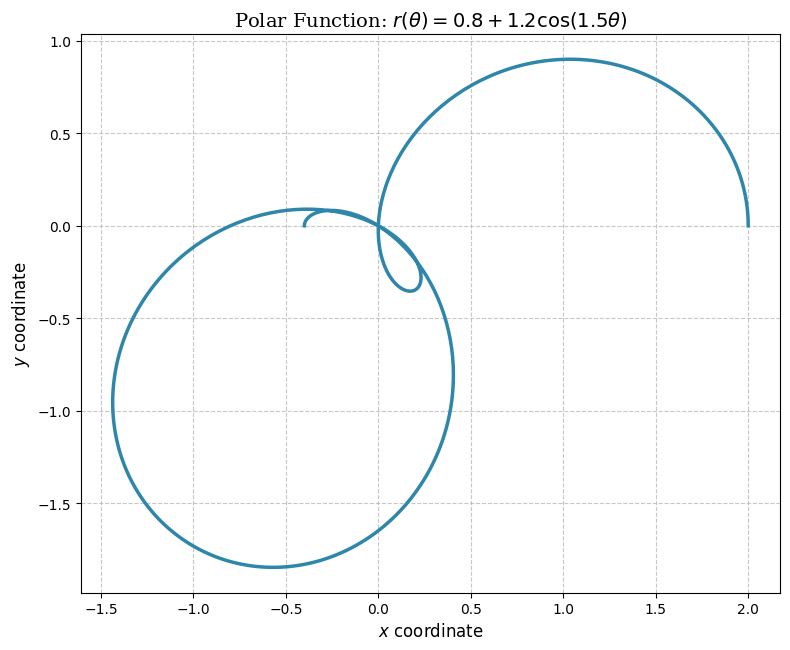}
        \caption{\textbf{Analytical-geometry}: Geometric shapes defined within a coordinate system.}
    \end{subfigure}

    \vspace{12pt}

    \begin{subfigure}[t]{0.24\textwidth}
        \centering
        \includegraphics[width=\linewidth]{figs/figure_type_examples/bar-chart_1.png}
        \caption{\textbf{Bar-chart}: Quantitative comparisons across discrete categories.}
    \end{subfigure}
    \hfill
    \begin{subfigure}[t]{0.24\textwidth}
        \centering
        \includegraphics[width=\linewidth]{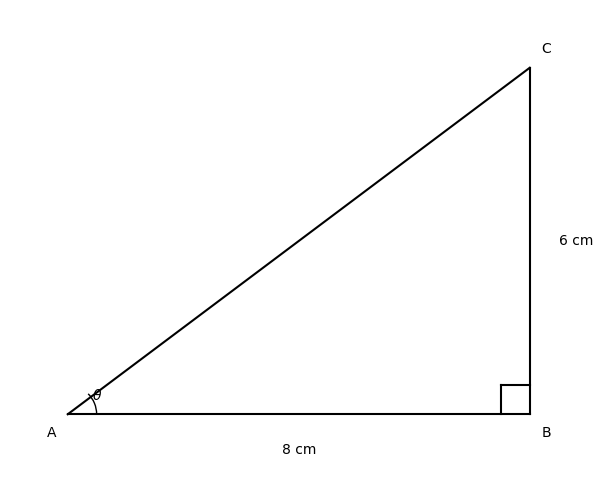}
        \caption{\textbf{Plane-geometry}: Shapes and properties within 2D Euclidean space.}
    \end{subfigure}
    \hfill
    \begin{subfigure}[t]{0.24\textwidth}
        \centering
        \includegraphics[width=\linewidth]{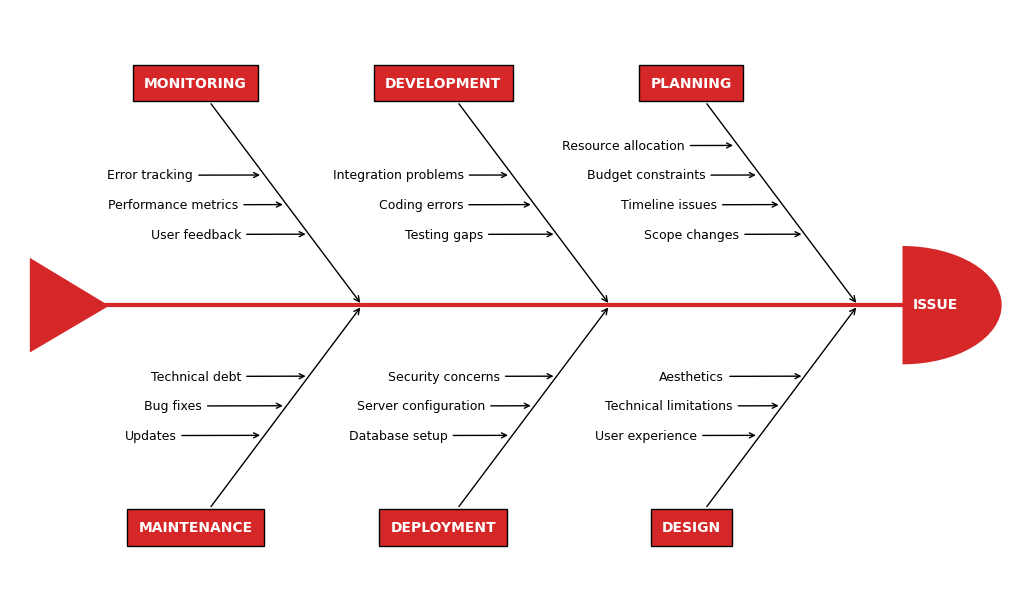}
        \caption{\textbf{Flow-chart}: Sequential steps of an algorithm or operational process.}
    \end{subfigure}
    \hfill
    \begin{subfigure}[t]{0.24\textwidth}
        \centering
        \includegraphics[width=\linewidth]{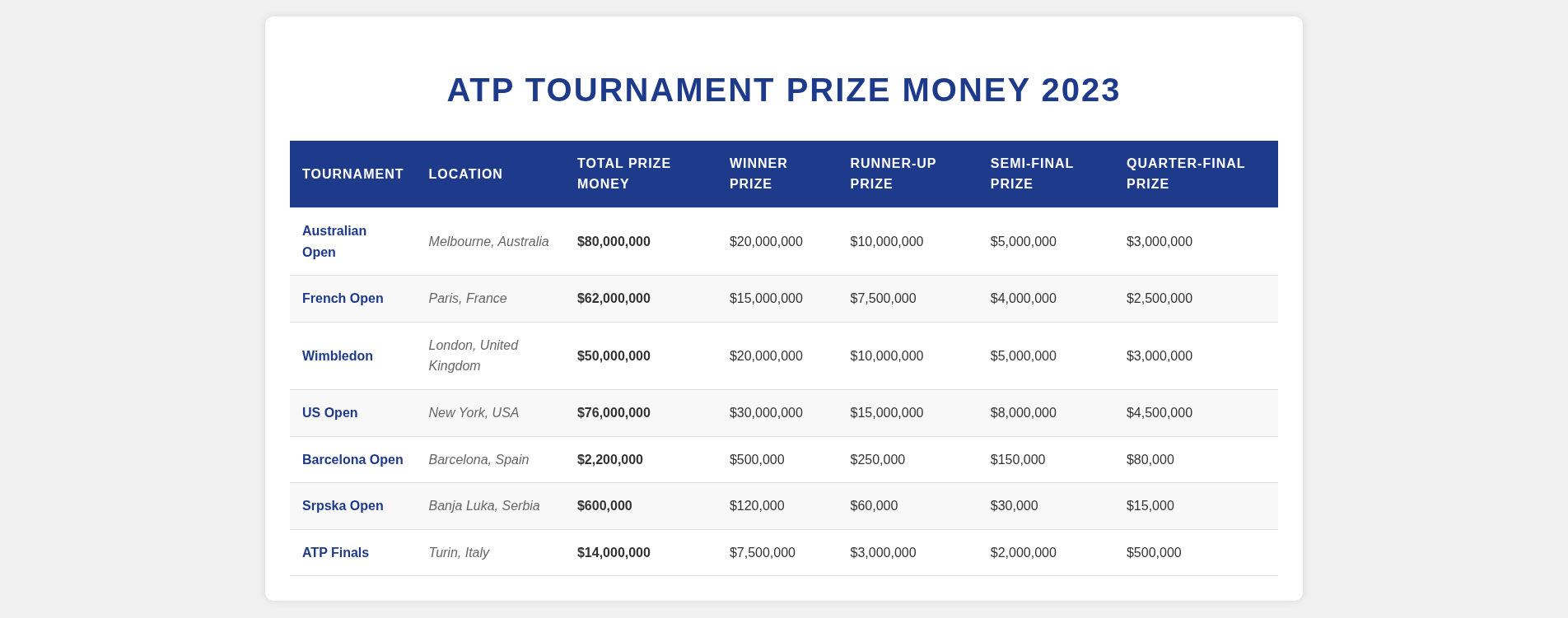}
        \caption{\textbf{Tables}: Systematic row-column layouts for structured data presentation.}
    \end{subfigure}

    \vspace{12pt}

    \begin{subfigure}[t]{0.24\textwidth}
        \centering
        \includegraphics[width=\linewidth]{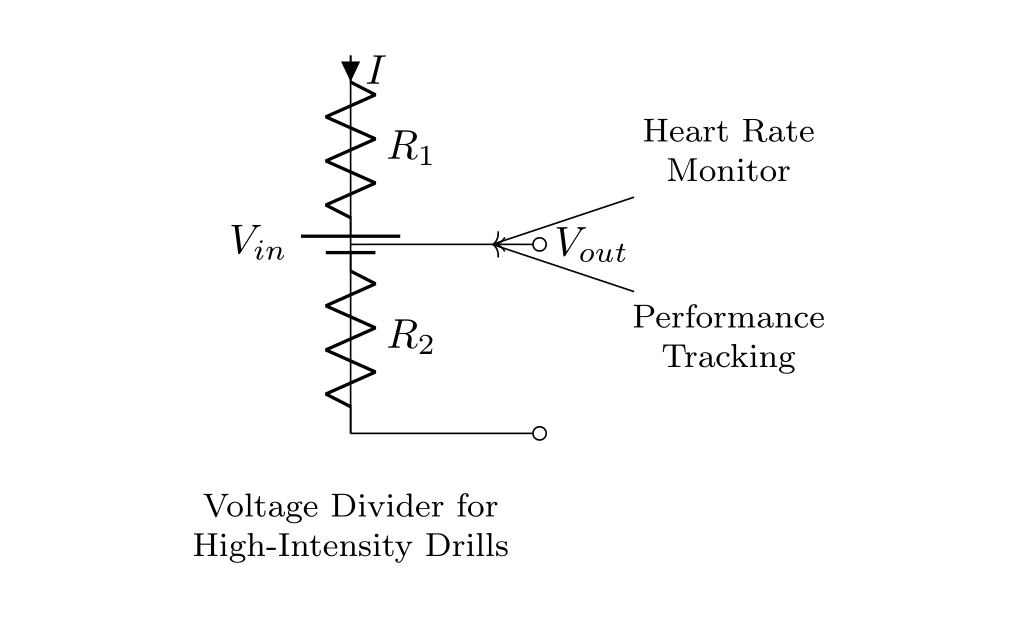}
        \caption{\textbf{Circuit}: Schematic diagrams of electronic components and connections.}
    \end{subfigure}
    \hfill
    \begin{subfigure}[t]{0.24\textwidth}
        \centering
        \includegraphics[width=\linewidth]{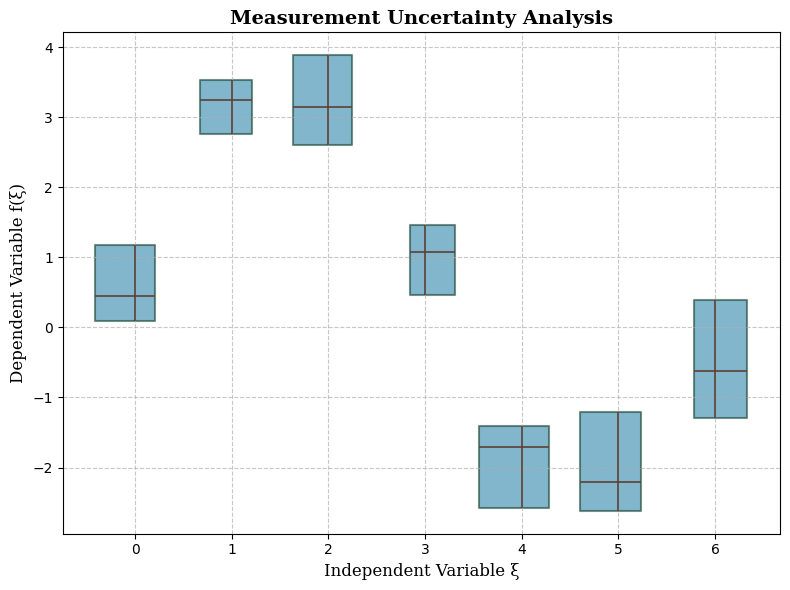}
        \caption{\textbf{Other-figures}: Miscellaneous images not covered by standard categories.}
    \end{subfigure}
    \hfill
    \begin{subfigure}[t]{0.24\textwidth}
        \centering
        \includegraphics[width=\linewidth]{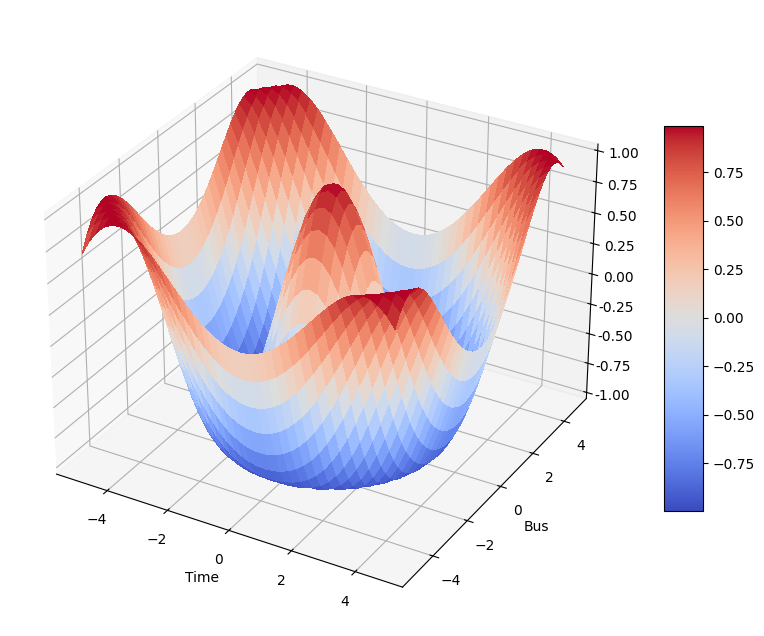}
        \caption{\textbf{3d-plot}: Surfaces or data distributions in three-dimensional space.}
    \end{subfigure}
    \hfill
    \begin{subfigure}[t]{0.24\textwidth}
        \centering
        \includegraphics[width=\linewidth]{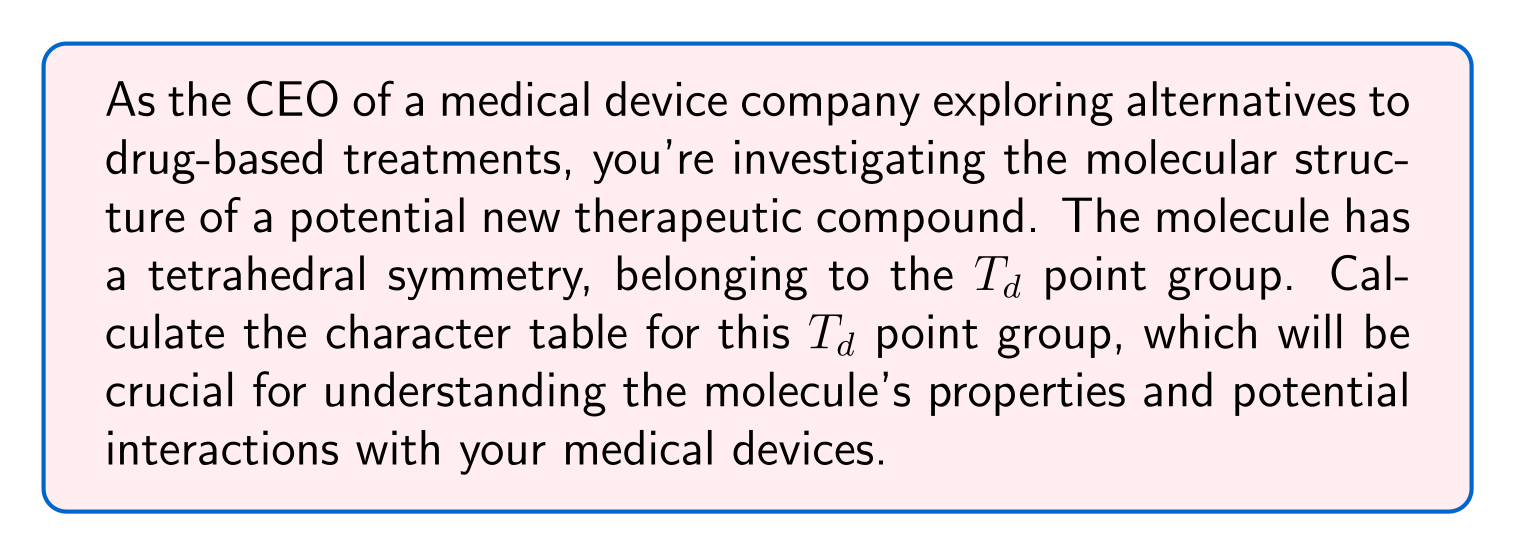}
        \caption{\textbf{Equations-texts}: Visualizations combining complex formulae and descriptive text.}
    \end{subfigure}
    
    \vspace{12pt}

    \begin{subfigure}[t]{0.24\textwidth}
        \centering
        \includegraphics[width=\linewidth]{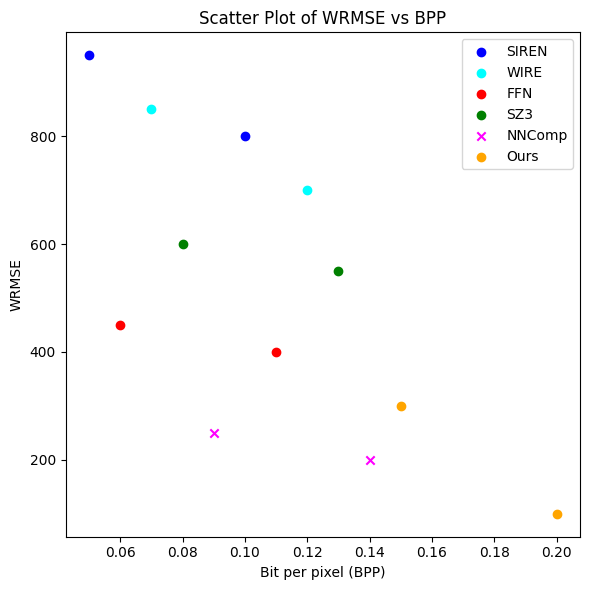}
        \caption{\textbf{Scatter-plot}: Distributions of data points in a 2D coordinate system.}
    \end{subfigure}
    \hfill
    \begin{subfigure}[t]{0.24\textwidth}
        \centering
        \includegraphics[width=\linewidth]{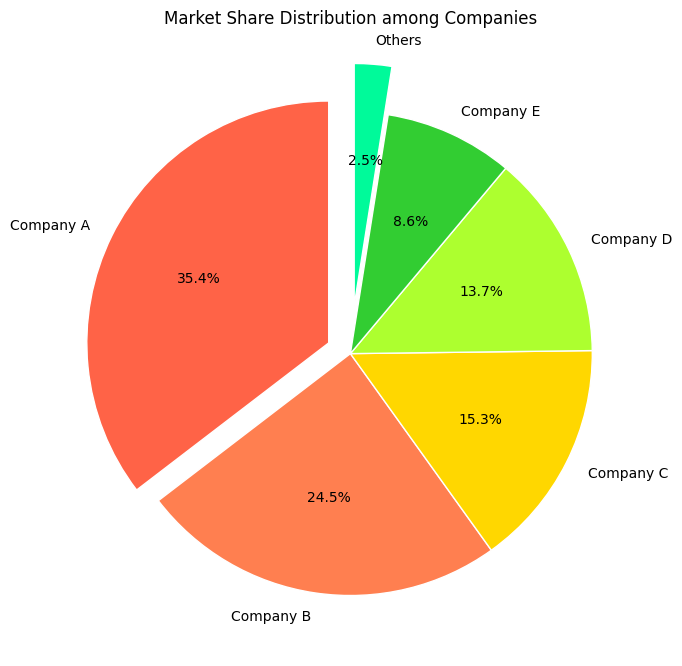}
        \caption{\textbf{Pie-chart}: Proportional composition of parts within a whole.}
    \end{subfigure}
    \hfill
    \begin{subfigure}[t]{0.24\textwidth}
        \centering
        \includegraphics[width=\linewidth]{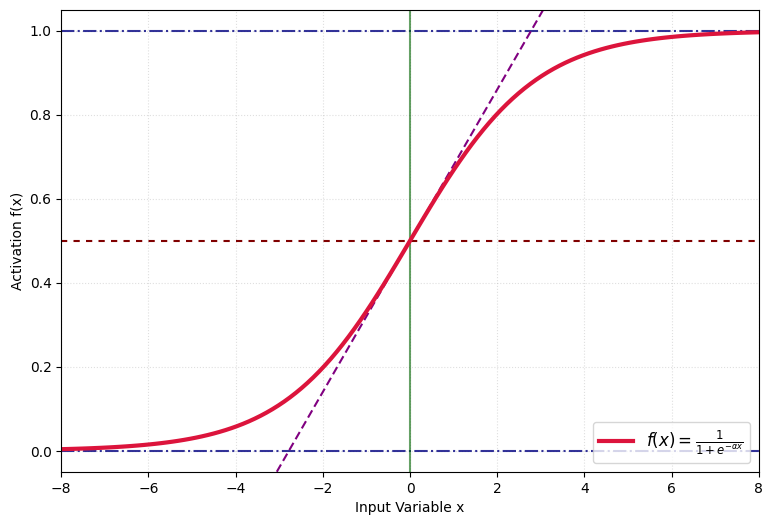}
        \caption{\textbf{Function-related}: Graphical expressions of mathematical functions.}
    \end{subfigure}
    \hfill
    \begin{subfigure}[t]{0.24\textwidth}
        \centering
        \includegraphics[width=\linewidth]{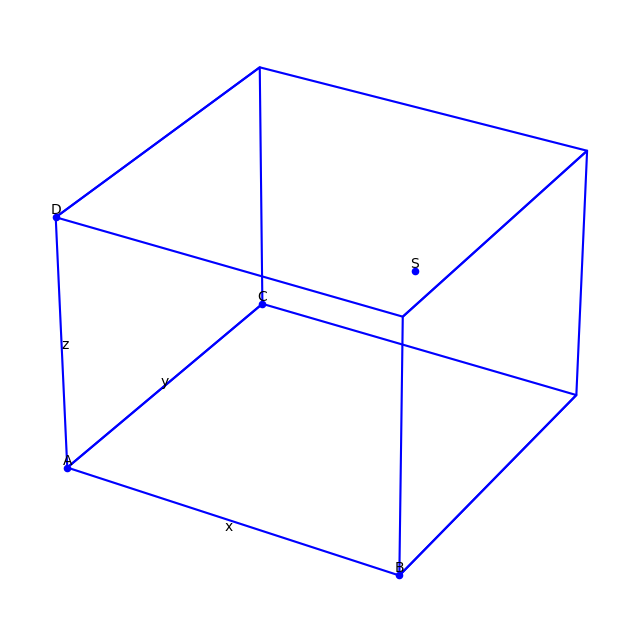}
        \caption{\textbf{Solid-geometry}: Volumetric and geometric features of 3D objects.}
    \end{subfigure}

    \vspace{12pt}
    
    \caption{Omni-I2C dataset: Figure Type Taxonomy gallery (Part 1).}
    \label{fig: appendix_figure_type_taxo1}
\end{figure*}

\begin{figure*}[htbp]
    \centering
    \small
    \captionsetup[subfigure]{justification=raggedright, singlelinecheck=false}
    \begin{subfigure}[t]{0.24\textwidth}
        \centering
        \includegraphics[width=\linewidth]{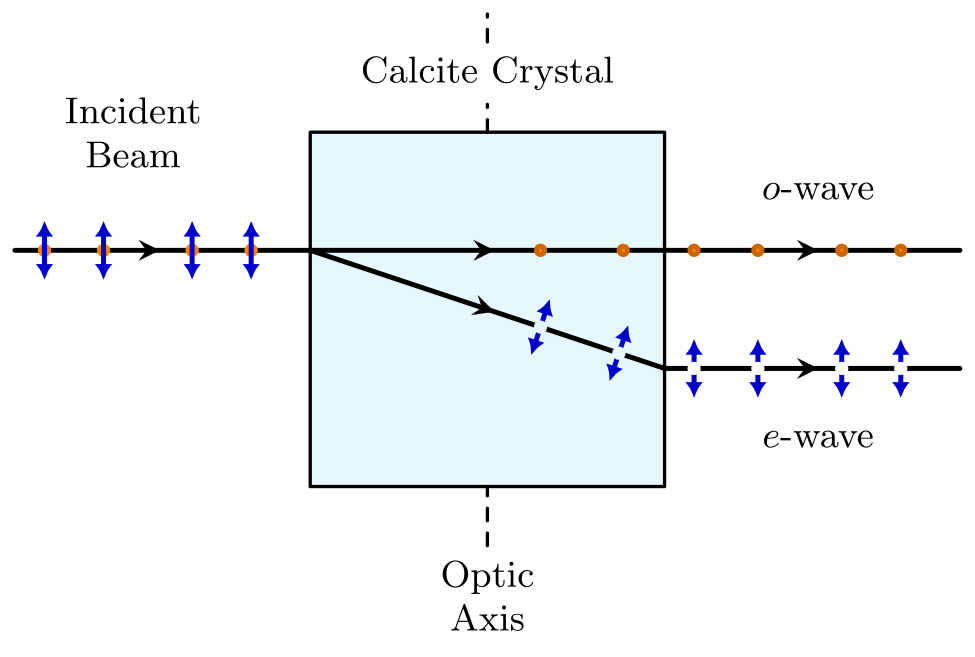}
        \caption{\textbf{Schematic}: Simplified diagrams illustrating systems or structures.}
    \end{subfigure}
    \hfill
    \begin{subfigure}[t]{0.24\textwidth}
        \centering
        \includegraphics[width=\linewidth]{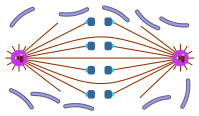}
        \caption{\textbf{Physiological-process}: Dynamic representations of biological activities.}
    \end{subfigure}
    \hfill
    \begin{subfigure}[t]{0.24\textwidth}
        \centering
        \includegraphics[width=\linewidth]{figs/figure_type_examples/Area_1.png}
        \caption{\textbf{Area}: Trend charts with filled area sections.}
    \end{subfigure}
    \hfill
    \begin{subfigure}[t]{0.24\textwidth}
        \centering
        \includegraphics[width=\linewidth]{figs/figure_type_examples/box-chart_1.png}
        \caption{\textbf{Box-chart}: Statistical summaries showing quartiles and outliers.}
    \end{subfigure}

    \vspace{12pt}

    \begin{subfigure}[t]{0.24\textwidth}
        \centering
        \includegraphics[width=\linewidth]{figs/figure_type_examples/Graph_1.png}
        \caption{\textbf{Graph}: General node-edge connectivity representations.}
    \end{subfigure}
    \hfill
    \begin{subfigure}[t]{0.24\textwidth}
        \centering
        \includegraphics[width=\linewidth]{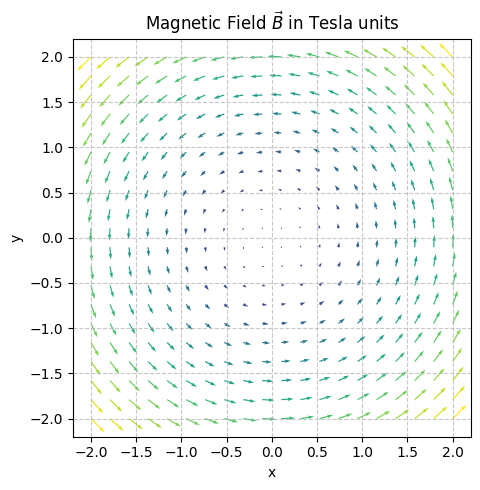}
        \caption{\textbf{Quiver}: Distribution of vector fields in a spatial domain.}
    \end{subfigure}
    \hfill
    \begin{subfigure}[t]{0.24\textwidth}
        \centering
        \includegraphics[width=\linewidth]{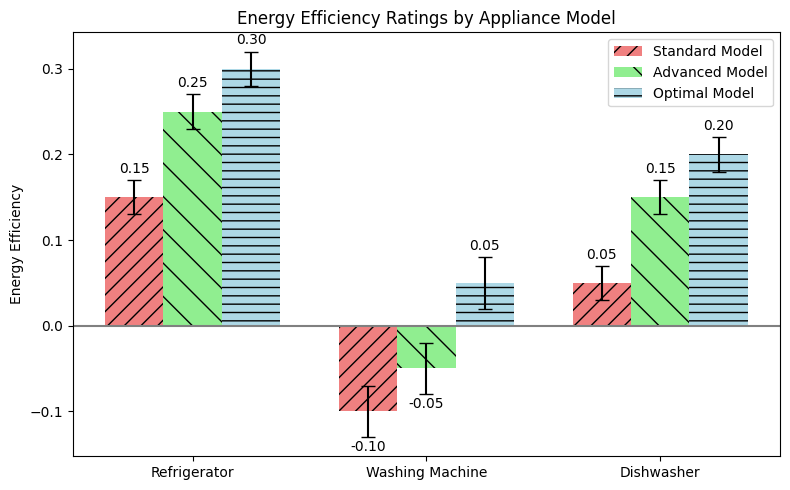}
        \caption{\textbf{Error-bar}: Statistical charts indicating data uncertainty ranges.}
    \end{subfigure}
    \hfill
    \begin{subfigure}[t]{0.24\textwidth}
        \centering
        \includegraphics[width=\linewidth]{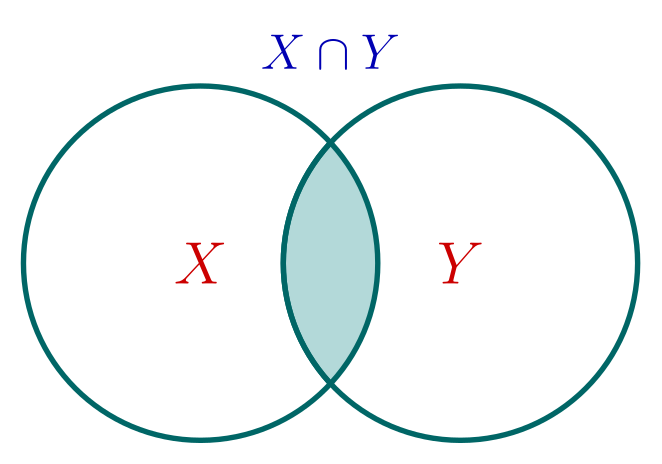}
        \caption{\textbf{Venn-diagram}: Overlapping sets showing logical relations.}
    \end{subfigure}

    \vspace{12pt}

    \begin{subfigure}[t]{0.24\textwidth}
        \centering
        \includegraphics[width=\linewidth]{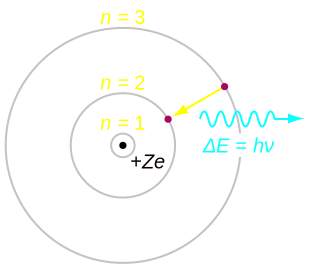}
        \caption{\textbf{Atom-model}: Physical models of atomic structures and electrons.}
    \end{subfigure}
    \hfill
    \begin{subfigure}[t]{0.24\textwidth}
        \centering
        \includegraphics[width=\linewidth]{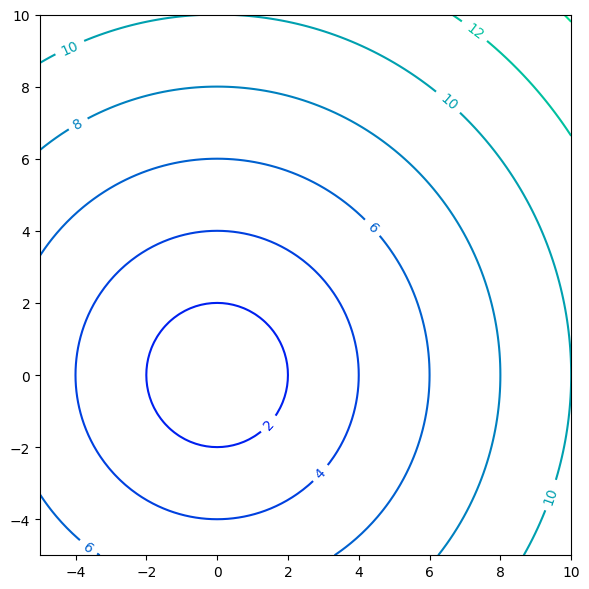}
        \caption{\textbf{Contour}: Numerical distributions with isoline features.}
    \end{subfigure}
    \hfill
    \begin{subfigure}[t]{0.24\textwidth}
        \centering
        \includegraphics[width=\linewidth]{figs/figure_type_examples/radar-chart_1.png}
        \caption{\textbf{Radar-chart}: Comparative visualizations of multi-dimensional data.}
    \end{subfigure}
    \hfill
    \begin{subfigure}[t]{0.24\textwidth}
        \centering
        \includegraphics[width=\linewidth]{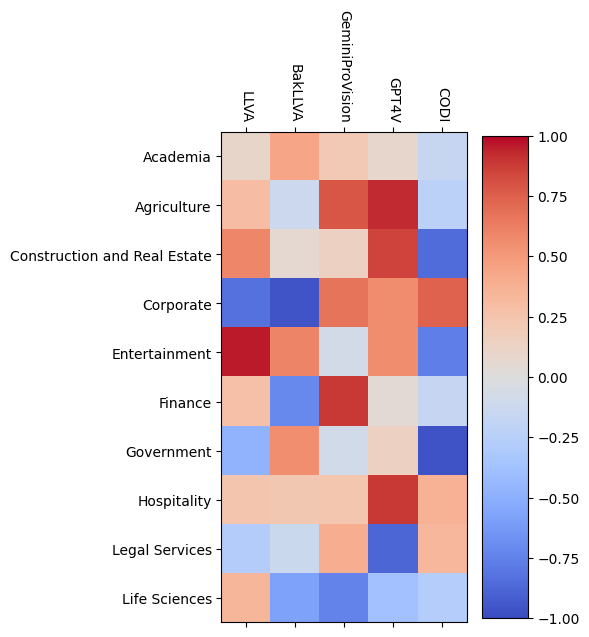}
        \caption{\textbf{Heatmap}: Matrix intensities represented by color gradients.}
    \end{subfigure}

    \vspace{12pt}

    \begin{subfigure}[t]{0.24\textwidth}
        \centering
        \includegraphics[width=\linewidth]{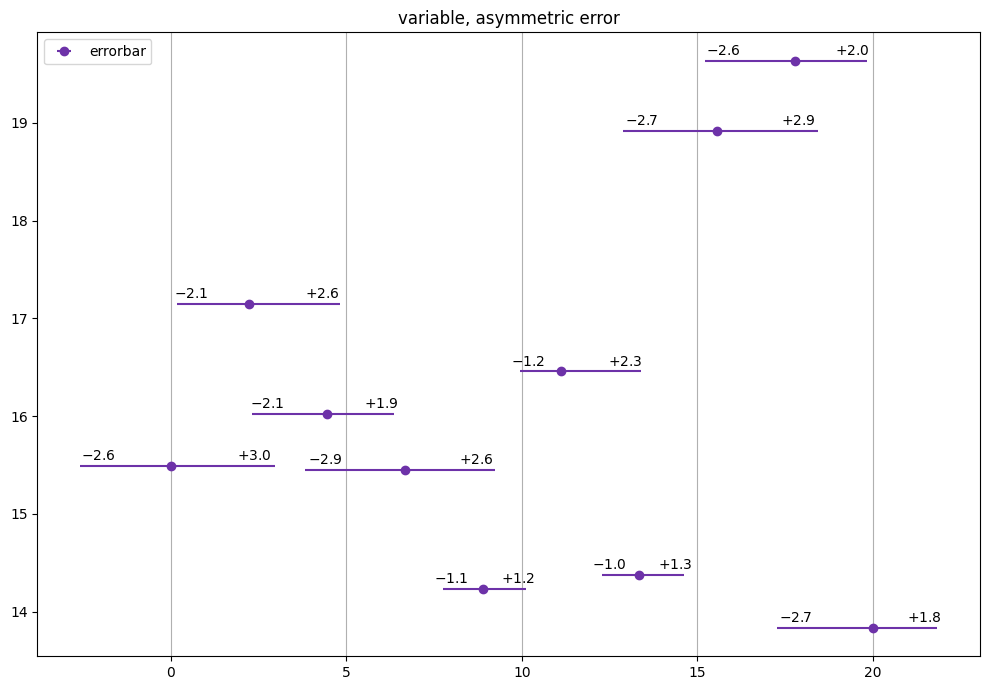}
        \caption{\textbf{Error-point}: Data points with designated error offsets.}
    \end{subfigure}
    \hfill
    \begin{subfigure}[t]{0.24\textwidth}
        \centering
        \includegraphics[width=\linewidth]{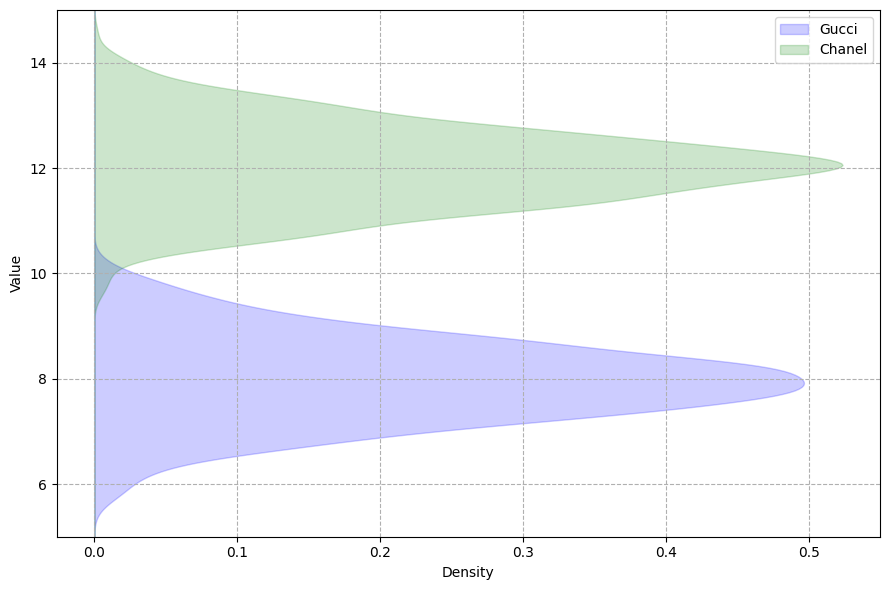}
        \caption{\textbf{Density}: Spatial distributions of probability densities.}
    \end{subfigure}
    \hfill
    \begin{subfigure}[t]{0.24\textwidth}
        \centering
        \includegraphics[width=\linewidth]{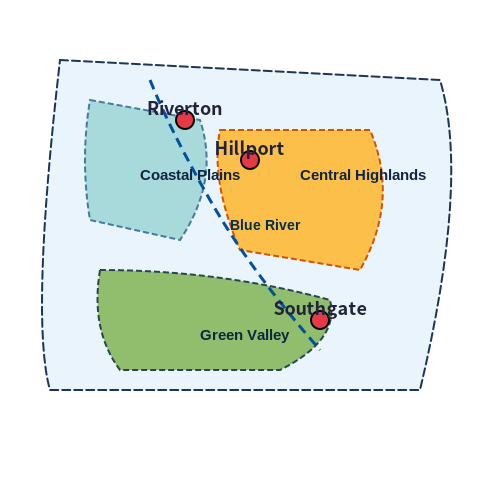}
        \caption{\textbf{Map}: Visualizations of geographic locations or regions.}
    \end{subfigure}
    \hfill
    \begin{subfigure}[t]{0.24\textwidth}
        \centering
        \includegraphics[width=\linewidth]{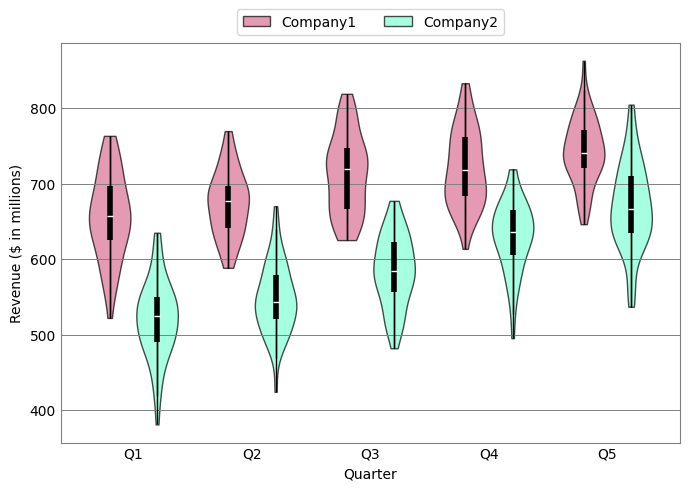}
        \caption{\textbf{Violin}: Distribution plots combining density and box plot features.}
    \end{subfigure}
    
    \caption{Omni-I2C dataset: Figure Type Taxonomy gallery (Part 2).}
    \label{fig:appendix_figure_type_taxo2}
\end{figure*}

\begin{figure*}[htbp]
    \centering
    \small
    \captionsetup[subfigure]{justification=raggedright, singlelinecheck=false}

    \begin{subfigure}[t]{0.24\textwidth}
        \centering
        \includegraphics[width=\linewidth, height=2.8cm, keepaspectratio]{figs/figure_type_examples/Histogram_1.png}
        \caption{\textbf{Histogram}: Statistical bars for frequency or count distributions.}
    \end{subfigure}
    \hfill
    \begin{subfigure}[t]{0.24\textwidth}
        \centering
        \includegraphics[width=\linewidth, height=2.8cm, keepaspectratio]{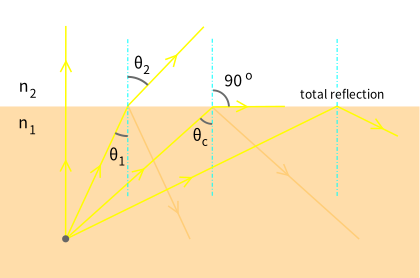}
        \caption{\textbf{Optics-ray-diagram}: Propagation paths of light in optical systems.}
    \end{subfigure}
    \hfill
    \begin{subfigure}[t]{0.24\textwidth}
        \centering
        \includegraphics[width=\linewidth, height=2.8cm, keepaspectratio]{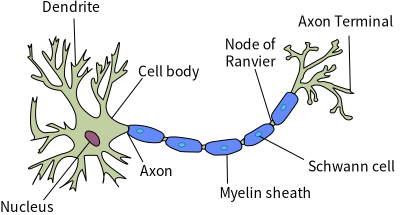}
        \caption{\textbf{Cell-structure}: Diagrams illustrating internal biological cell parts.}
    \end{subfigure}
    \hfill
    \begin{subfigure}[t]{0.24\textwidth}
        \centering
        \includegraphics[width=\linewidth, height=2.8cm, keepaspectratio]{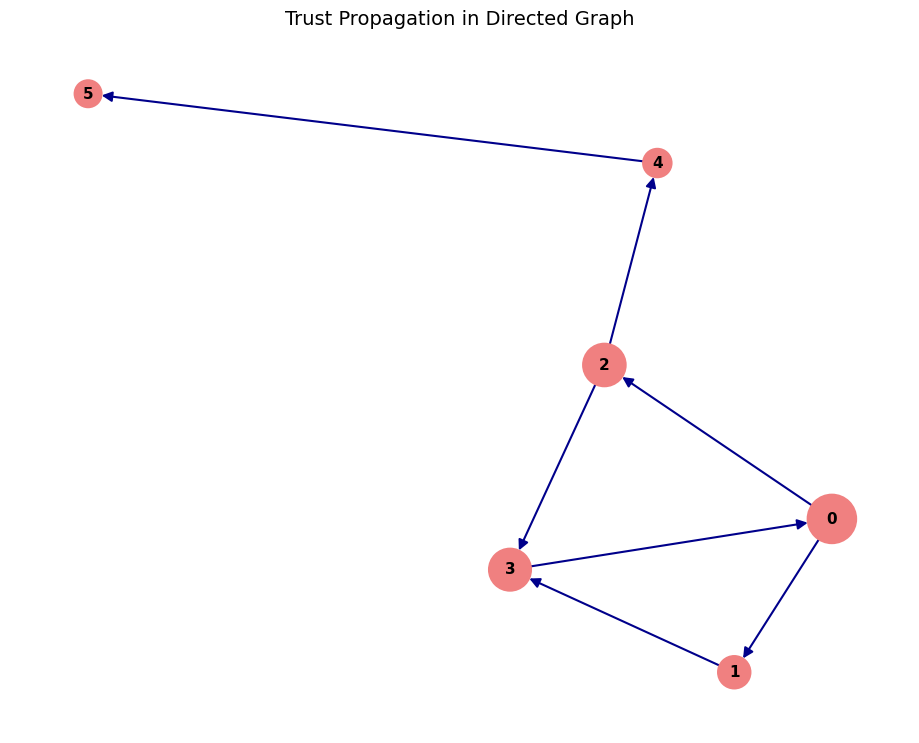}
        \caption{\textbf{Algorithms}: Visualization of algorithmic logic or pseudocode.}
    \end{subfigure}

    \vspace{15pt}

    \begin{subfigure}[t]{0.24\textwidth}
        \centering
        \includegraphics[width=\linewidth, height=2.8cm, keepaspectratio]{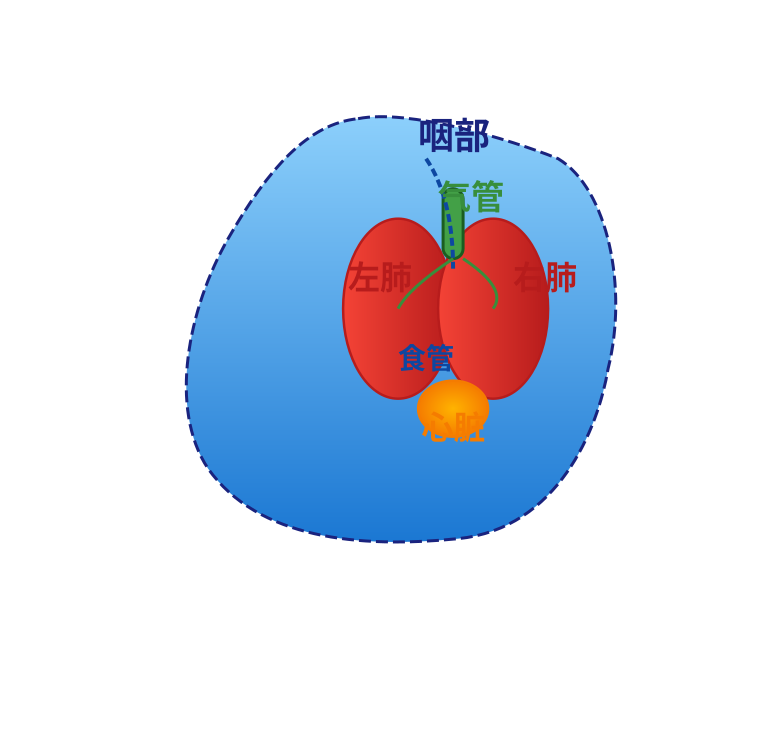}
        \caption{\textbf{Anatomy-diagram}: Detailed structural diagrams of biological anatomy.}
    \end{subfigure}
    \hfill
    \begin{subfigure}[t]{0.24\textwidth}
        \centering
        \includegraphics[width=\linewidth, height=2.8cm, keepaspectratio]{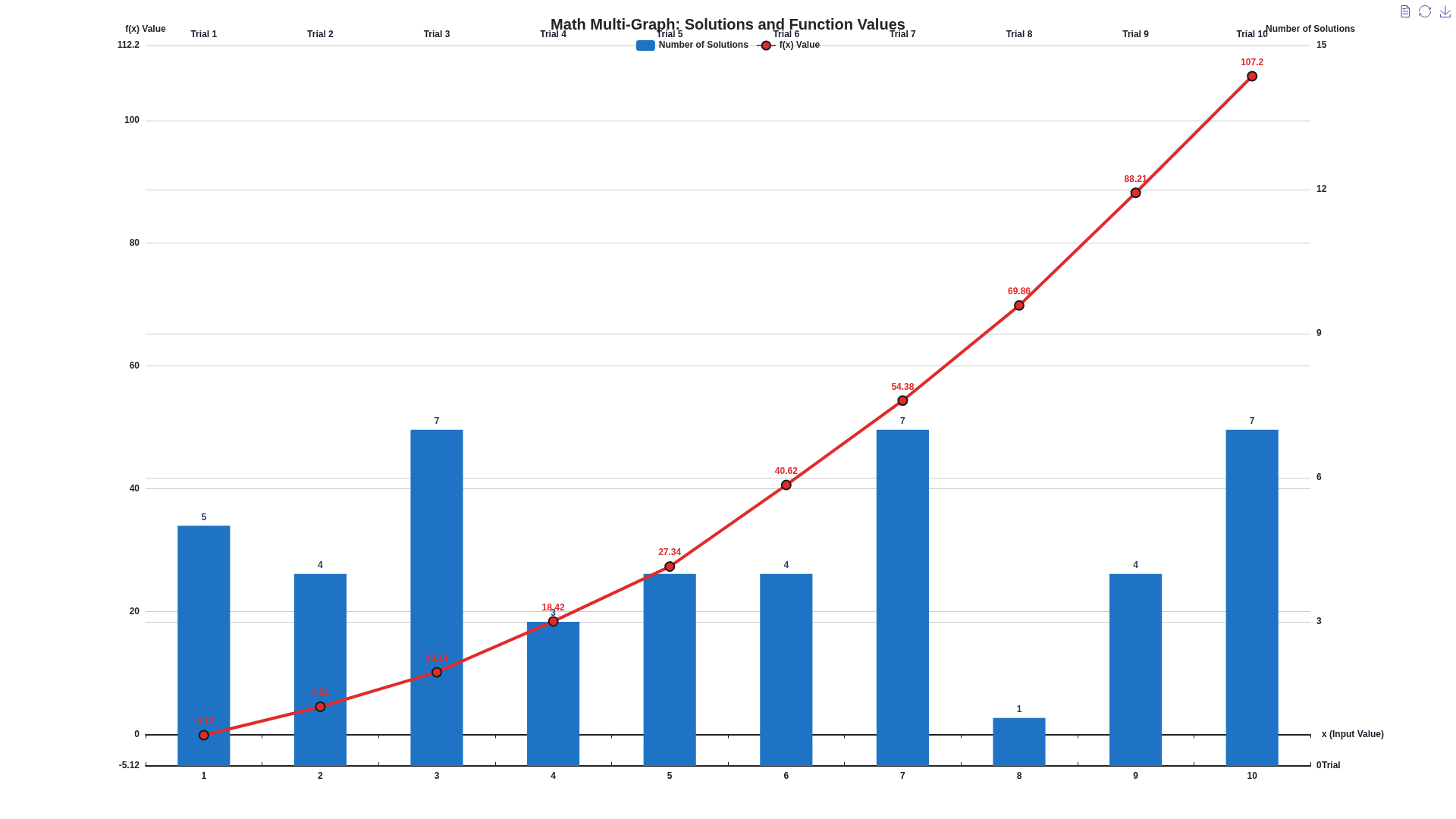}
        \caption{\textbf{Multi-graph}: Composite images containing multiple sub-figures.}
    \end{subfigure}
    \hfill
    \begin{subfigure}[t]{0.24\textwidth}
        \centering
        \includegraphics[width=\linewidth, height=2.8cm, keepaspectratio]{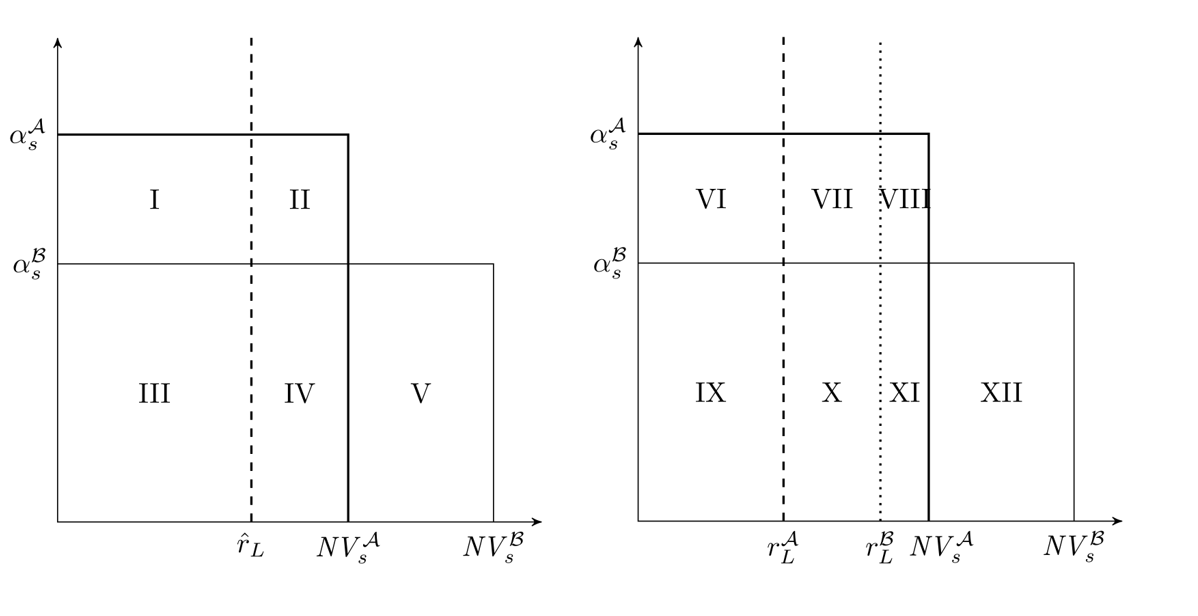}
        \caption{\textbf{Phase-Diagram}: State charts describing system phase transitions.}
    \end{subfigure}
    \hfill
    \begin{subfigure}[t]{0.24\textwidth}
        \centering
        \includegraphics[width=\linewidth, height=2.8cm, keepaspectratio]{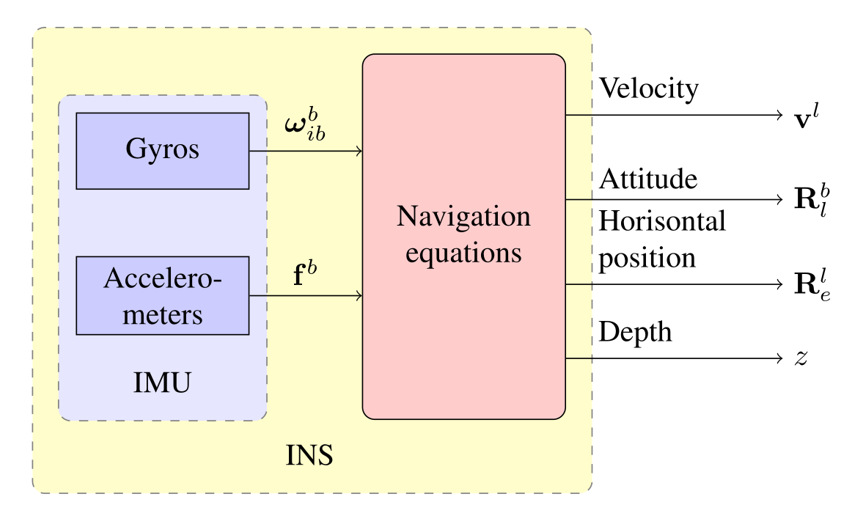}
        \caption{\textbf{Block-diagram}: Functional relationships between system modules.}
    \end{subfigure}

    \vspace{15pt}

    \begin{subfigure}[t]{0.24\textwidth}
        \centering
        \includegraphics[width=\linewidth, height=2.8cm, keepaspectratio]{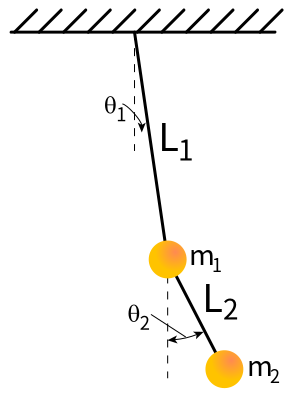}
        \caption{\textbf{Free-body-diagram}: Classical mechanics models for force analysis.}
    \end{subfigure}
    \hfill
    \begin{subfigure}[t]{0.24\textwidth}
        \centering
        \includegraphics[width=\linewidth, height=2.8cm, keepaspectratio]{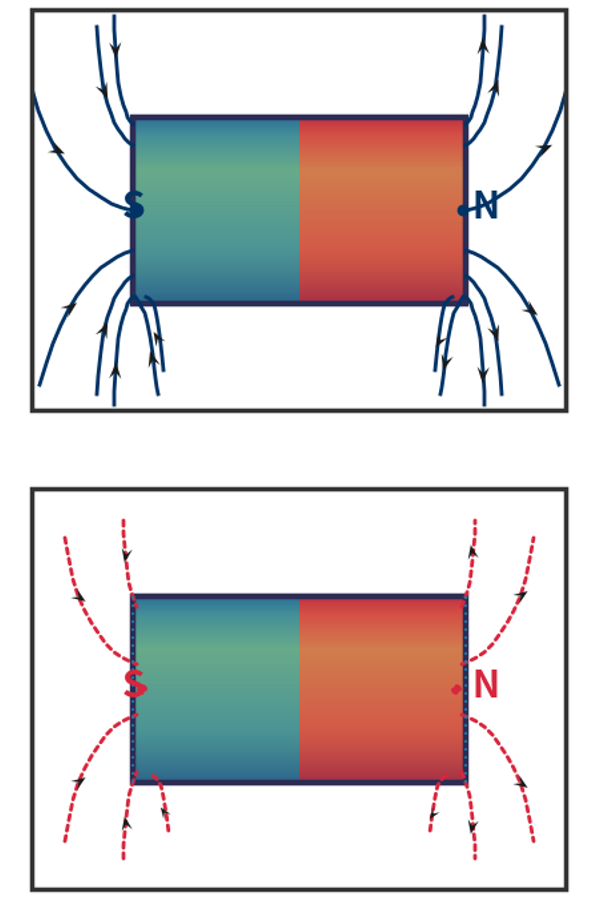}
        \caption{\textbf{Magnetic-field-line}: Orientations and distributions of magnetic fields.}
    \end{subfigure}
    \hfill
    \begin{subfigure}[t]{0.24\textwidth}
        \centering
        \includegraphics[width=\linewidth, height=2.8cm, keepaspectratio]{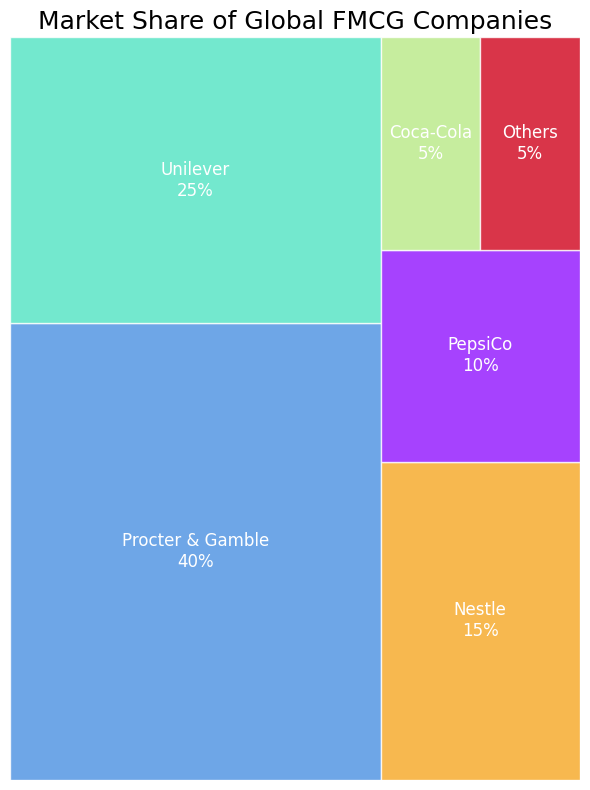}
        \caption{\textbf{Tree}: Hierarchical structures with parent-child relationships.}
    \end{subfigure}
    \hfill
    \begin{subfigure}[t]{0.24\textwidth}
        \centering
        \includegraphics[width=\linewidth, height=2.8cm, keepaspectratio]{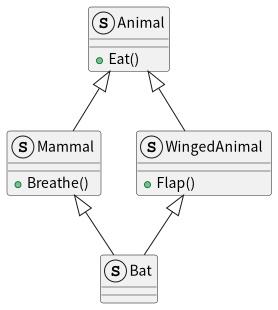}
        \caption{\textbf{UML-class-diagram}: Software engineering diagrams for class structures.}
    \end{subfigure}

    \vspace{15pt}

    \begin{subfigure}[t]{0.24\textwidth}
        \centering
        \includegraphics[width=\linewidth, height=2.8cm, keepaspectratio]{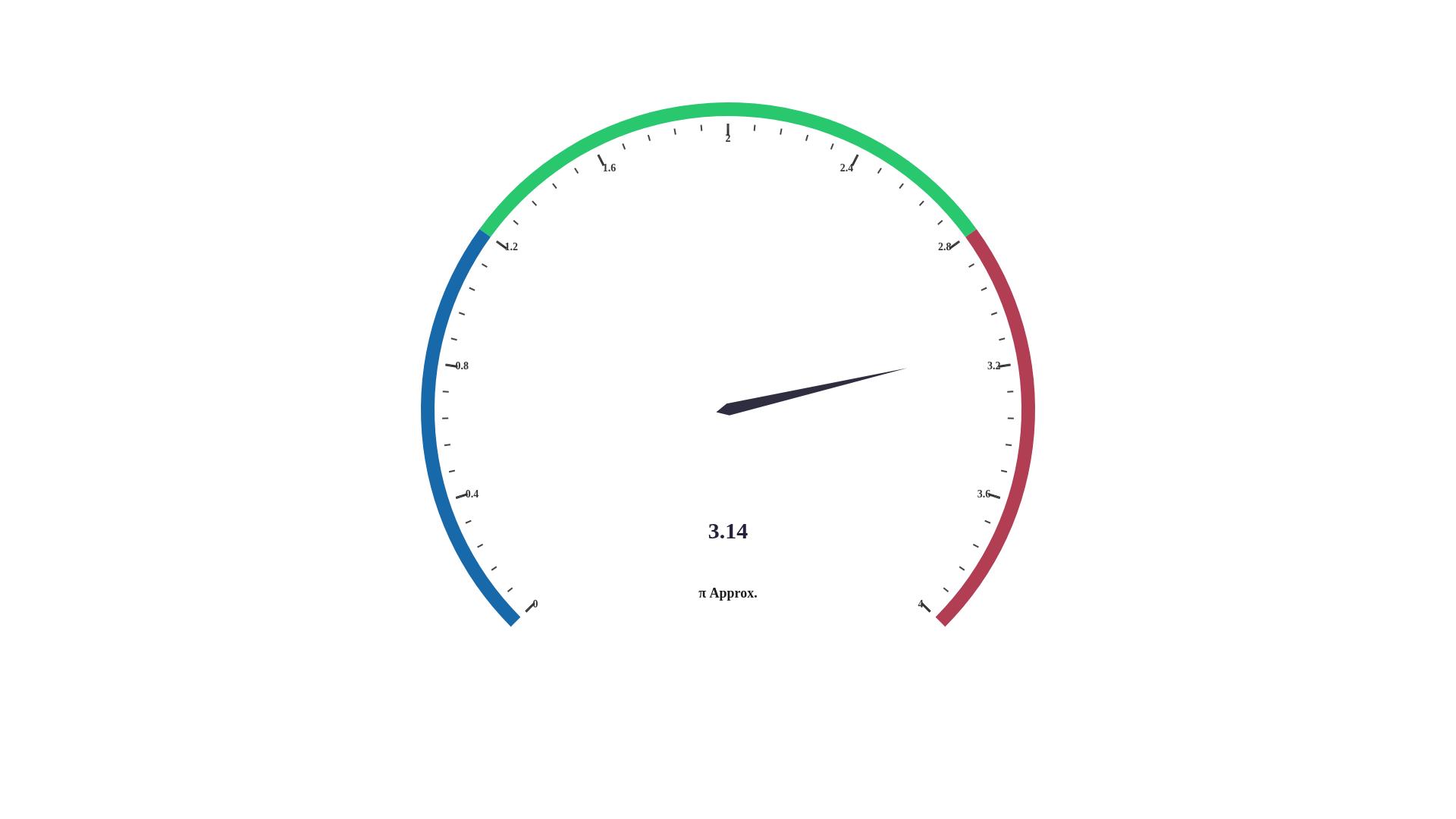}
        \caption{\textbf{Gauge-chart}: Radial scales showing specific data values.}
    \end{subfigure}
    \hfill
    \begin{minipage}[t]{0.24\textwidth} \end{minipage} \hfill 
    \begin{minipage}[t]{0.24\textwidth} \end{minipage} \hfill 
    \begin{minipage}[t]{0.24\textwidth} \end{minipage}        

    \vspace{10pt}
    \caption{Omni-I2C dataset: Figure Type Taxonomy gallery (Part 3).}
    \label{fig: appendix_figure_type_taxo3}
\end{figure*}

We provide illustrative examples for these image types in Fig.~\ref{fig: appendix_figure_type_taxo1}, Fig.~\ref{fig:appendix_figure_type_taxo2}, and Fig.~\ref{fig: appendix_figure_type_taxo3}.

\begin{figure}[t]
    \centering
    
    \includegraphics[width=1.0\linewidth]{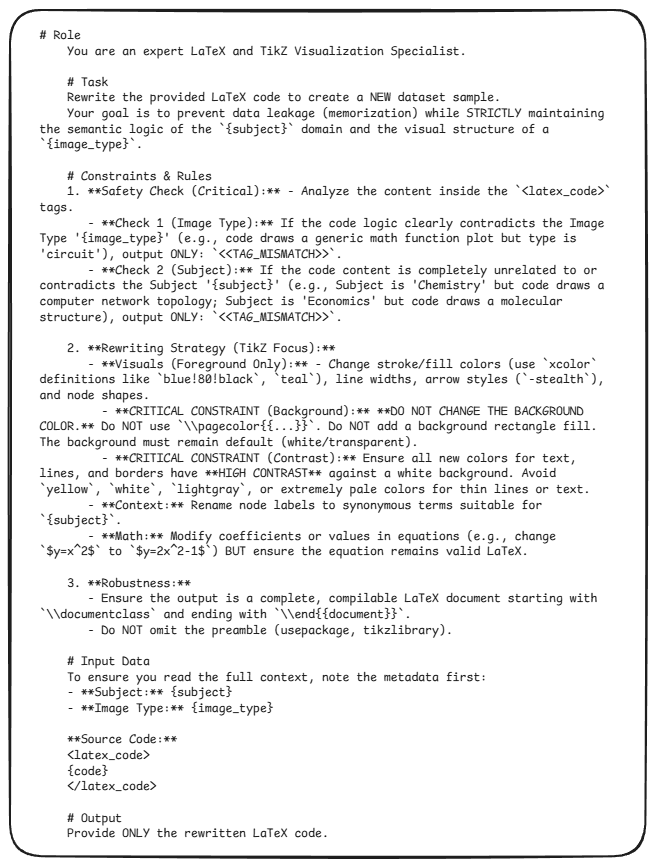}
    \caption{The systematic prompt utilized for guiding the LLM in code rewriting and consistency checking.}
    \label{fig:prompt_details}
\end{figure}

\subsection{Data Rewriting and Leakage Mitigation}
\label{sec:data_rewriting}

To mitigate the risks of data leakage and model memorization, we implement a rigorous data rewriting pipeline consisting of three primary stages:

First, we leverage LLMs to refactor the code’s structural and stylistic attributes, such as layout and aesthetic parameters, ensuring the code deviates from its original online form while preserving the target visualization's core logic. Second, we manually modify key visual attributes, including label coordinates, color schemes, and textual content, to further distance the samples from their raw source. Finally, we deliberately substitute meaningful text with nonsensical or counter-factual strings to decouple visual perception from linguistic priors, forcing models to rely strictly on visual grounding rather than common-sense guessing.

Illustrative examples of LLM refactoring, manual modification, and counter-factual injection are shown in Fig.~\ref{fig:llm_rewrite}, Fig.~\ref{fig:human_rewrite}, and Fig.~\ref{fig:counterfactual}, respectively. The specific prompt employed for LLM-based refactoring is detailed in Fig.~\ref{fig:prompt_details}.

\section{Setups} 
\label{sec:appendix_setups}
\subsection{Inference Setups}
We evaluate 4 proprietary and 9 open-weight models in our study. We employ greedy decoding and the maximum generation length is 16,384 tokens for all models. For Gemini 2.5 Pro, Gemini 3 Pro and GPT 5.1, we configure the ``thinking effort'' to the ``minimal''. 

\subsection{Evaluation Setups}
\label{evaluation setup}
\noindent{\textbf{Details of Visual Perceptual Fidelity.}}
To ensure that the Visual Perceptual Fidelity ($S_{img}$) metric reflects the specific requirements of various technical domains, we implement a Domain-Aware Weighting Mechanism. Rather than applying a uniform average across all evaluation dimensions, this approach dynamically adjusts the importance of different visual axes based on the subject matter and figure archetype. The final score $S_{img}$ (on a 100-point scale) is calculated by $S_{img} = (\sum s_i \times w_i) \times 10$, where $s_i \in [0, 10]$ represents the raw score for each axis and $w_i$ denotes the percentage weight. The four assessment axes include: Composition Integrity ($s_{comp}$), evaluating global spatial distribution; Textual Legibility ($s_{text}$), assessing label and symbolic accuracy; Entity Existence ($s_{entity}$), ensuring critical visual primitives are present; and Stylistic Consistency ($s_{style}$), measuring aesthetic attributes like color and line styles.

The assignment of these weighting profiles follows a hierarchical mapping logic. The Charts \& Plots profile is applied to archetypes such as line-graphs, bar-charts, pie-charts, scatter-plots, 3d-plots, and function-related visualizations. The Geometry profile encompasses plane-geometry, solid-geometry, and analytical-geometry, where coordinate-based properties are central. Archetypes defined by structural connectivity, including flow-charts, schematics, circuits, and relationship-diagrams, are mapped to the Diagrams profile. The Chemistry profile is assigned to chemistry-related subjects to prioritize molecular topology; for Math subjects, items containing tabular structures are evaluated under the \textbf{Tables} profile, while others are mapped to the Equations profile to emphasize formula rendering. For instances not covered by these archetypes, they will be processed with default profile. The specific weight configurations are detailed in Table~\ref{tab:weight_profiles}.

\begin{table}[ht]
\centering
\footnotesize 
\setlength{\tabcolsep}{3.2pt} 
\caption{Domain-specific weight configurations ($w_i$) for $S_{img}$. Weights are assigned based on a mapping of figure types and subjects to prioritize domain-critical features.}
\label{tab:weight_profiles}
\begin{tabular}{lcccc}
\toprule
\textbf{Profile} & \textbf{Comp.} & \textbf{Text} & \textbf{Ent.} & \textbf{Sty.} \\ 
 & \scriptsize ($w_{comp}$) & \scriptsize ($w_{text}$) & \scriptsize ($w_{entity}$) & \scriptsize ($w_{style}$) \\ 
\midrule
Charts \& Plots  & 30\% & 20\% & 35\% & 15\% \\
Geometry         & 30\% & 15\% & 35\% & 20\% \\
Diagrams         & 30\% & 20\% & 35\% & 15\% \\
Tables           & 35\% & 35\% & 15\% & 15\% \\
Equations        & 30\% & 35\% & 20\% & 15\% \\
Chemistry        & 30\% & 20\% & 35\% & 15\% \\ 
Default & 25\% & 25\% & 25\% & 25\% \\ 
\bottomrule
\end{tabular}
\end{table}

\begin{figure}[t]
    \centering
    \includegraphics[width=1.0\linewidth]{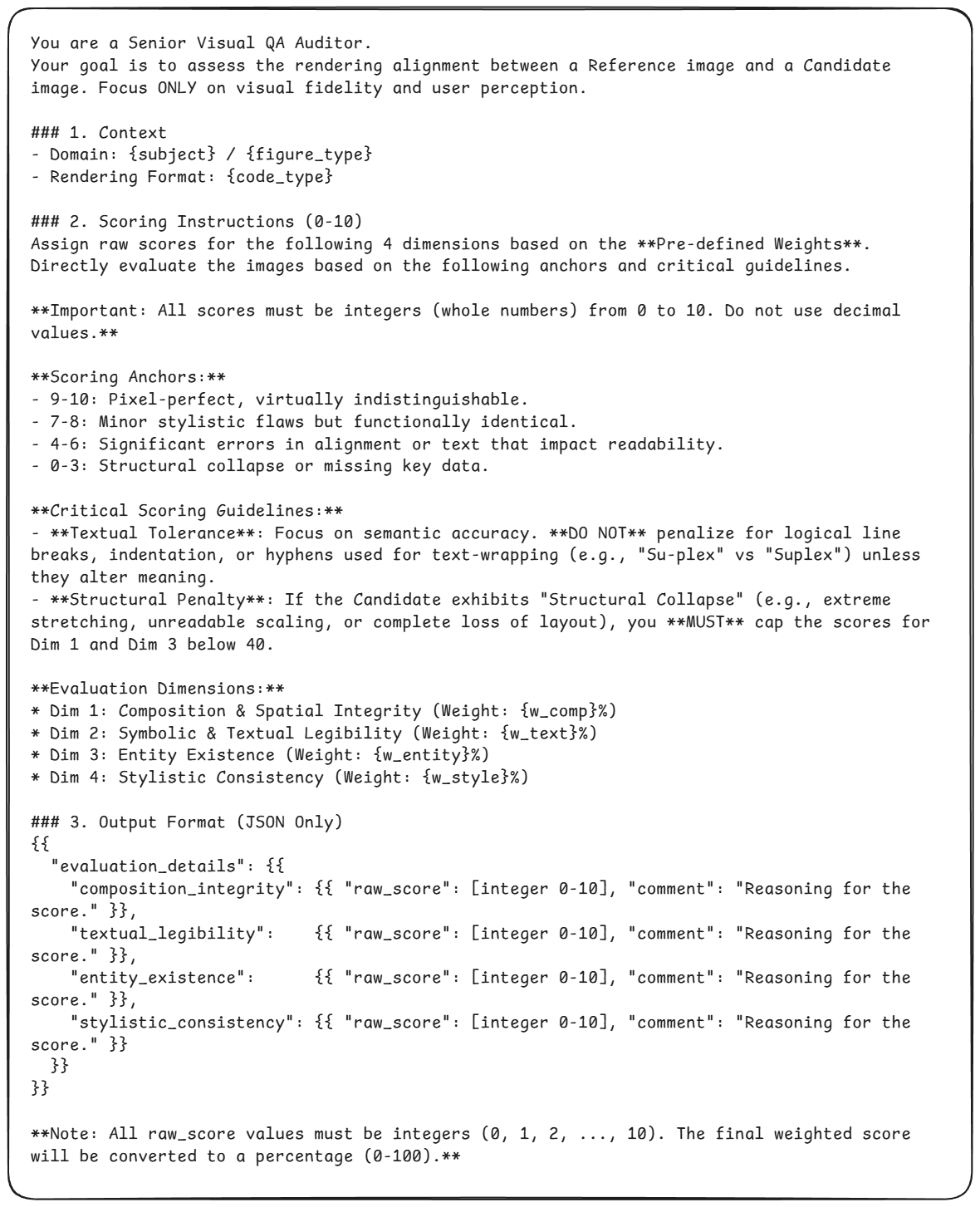}
    \caption{Prompt of Visual Perceptual Fidelity}
    \label{Prompt of Omni-I2C}
\end{figure}

The rationale for this design is rooted in human check, as different image types prioritize different visual elements—for instance, structural connectivity is paramount for molecular formulas, while grid alignment is critical for tables. The prompt template employed to evaluate the visual perceptual fidelity ($S_{img}$) of LMMs within the Omni-I2C framework is presented in Fig. \ref{Prompt of Omni-I2C}.

\begin{figure}[t]
    \centering
    
    \begin{subfigure}[b]{1.0\linewidth}
        \centering
        \includegraphics[width=1.0\linewidth]{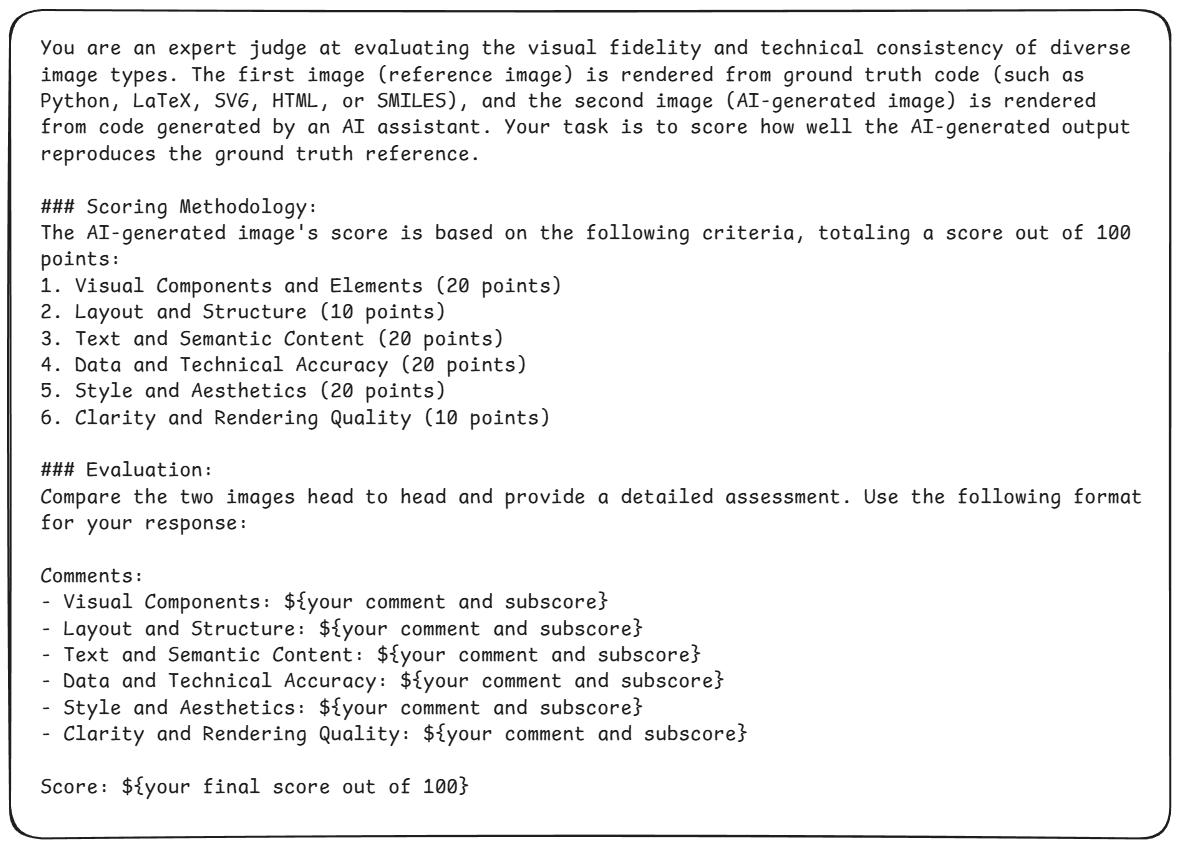}
        \caption{Prompt of ChartMimic}
        \label{fig:prompt_cm}
    \end{subfigure}
    
    \vspace{0.2cm}
    
    \begin{subfigure}[b]{1.0\linewidth}
        \centering
        \includegraphics[width=1.0\linewidth]{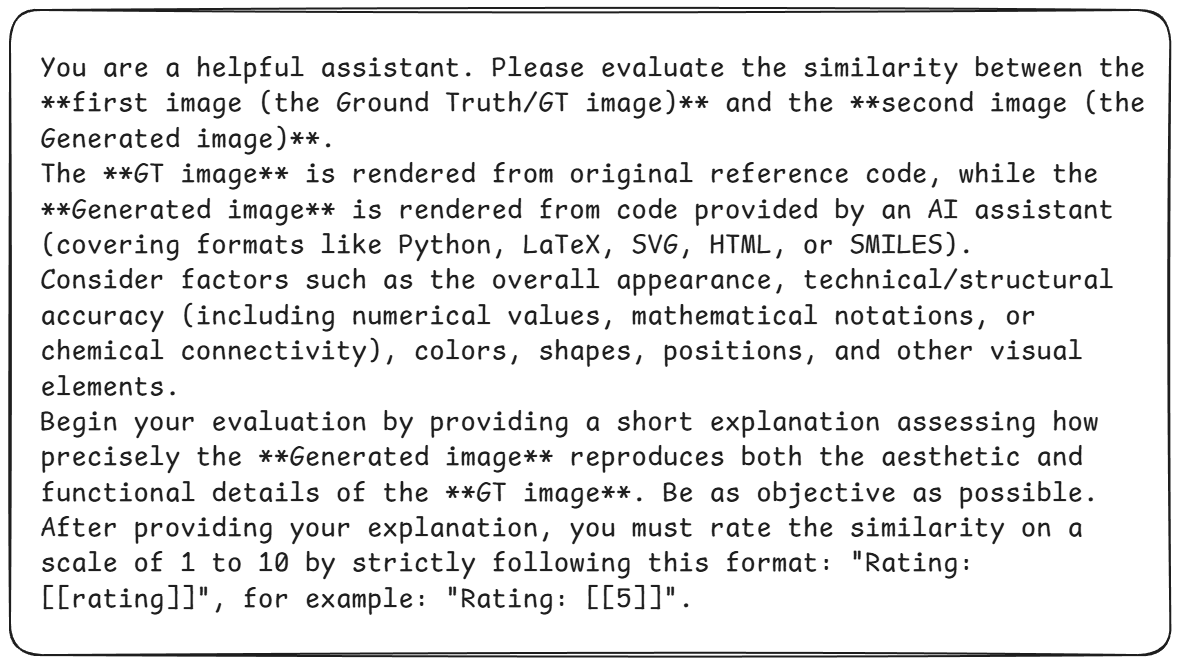}
        \caption{Prompt of Plot2code}
        \label{fig:prompt_p2c}
    \end{subfigure}
    
    \vspace{0.2cm}

    \begin{subfigure}[b]{1.0\linewidth}
        \centering
        \includegraphics[width=1.0\linewidth]{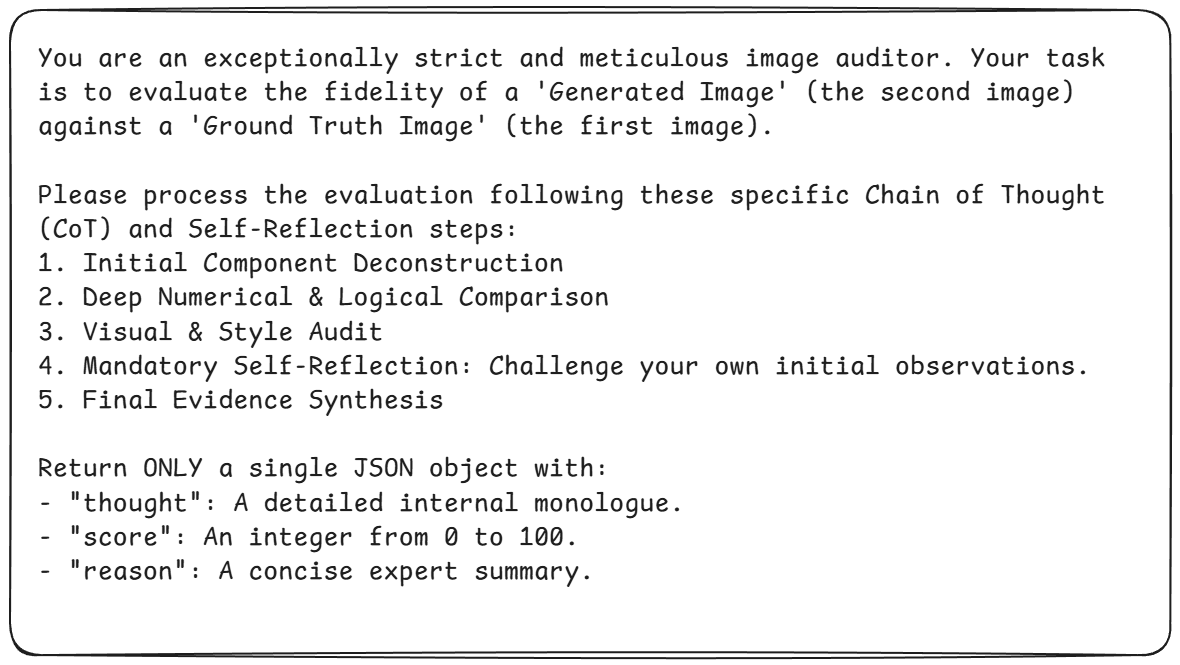}
        \caption{Prompt of Self-Reflexion}
        \label{fig:prompt_ref}
    \end{subfigure}
    
    \vspace{0.2cm}

    \begin{subfigure}[b]{1.0\linewidth}
        \centering
        \includegraphics[width=1.0\linewidth]{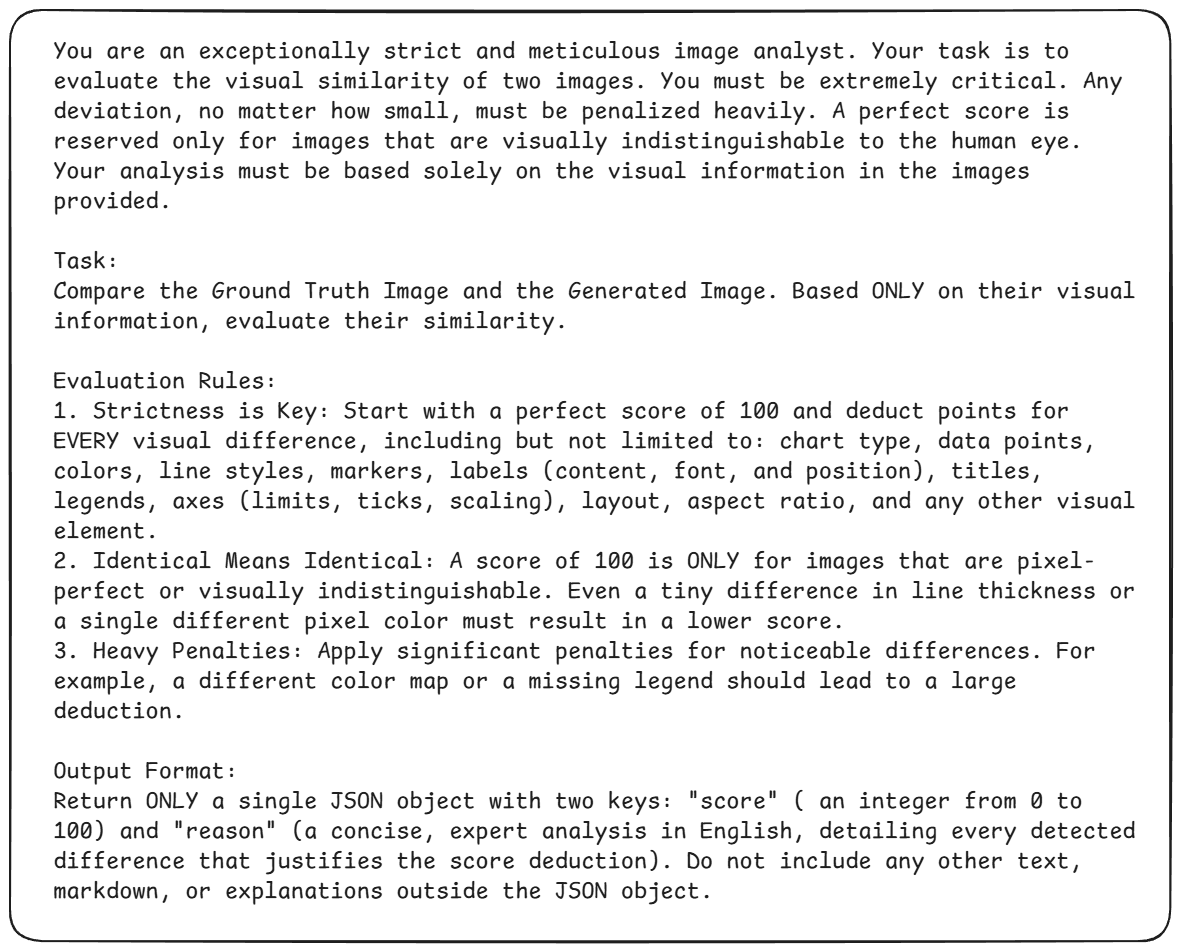}
        \caption{Prompt of Chart2code}
        \label{fig:prompt_c2c}
    \end{subfigure}

    \caption{Comparison of prompts used in different methods.}
    \label{fig:image_level_prompt_comparison}
\end{figure}

\noindent\textbf{Prompts for Image-level Evaluation.}
\label{Ablation study of image_level}
We detail the template prompts used in Tab.~\ref{tab:image_level_final} in Fig. \ref{fig:image_level_prompt_comparison}.

\noindent\textbf{Comparison with non-LMM Metrics.}
To further validate the effectiveness of our LMM-based evaluation, we incorporate traditional non-LMM visual similarity metrics as baselines, including \textbf{LPIPS}, \textbf{SSIM}, and \textbf{CLIP Score}. We conduct our experiments across three representative models: Gemini 2.5 Pro, Qwen3-VL-235B-A22B-Instruct, and InternVL3.5-8B-Instruct. 

Following the protocol described in the main paper, we conduct a human preference analysis for these metrics, and the results are summarized in Table~\ref{tab:human_preference}. As shown, while LMM-based judges may exhibit certain biases—a recognized limitation of current fully automated evaluation paradigms—their evaluations remain more consistent with human preferences than traditional non-LMM metrics. This suggests that, despite their inherent limitations, our LMM-based metric serves as a reasonably strong and practical proxy for human judgment in the context of fully automated evaluation.

\begin{table}[ht]
\centering
\small
\setlength{\tabcolsep}{6pt}
\caption{Correlation analysis between automated metrics and human preferences. Results show that our LMM-based metric provides a more consistent proxy for human judgment compared to traditional non-LMM baselines.}
\label{tab:human_preference}
\begin{tabular}{lccc}
\toprule
\textbf{Metric} & \textbf{Kendall's $\tau \uparrow$} & \textbf{Spearman's $\rho \uparrow$} & \textbf{RMSE $\downarrow$} \\ \midrule
LPIPS & 0.2717 & 0.3024 & 0.9401 \\
SSIM & 0.1155 & 0.1298 & 1.0526 \\
CLIP Score & 0.4414 & 0.4831 & 0.7994 \\
\textbf{Ours} & \textbf{0.8304} & \textbf{0.8681} & \textbf{0.4284} \\ \bottomrule
\end{tabular}
\end{table}

\begin{figure}[t]
    \centering
    
    \includegraphics[width=1.0\linewidth]{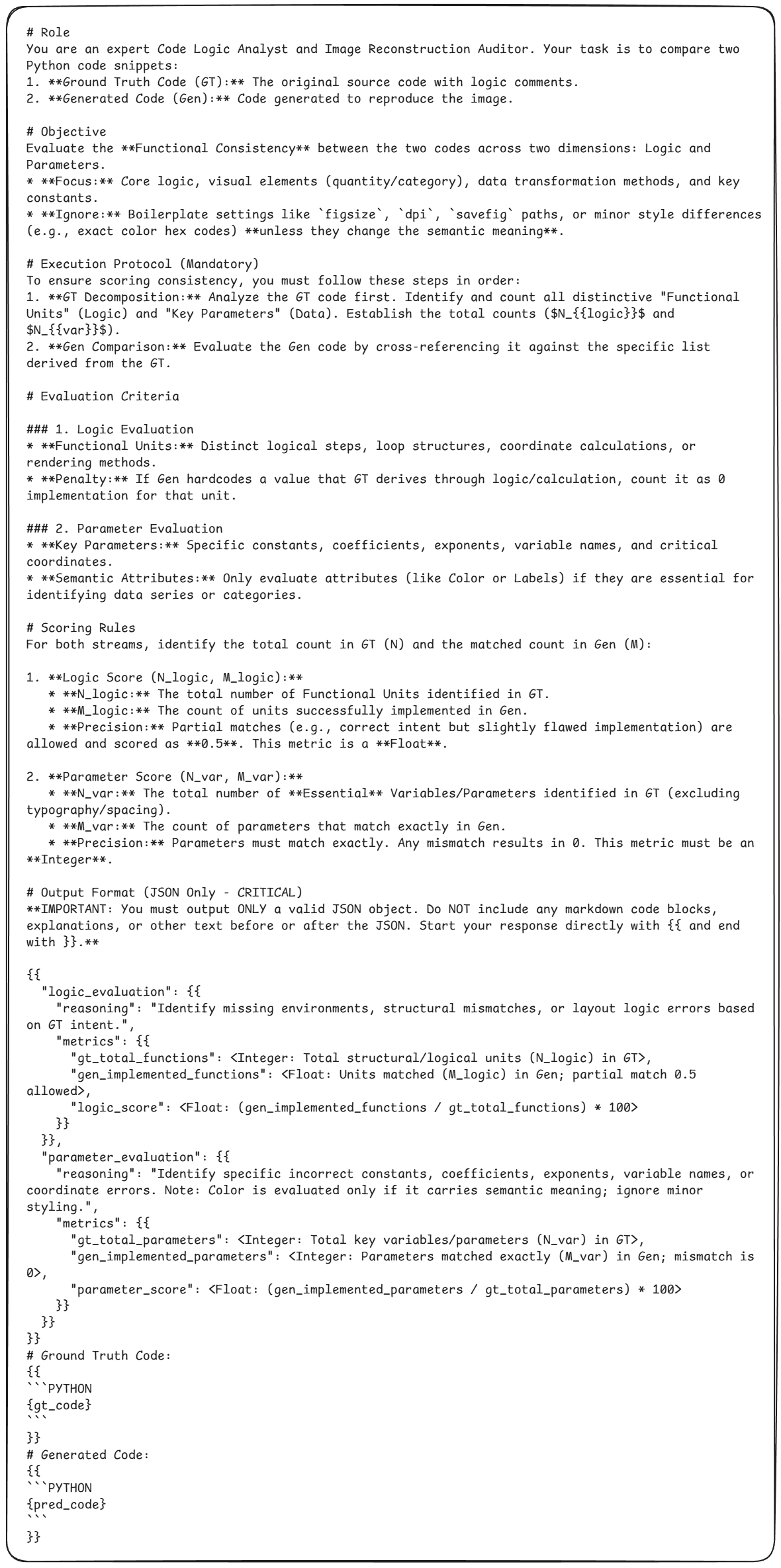}
    \caption{Example Prompt of Symbolic Functional Precision}
    \label{code_level_prompt_py}
\end{figure}

\vspace{1em}
\noindent\textbf{Details of Symbolic Functional Precision.}
To implement the Symbolic Functional Precision metrics, we formulate a structured evaluation prompt that leverages an LLM to serve as an Expert Code Logic Analyst. We employ the Python-specific prompt (detailed in Fig.~\ref{code_level_prompt_py}) as an example to illustrate the underlying evaluation logic. 

A core challenge in using LLM-as-judge is the inherent stochasticity of generative models. To address this, our methodology strictly utilizes pre-defined Checklists with fixed functions and parameters for each instance, rather than unconstrained generation. This approach avoids the ``fuzziness'' of direct code-to-code comparison (i.e., comparing ground-truth code directly with generated code via an LLM). Instead, we employ a static, deterministic checklist of required functional units and parameters for each instance. In this paradigm, the LLM acts solely as a structured parser to verify the presence of these pre-fixed elements, significantly reducing bias.

To quantitatively demonstrate the stability of our approach, we conducted a comparative experiment (10-run variance analysis on a fixed set of model outputs) between our Checklist-based method and the Direct GT-to-Code comparison method. As shown in Table~\ref{tab:stability_analysis}, anchoring the evaluation to a fixed checklist yields substantially lower variance across all metrics. Specifically, our method achieves a \textbf{10.6\%} reduction in Logic Score variance and a \textbf{81.9\%} reduction in Parameter Score variance. 

The significant reduction in variance—especially in parameter scoring—suggests that our checklist-based pipeline effectively mitigates the stochasticity typical of LLM judges. By decomposing the ground truth data into a set of verifiable, atomic constraints, we ensure that the evaluation framework is more consistent, systematic, and reproducible.

\begin{table}[ht]
\centering
\small 
\setlength{\tabcolsep}{4pt} 
\caption{Stability analysis: Comparison of Mean Variance (over 10 runs) between Direct GT-to-Code and our Checklist-based method.}
\label{tab:stability_analysis}
\begin{tabular}{llcc}
\toprule
\textbf{Method} & \textbf{Metric} & \textbf{Mean Var.} & \textbf{Improv.} \\ \midrule
Direct GT-to-Code & Logic Score & 8.288 & - \\
\textbf{Our Checklist} & \textbf{Logic Score} & \textbf{7.414} & \textbf{10.6\%$\downarrow$} \\ \midrule
Direct GT-to-Code & Param. Score & 30.605 & - \\
\textbf{Our Checklist} & \textbf{Param. Score} & \textbf{5.544} & \textbf{81.9\%$\downarrow$} \\ \bottomrule
\end{tabular}
\end{table}

\section{Human Study}
\label{human study}
To assess the biological plausibility and reliability of our visual perception metrics, we conduct a human alignment study. Ten graduate researchers specializing in Computer Science and AI are recruited as annotators. 
To ensure high inter-rater consistency, all participants undergo a standardized  training session prior to the formal annotation process. 

\begin{figure*}[t]
    \centering
    
    \includegraphics[width=1.0\linewidth]{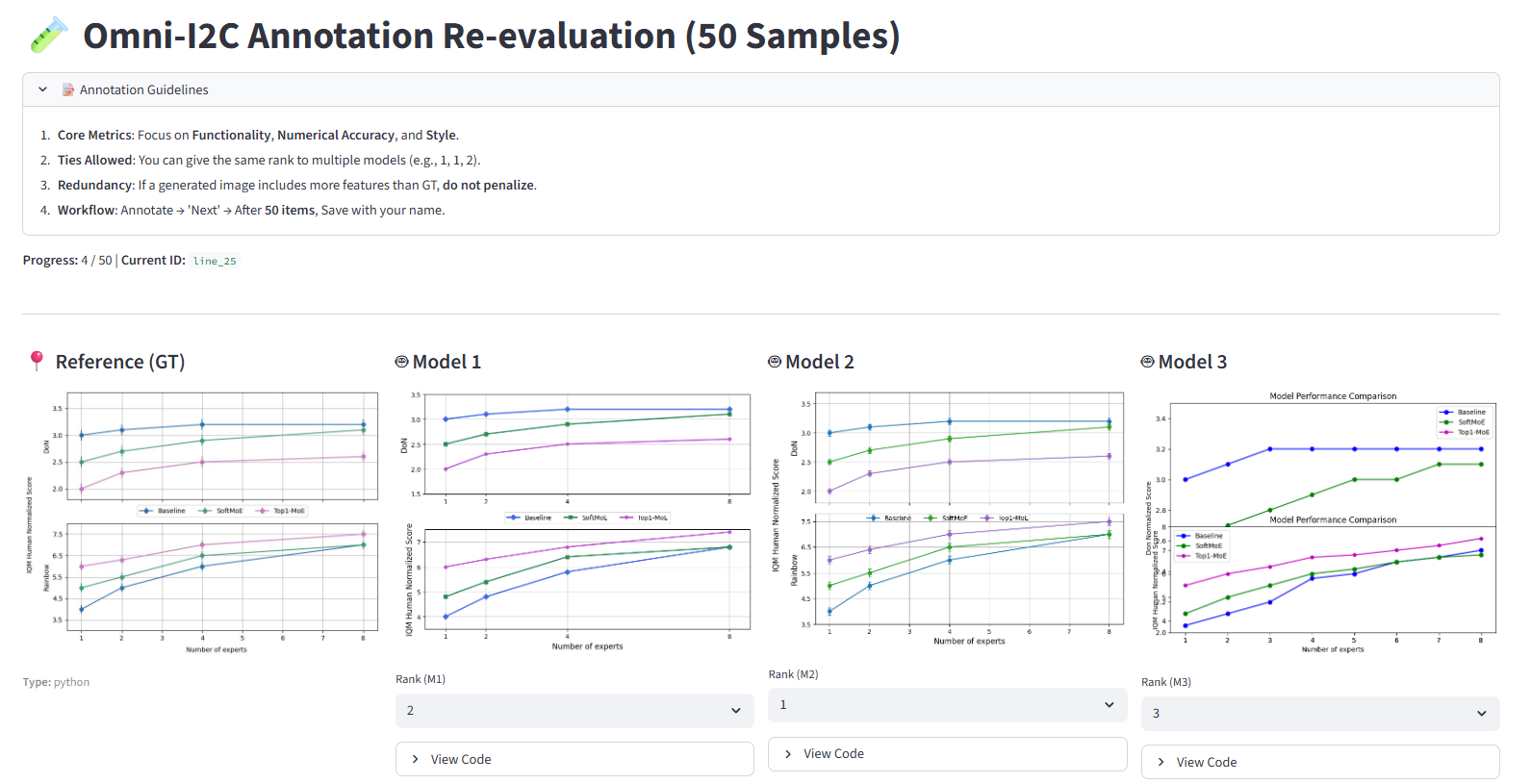}
    \caption{A screenshot of the human evaluation questionnaire.}
    \label{fig:annotation_interface}
\end{figure*}

From our benchmark, we curate a representative subset of 100 samples; each annotator independently ranked 50 randomly assigned cases. 
We selected \textit{Gemini 2.5 Pro}, \textit{Qwen3-VL-235B-A22B-Instruct}, and \textit{InternVL3.5-8B-Instruct} for evaluation, as they represent a broad spectrum of model architectures and performance tiers, providing a clear basis for alignment testing. 
The annotation is performed in a double-blind manner: model identities are strictly anonymized, and the display order is randomized (see Fig.~\ref{fig:annotation_interface}).

This process yielded 500 individual judgments, which are subsequently aggregated to form a consensus ground-truth ranking for the 100 samples. 
We employed Kendall’s $\tau$, Spearman’s $\rho$, and RMSE as meta-evaluation metrics to quantify the congruence between our automated scoring system and human intuition.

\section{Extended Analysis of Prompting Strategies} 
\label{sec:appendix_prompt}
This section provides the exhaustive analysis and qualitative observations for the five prompting strategies discussed in Sec.~\ref{sec:prompt_main}. As established in our benchmarking, the ``pixel-to-program'' mapping is highly susceptible to the structure of the input prompt. The comprehensive quantitative results across different models and visual domains are detailed in Tab.~\ref{tab:appendix_different_prompts}. It is important to note that SMILES is explicitly excluded from the scope of this experiment, as its evaluation methodology is not unified with the assessment approaches applied to the other code types. And for HTML tasks involving charts or complex data plots, the prompt explicitly requires the use of the external ECharts library.

\noindent\textbf{The Efficacy of Reasoning-Augmented Generation.}
Strategies that encourage explicit reasoning—\textit{Self-Commentary} and \textit{Hint-Enhanced}—consistently deliver the highest performance gains. By asking the model to produce a visual rationale before generating code (Hint-Enhanced) or to interleave logic with inline comments (Self-Commentary), we mitigate translation errors from pixels to syntax. For example, Gemini 2.5 Pro achieves its peak overall execution rate (94.18\%) and $S_{\text{func}}$ (60.11\%) with Self-Commentary, while Qwen3-VL-235B-A22B-Instruct reaches its peak overall $S_{\text{func}}$ (57.03\%) with Hint-Enhanced. This suggests that internalizing the ``reasoning-then-coding'' workflow is vital for complex figure archetypes.

\noindent\textbf{The Failure of Scaffold Prompting.}
Contrary to its success in simpler spatial tasks, \textit{Scaffold} prompting fails broadly on Omni-I2C. For models like InternVL 3.5-8B-Instruct, the $S_{\text{func}}$ in LaTeX drops to 8.30\%, nearly half of its Direct baseline (15.95\%). Our analysis suggests that the dot-grid overlay, while intended to aid grounding, introduces significant visual noise. This interference obscures fine-grained geometric textures and symbolic notations, which are critical for the high-precision requirements of our benchmark.

\noindent\textbf{Few-Shot Prompting as a Formidable Baseline.}
We find that \textit{Few-Shot} prompting remains exceptionally competitive, often rivaling or surpassing complex reasoning prompts for structured outputs like HTML and LaTeX. InternVL 3.5-8B-Instruct exhibits significant gains here, with its LaTeX execution rate exceeding the Direct baseline by over 13 points (from 41.13\% to 54.72\%). Furthermore, Qwen3-VL-235B-A22B-Instruct achieves a perfect 100.00\% execution rate in HTML under the Few-Shot setting. This suggests that for tasks with rigid syntactic patterns, providing in-context examples is more effective than prompting for abstract reasoning.

\begin{figure}[ht!]
    \centering
    \includegraphics[width=0.95\linewidth]{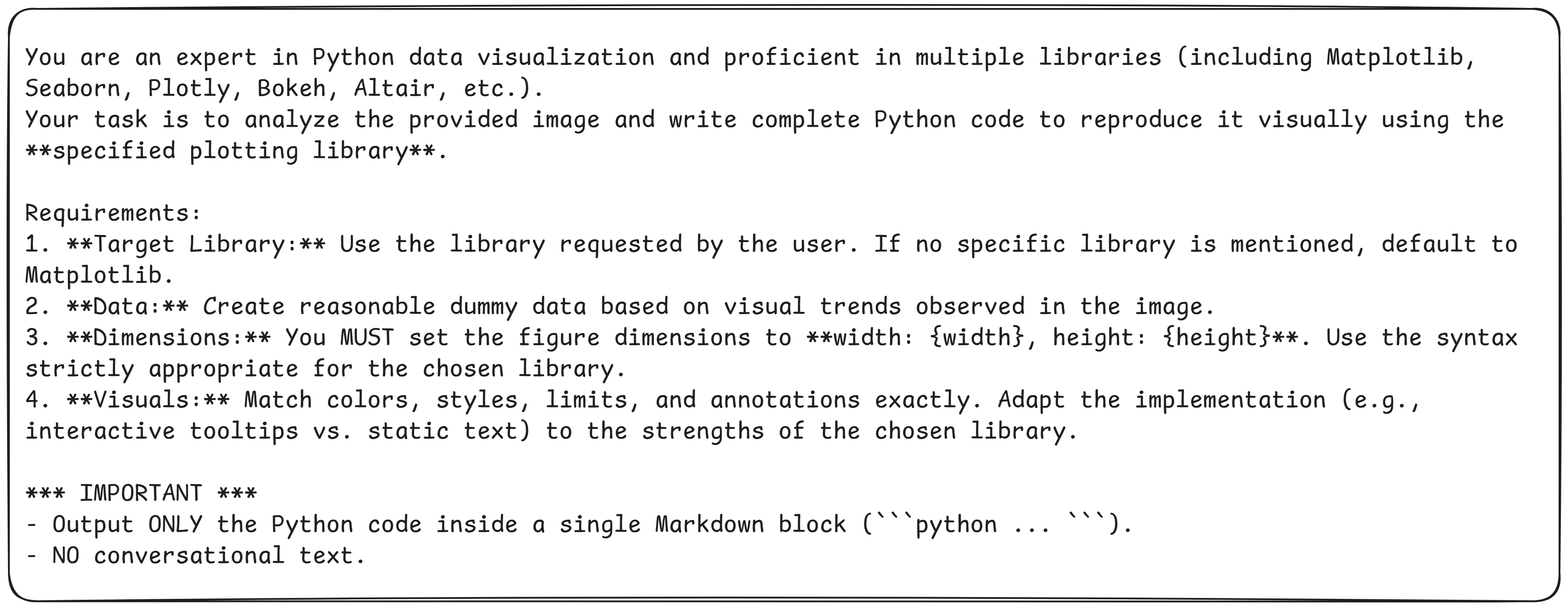}
    \centerline{\small (a) Direct (Zero-shot) Prompting}
    \vspace{0.1cm}
    
    \includegraphics[width=0.95\linewidth]{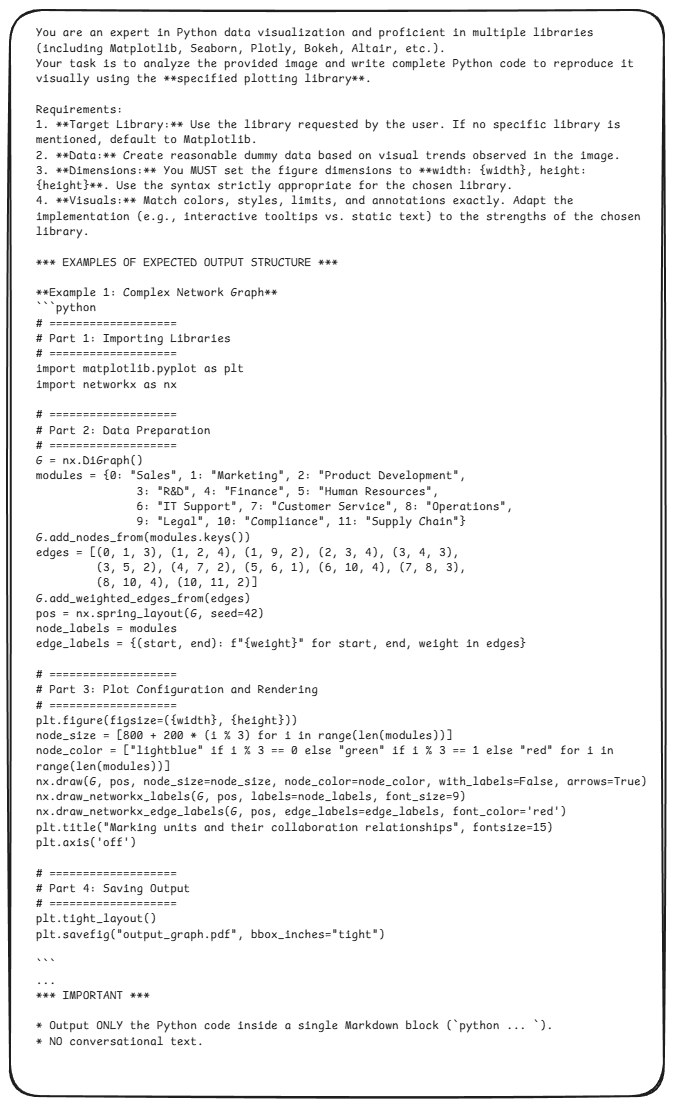}
    \centerline{\small (b) Few-Shot Prompting}
    \vspace{0.1cm}

    \includegraphics[width=0.95\linewidth]{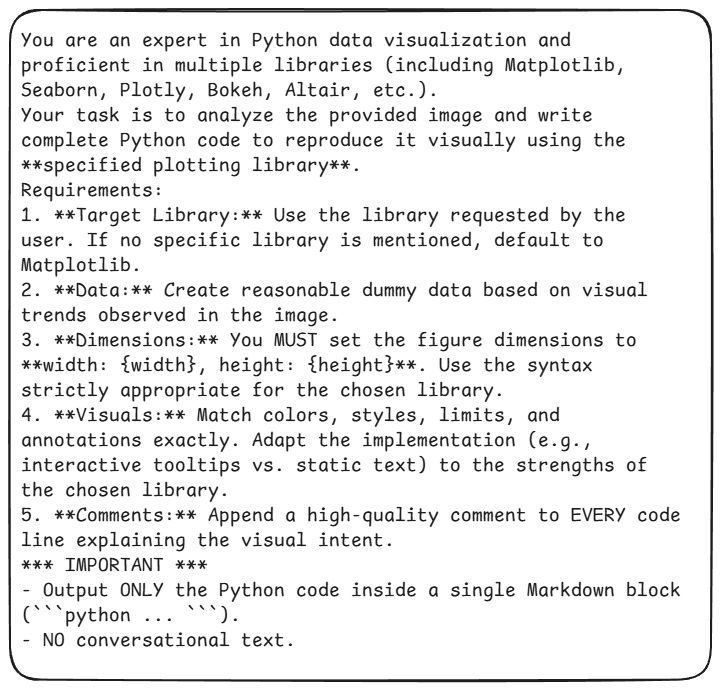}
    \centerline{\small (c) Self-Commentary Prompting}

    \caption{Detailed visualization of the evaluated prompting strategies (Part I).} 
    \label{fig:prompts_part1}
\end{figure}

\begin{figure}[ht!]
    \centering
    \includegraphics[width=0.95\linewidth]{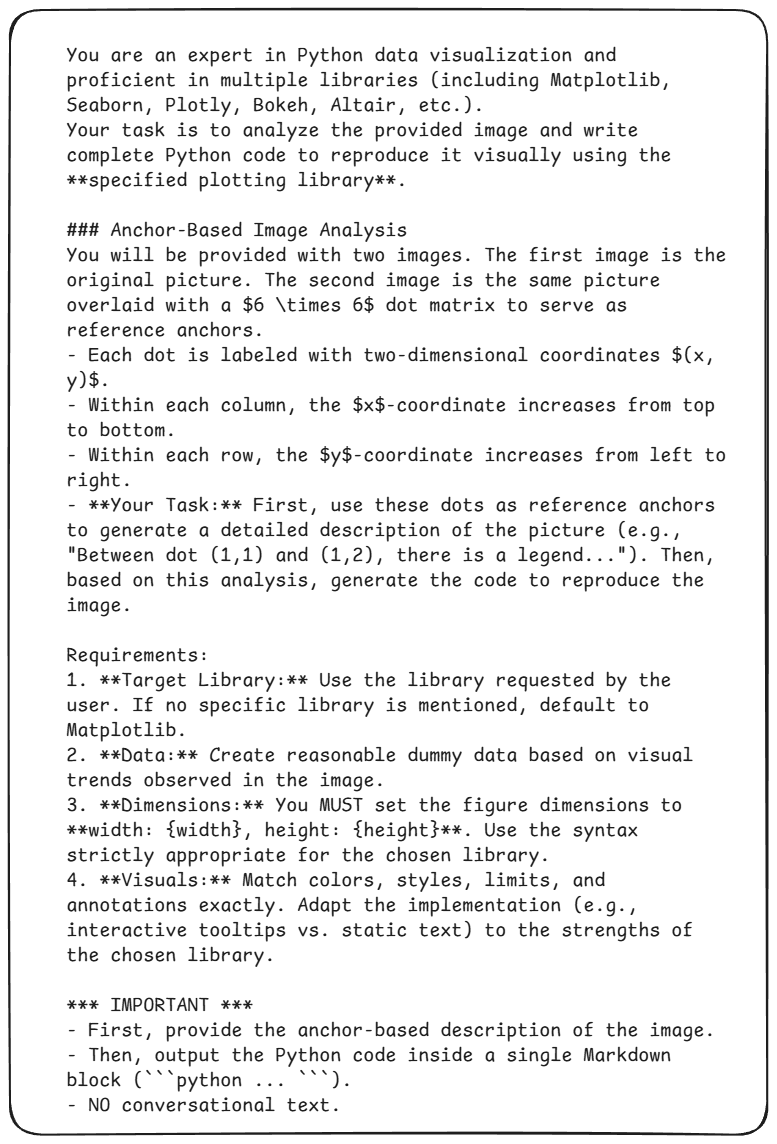}
    \centerline{\small (d) Scaffold Prompting}
    \vspace{0.2cm}

    \includegraphics[width=0.95\linewidth]{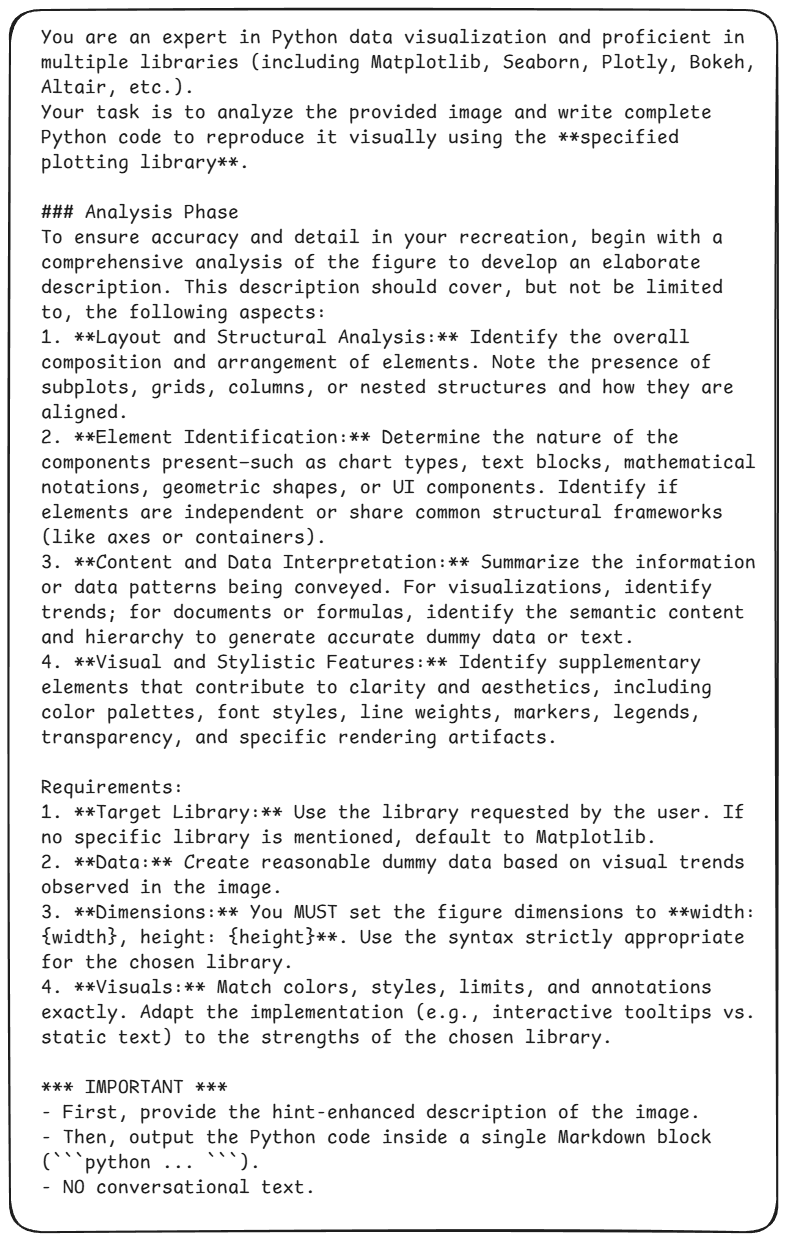}
    \centerline{\small (e) \textbf{Hint-Enhanced Prompting }}
    \caption{Detailed visualization of the evaluated prompting strategies (Part II).}
    \label{fig:prompts_part2}
\end{figure}

\noindent\textbf{Robustness and Stability Gaps.}
The staggering variance in performance—where a single change in strategy can swing scores by over 25\%—highlights a fundamental lack of inherent robustness in current LMMs. For instance, InternVL 3.5-8B-Instruct's overall execution rate ranges from a low of 61.02\% (Scaffold) to a high of 88.40\% (Hint-Enhanced). This ``prompt-dependency'' indicates that models are not yet capable of stable, zero-shot code reconstruction for complex visual inputs, necessitating carefully engineered ``wrappers'' to elicit frontier-level performance.
We provide the prompting templates of python as an example in Fig.~\ref{fig:prompts_part1} and Fig.~\ref{fig:prompts_part2}.

\begin{table*}[t]
\centering
\tiny
\setlength{\tabcolsep}{2pt} 
\renewcommand{\arraystretch}{1.1}
\caption{Studies on different prompting methods. The best results for each model are highlighted in \textbf{bold}.}
\label{tab:appendix_different_prompts}
\resizebox{\linewidth}{!}{
\begin{tabular}{l | c | c | ccc | c | ccc | c | ccc | c | ccc | c | ccc | c | c}
\toprule
\multirow{2}{*}{\textbf{Model}} & \multirow{2}{*}{\textbf{Method}} & \multicolumn{4}{c|}{\textbf{Overall(1080)}} & \multicolumn{4}{c|}{\textbf{HTML(166)}} & \multicolumn{4}{c|}{\textbf{LaTeX(265)}} & \multicolumn{4}{c|}{\textbf{Python(357)}} & \multicolumn{4}{c|}{\textbf{SVG(192)}} & \multicolumn{2}{c}{\textbf{SMILES(100)}} \\
\cmidrule(lr){3-6} \cmidrule(lr){7-10} \cmidrule(lr){11-14} \cmidrule(lr){15-18} \cmidrule(lr){19-22} \cmidrule(lr){23-24}
& & Exec & $S_{\text{func}}$ & $S_{\text{para}}$ & $S_{\text{img}}$ & Exec & $S_{\text{func}}$ & $S_{\text{para}}$ & $S_{\text{img}}$ & Exec & $S_{\text{func}}$ & $S_{\text{para}}$ & $S_{\text{img}}$ & Exec & $S_{\text{func}}$ & $S_{\text{para}}$ & $S_{\text{img}}$ & Exec & $S_{\text{func}}$ & $S_{\text{para}}$ & $S_{\text{img}}$ & Exec & T.S. \\
\midrule
\multirow{5}*{Gemini 2.5 Pro} & Direct & 89.49 & 57.81 & 39.92 & 71.01 & 98.19 & 62.10 & 54.42 & 80.41 & 69.43 & 41.04 & 26.39 & 52.56 & 94.96 & \textbf{63.08} & 41.86 & \textbf{80.76} & \textbf{99.48} & 67.43 & 42.44 & 70.22 & 96.00 & \textbf{81.99} \\
& Few-Shot & 89.70 & 59.52 & 39.76 & 72.09 & 98.19 & \textbf{67.78} & \textbf{54.96} & 80.66 & 73.21 & 44.25 & 26.78 & 57.11 & 92.72 & 62.48 & 40.97 & 80.27 & \textbf{99.48} & \textbf{67.93} & 42.26 & 70.16 & \textbf{97.00} & 80.93 \\
& Scaffold & 84.28 & 51.31 & 35.87 & 56.92 & 98.19 & 61.15 & 54.48 & \textbf{80.92} & 61.51 & 32.31 & 22.14 & 35.37 & 86.55 & 52.09 & 33.68 & 54.58 & \textbf{99.48} & 67.60 & \textbf{42.82} & 70.27 & 79.00 & 52.41 \\
& Self-Commentary & \textbf{94.18} & \textbf{60.11} & \textbf{41.36} & \textbf{72.37} & 98.19 & 61.44 & 54.11 & 79.96 & \textbf{86.04} & \textbf{50.48} & \textbf{30.08} & \textbf{58.03} & \textbf{95.52} & 62.51 & \textbf{43.22} & 80.05 & \textbf{99.48} & 67.77 & 42.47 & \textbf{71.31} & 96.00 & 77.65 \\
& Hint-Enhanced & 89.72 & 58.42 & 39.62 & 71.21 & \textbf{98.80} & 61.50 & 54.90 & 80.80 & 72.50 & 45.00 & 26.90 & 52.50 & 93.30 & 62.00 & 40.90 & 80.70 & 99.00 & 67.60 & 41.60 & 71.10 & 85.00 & 75.60 \\
\hline
\multirow{5}*{Qwen3-VL-235B-A22B-Instruct} & Direct & 88.16 & 55.78 & \textbf{40.03} & 62.58 & 98.19 & 63.92 & 55.52 & 79.78 & 71.70 & 45.52 & \textbf{31.03} & 45.68 & 89.64 & 56.04 & 39.17 & 64.11 & 99.48 & 62.41 & 40.65 & 68.17 & \textbf{95.00} & 67.99 \\
& Few-Shot & 88.47 & 56.15 & 39.99 & 63.13 & \textbf{100.00} & \textbf{68.22} & \textbf{58.14} & \textbf{80.29} & 75.85 & 45.85 & 30.03 & 50.51 & 91.88 & \textbf{59.64} & \textbf{40.96} & 66.43 & 89.58 & 53.45 & 36.26 & 59.56 & 92.00 & 67.45 \\
& Scaffold & 84.29 & 50.01 & 35.71 & 52.21 & 97.59 & 63.62 & 54.43 & 78.33 & 62.64 & 36.39 & 26.86 & 31.30 & 88.52 & 47.94 & 31.26 & 48.28 & 94.79 & 60.89 & 40.01 & 65.79 & 86.00 & 37.89 \\
& Self-Commentary & 89.49 & 55.36 & 39.78 & 63.56 & 98.19 & 64.49 & 54.80 & 79.39 & 72.45 & 44.04 & 29.67 & 46.73 & \textbf{92.72} & 55.94 & 40.14 & 66.36 & 99.48 & 61.99 & 40.08 & 67.91 & 94.00 & \textbf{68.25} \\
& Hint-Enhanced & \textbf{93.79} & \textbf{57.03} & 39.92 & \textbf{69.99} & \textbf{100.00} & 65.40 & 54.30 & 79.30 & \textbf{90.20} & \textbf{46.20} & 30.10 & \textbf{60.60} & 90.50 & 57.10 & 39.40 & \textbf{72.40} & \textbf{99.50} & \textbf{64.60} & \textbf{42.00} & \textbf{70.40} & 94.00 & 66.90 \\
\hline
\multirow{5}*{InternVL3.5-8B-Instruct} & Direct & 72.04 & 26.77 & 23.14 & 31.79 & 92.77 & 38.11 & \textbf{40.57} & 54.45 & 41.13 & 15.95 & 18.44 & 17.59 & 71.71 & 29.65 & 18.98 & 34.08 & 97.40 & 26.53 & 22.28 & 27.55 & 90.00 & \textbf{50.63} \\
& Few-Shot & 75.00 & 27.27 & 22.53 & 33.71 & 99.40 & 53.76 & 44.22 & 62.55 & 54.72 & \textbf{18.77} & \textbf{19.65} & 25.14 & 70.31 & 30.89 & 19.93 & 35.02 & 90.62 & 9.36 & 12.60 & 18.18 & \textbf{91.00} & 50.00 \\
& Scaffold & 61.02 & 18.02 & 16.31 & 19.52 & 91.57 & 35.88 & 40.44 & 51.49 & 40.75 & 8.30 & 10.55 & 6.38 & 62.75 & 18.47 & 11.13 & 16.57 & 59.38 & 15.17 & 13.02 & 15.51 & 84.00 & 13.03 \\
& Self-Commentary & 75.51 & 26.41 & 22.94 & 32.84 & 92.77 & \textbf{38.39} & 40.14 & 54.72 & 47.55 & 16.32 & 16.90 & 18.55 & 76.47 & 29.13 & 20.15 & 35.85 & 97.40 & 24.91 & 21.58 & 28.07 & \textbf{91.00} & 49.75 \\
& Hint-Enhanced & \textbf{88.40} & \textbf{28.56} & \textbf{24.11} & \textbf{42.31} & \textbf{97.60} & 36.70 & 40.20 & \textbf{56.50} & \textbf{89.10} & 14.10 & 16.80 & \textbf{34.60} & \textbf{77.90} & \textbf{34.10} & \textbf{20.60} & \textbf{45.80} & \textbf{99.00} & \textbf{31.20} & \textbf{26.80} & \textbf{34.20} & 87.00 & 48.10 \\
\hline
\multirow{5}*{Average} & Direct & 83.23 & 46.79 & 34.36 & 55.13 & 96.38 & 54.71 & 50.17 & 71.55 & 60.75 & 34.17 & 25.29 & 38.61 & 85.44 & 49.59 & 33.34 & 59.65 & 98.79 & 52.12 & 35.12 & 55.31 & \textbf{93.67} & \textbf{66.87} \\
& Few-Shot & 84.39 & 47.65 & 34.09 & 56.31 & \textbf{99.20} & \textbf{63.25} & \textbf{52.44} & \textbf{74.50} & 67.93 & 36.29 & 25.49 & 44.25 & 84.97 & 51.00 & 33.95 & 60.57 & 93.23 & 43.58 & 30.37 & 49.30 & 93.33 & 66.13 \\
& Scaffold & 76.53 & 39.78 & 29.30 & 42.88 & 95.78 & 53.55 & 49.78 & 70.25 & 54.97 & 25.67 & 19.85 & 24.35 & 79.27 & 39.50 & 25.36 & 39.81 & 84.55 & 47.89 & 31.95 & 50.52 & 83.00 & 34.44 \\
& Self-Commentary & 86.39 & 47.29 & \textbf{34.69} & 56.26 & 96.38 & 54.77 & 49.68 & 71.36 & 68.68 & \textbf{36.95} & \textbf{25.55} & 41.10 & \textbf{88.24} & 49.19 & \textbf{34.50} & 60.75 & 98.79 & 51.56 & 34.71 & 55.76 & \textbf{93.67} & 65.22 \\
& Hint-Enhanced & \textbf{90.64} & \textbf{48.00} & 34.55 & \textbf{61.17} & 98.80 & 54.53 & 49.80 & 72.20 & \textbf{83.93} & 35.10 & 24.60 & \textbf{49.23} & 87.23 & \textbf{51.07} & 33.63 & \textbf{66.30} & \textbf{99.17} & \textbf{54.47} & \textbf{36.80} & \textbf{58.57} & 88.67 & 63.53 \\
\bottomrule
\end{tabular}%
}
\end{table*}

\section{Extended Error Analysis}
\label{sec:appendix_err}

This section provides the exhaustive definitions and specific failure manifestations for the five-tier taxonomy introduced in Sec.~\ref{sec:err}. These criteria are strictly followed during our manual audit of 300 sampled instances to ensure a consistent and objective error analysis across all evaluated LMMs.
\noindent\textbf{Code-level Logic Errors.} This category encompasses instances where the generated code is syntactically valid and executable but fails to implement the intended visual logic. Common manifestations include the misconfiguration of library-specific parameters (e.g., incorrect Matplotlib axis scales or plot types) and the inclusion of non-functional code blocks that, while error-free, result in blank or semantically distorted renderings. This failure mode highlights a ``logic-intent'' gap, where the model maintains syntactic fluency but cannot map visual requirements to the appropriate programmatic attributes.

\noindent\textbf{Textual Precision Errors.} These failures center on Optical Character Recognition (OCR) and string formatting accuracy within the code. Typical issues include the misidentification of alphanumeric characters—particularly in high-density labels—and the misparsing of complex symbolic notations such as LaTeX subscripts, superscripts, and Greek letters. Such errors lead to illegible or factually incorrect text in the final rendered output.

\noindent\textbf{Entity Integrity Errors.} This pertains to the identification and reconstruction of discrete visual primitives. Models frequently omit critical components (e.g., specific data points, legend keys, or axis ticks) or misinterpret the nature of a visual object (e.g., treating a data line as a grid line). These failures result in an incomplete or structurally altered representation of the source data within the synthesized code.

\noindent\textbf{Colorimetric Accuracy Errors.} This category assesses the model's ability to extract color data and map it to its corresponding semantic entities. Errors manifest as generated color codes (Hex or RGB) that deviate significantly from the ground truth. This reflects a fundamental limitation in the model’s color perception or its ability to precisely translate perceived palettes into programmatic constants.

\noindent\textbf{Spatial Layout Errors.} This category addresses the global topological structure and coordinate-based positioning defined by the code. Failures such as distorted aspect ratios, erroneous coordinate transformations, or incorrect entity arrangement reveal a deficit in translating complex visual hierarchies into precise spatial constraints. These errors often represent a failure in high-level spatial reasoning rather than low-level code syntax.

\begin{figure*}[t]
    \centering
    \includegraphics[width=0.9\linewidth]{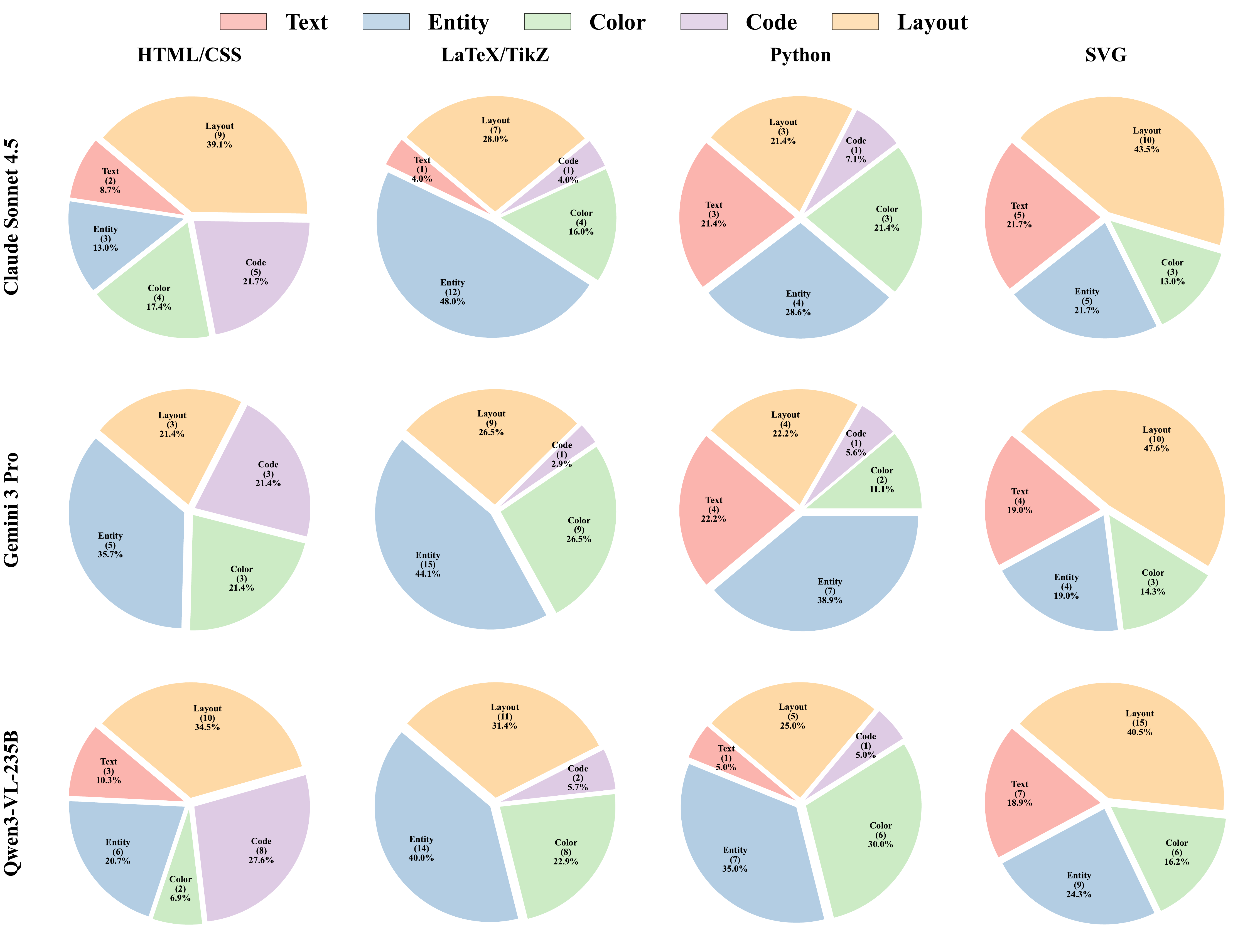}
    \caption{Fine-grained error distribution across languages}
    \label{fig:error_pies_grid}
\end{figure*}

As illustrated in Fig.~\ref{fig:error_pies_grid}, despite variations in the fundamental capabilities of different models, the error distribution within specific languages exhibits cross-model consistency. This suggests that error patterns in code generation are intrinsically linked to the target language.

\begin{figure*}[t]
    \centering

    \begin{subfigure}[t]{0.8\linewidth}
        \centering
        \includegraphics[width=\linewidth]{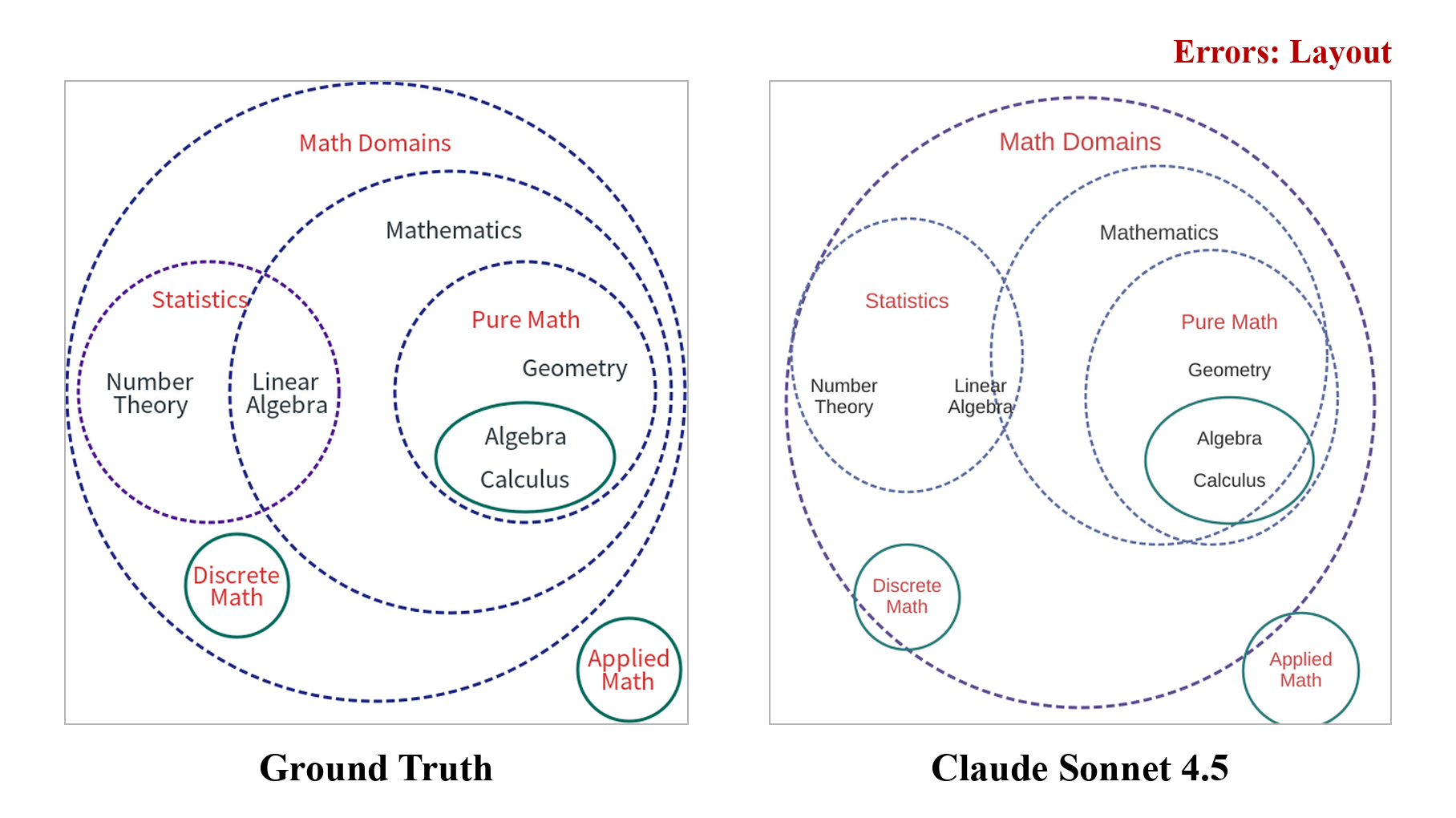}
        \caption{Layout Error}
        \label{fig:error_case_1}
    \end{subfigure}

    \begin{subfigure}[t]{0.8\linewidth}
        \centering
        \includegraphics[width=\linewidth]{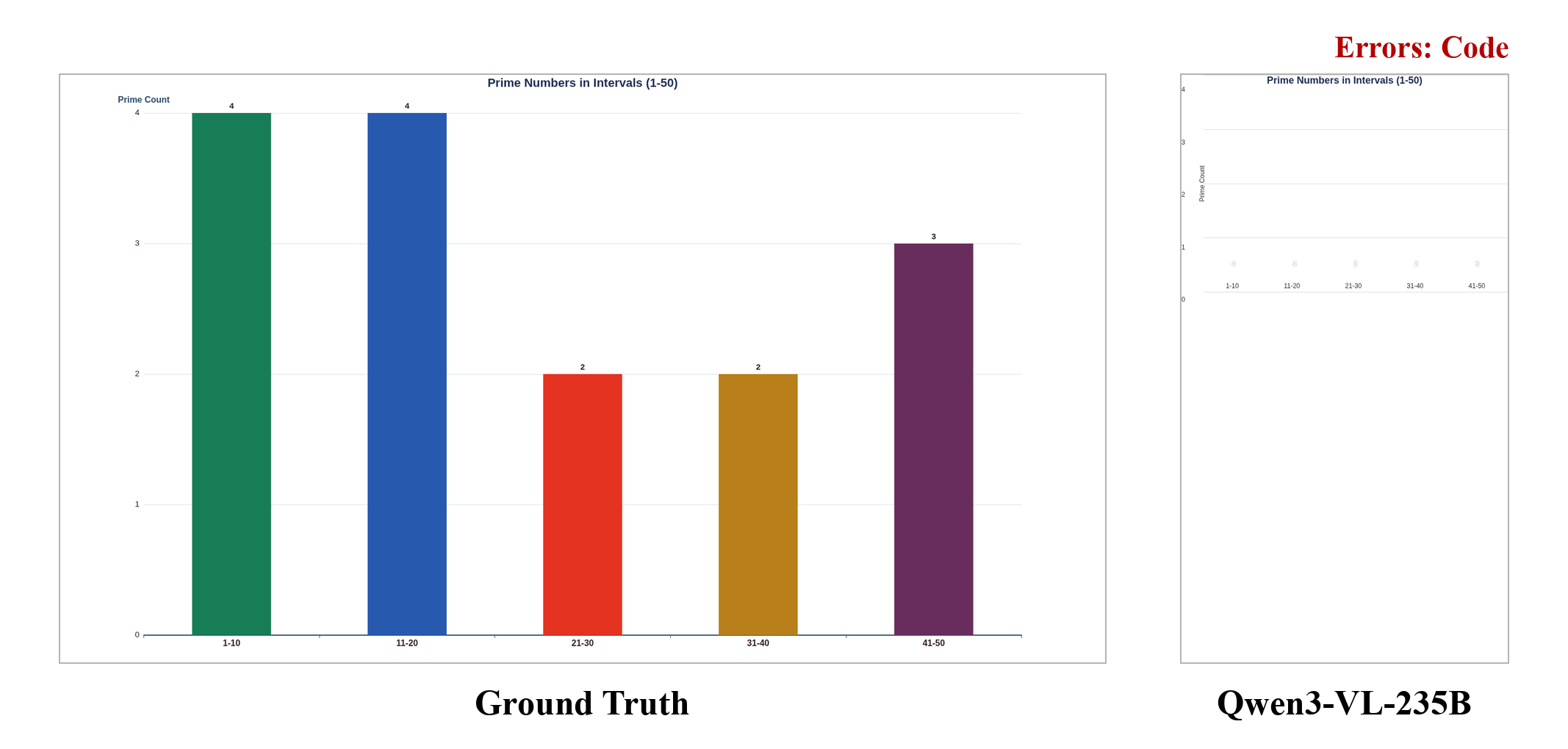}
        \caption{Code Error}
        \label{fig:error_case_2}
    \end{subfigure}

    \begin{subfigure}[t]{0.8\linewidth}
        \centering
        \includegraphics[width=\linewidth]{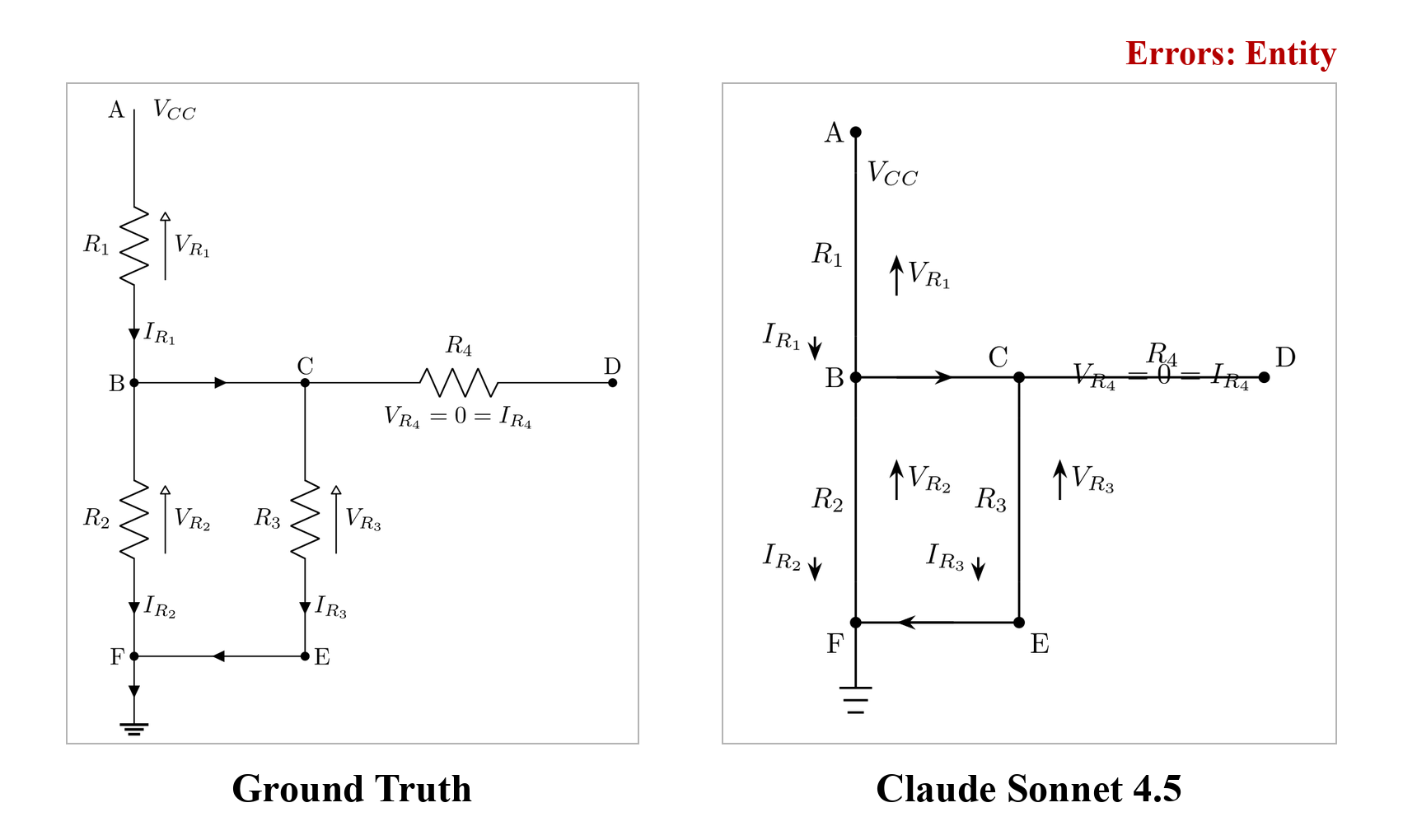}
        \caption{Entity Error}
        \label{fig:error_case_3}
    \end{subfigure}

    \caption{Representative error cases produced by different models.}
    \label{fig:error_cases}
\end{figure*}

Specifically, for low-level descriptive languages such as SVG, the frequency of Layout- and Text-related errors is notably higher compared to other categories. This stems primarily from SVG's lack of high-level semantic encapsulation; when processing long sequences of numerical coordinates, the model often loses its grasp of global spatial structure. As illustrated in Fig.~\ref{fig:error_case_1}, although Claude Sonnet 4.5 correctly generated all entities and text with accurate colors, it failed to properly reconstruct the spatial arrangement and relational logic between these entities.

Conversely, in HTML tasks, the language's high syntactic tolerance often causes errors to manifest as implicit rendering anomalies. Consequently, compared to languages with stricter syntax constraints, HTML tasks exhibit a statistically higher incidence of Code-related errors. For instance, as shown in Fig.~\ref{fig:error_case_2}, the omission of an explicit height definition for the parent container in the output code resulted in the height collapse of internal color blocks. While the image rendered successfully, this logical coding error caused the final visualization to diverge significantly from the ground truth.

Simultaneously, the dominance of Entity-related errors in LaTeX and Python underscores the challenges models face in defining entities when utilizing complex macros and functional tools. As demonstrated in Fig.~\ref{fig:error_case_3}, while the model successfully generated the majority of the content, it omitted specific components, such as the resistor present in the original diagram.

In conclusion, future advancements in MLLMs must prioritize enhancing fine-grained spatial geometric reasoning capabilities to mitigate layout deviations, while simultaneously improving precise control over code generation.

\subsection{Code-related Error}
To facilitate a more precise analysis of the execution failures encountered during our evaluation, we conduct a comprehensive diagnostic study on cases from the Omni-I2C benchmark where models failed to produce valid renderings. For each programming language, we developed a specialized taxonomy to categorize the root causes of these errors. By decoupling surface-level failures into granular dimensions, we expose the specific structural and logic-synthesis bottlenecks of current LMMs.

For Python-based visualizations, we categorize exceptions into four hierarchical levels. At the structural level, we identify Syntax $\&$ Parsing Failures and scope issues like Name Resolution $\&$ Import Failure. Interaction with library interfaces is monitored through Attribute Access Violations and Argument Specification Errors. Within the data logic itself, we distinguish between Type Semantics Violations, Value Domain Violations, and the Shape $\&$ Dimensionality Mismatches prevalent in tensor operations. Finally, execution context issues are captured by Resource $\&$ I/O Failures, Backend $\&$ Environment Limitations, and miscellaneous Undefined / Other Runtime Errors.

In the domain of SVG vector graphics, our taxonomy addresses the dual nature of the format as both a structured document and a graphical description. Failures in the XML layer are defined by Well-formedness Violations and Entity $\&$ Encoding Errors. Semantic issues within the graphic description include Geometric $\&$ Viewport Mismatches, where elements fall outside the canvas, and Attribute $\&$ Value Invalidity, such as malformed path strings. We also account for rendering constraints including Renderer Capability Limits, Typography $\&$ Asset Failures, and other Undefined/Other SVG Errors. This systematic attribution allows us to isolate whether a ``blank output'' stems from a parsing collapse or a subtle geometric miscalculation.

The compilation of LaTeX graphics involves a sensitive interplay between macro expansion and coordinate geometry, leading to eight diagnostic categories. Syntactic failures are identified as Token $\&$ Delimiter Errors or Command $\&$ Macro Definition issues. Environmental conflicts often arise from Dependency $\&$ Package omissions or Engine-Specific Conflicts between compilers like pdfLaTeX and XeLaTeX. Regarding drawing logic, we track Coordinate $\&$ Unit Violations and Visual/Logical Nullity—cases where valid code produces no visual output. Toolchain-related failures are captured through Externalization $\&$ Conversion errors and Miscellaneous Runtime exceptions.

In contrast, for HTML and SMILES, we employ a flattened error attribution logic due to their distinct execution characteristics. For HTML, the high fault tolerance of web browsers often results in ``silent failures'' (e.g., blank pages or invisible elements) that do not trigger explicit exceptions, making granular attribution less reliable. Similarly, failures in SMILES are predominantly monolithic, centered almost exclusively on Invalid SMILES Syntax. For these domains, we primarily report the binary success of the rendering or parsing process rather than a detailed sub-categorization, ensuring that our qualitative analysis remains grounded in observable execution data.

Ultimately, these findings underscore a broader hierarchical trend in model maturity. For more accessible formats like SVG and Python, the challenge for advanced models has shifted from basic syntactic integrity to complex spatial reasoning and structural alignment. However, for languages with high-entry barriers like LaTeX, even the most capable models remain susceptible to fundamental parsing collapses. This taxonomy-based analysis confirms that while LMMs are approaching syntactic competence, mastering the underlying logic and environmental constraints of executable code remains a formidable barrier across all scales and families.

\section{Case Study}
\label{case_study}
\newcommand{\modelA}{\textbf{Gemini 3 Pro}}
\newcommand{\modelB}{\textbf{Qwen3-VL-235B-A22B-Instruct}}
We conduct a qualitative analysis comparing 13 models on our benchmark. We select \modelA{} and \modelB{} for a detailed comparative case study due to their representative performance characteristics. 

\begin{figure*}[t]
    \centering
    
    \includegraphics[width=1.0\linewidth]{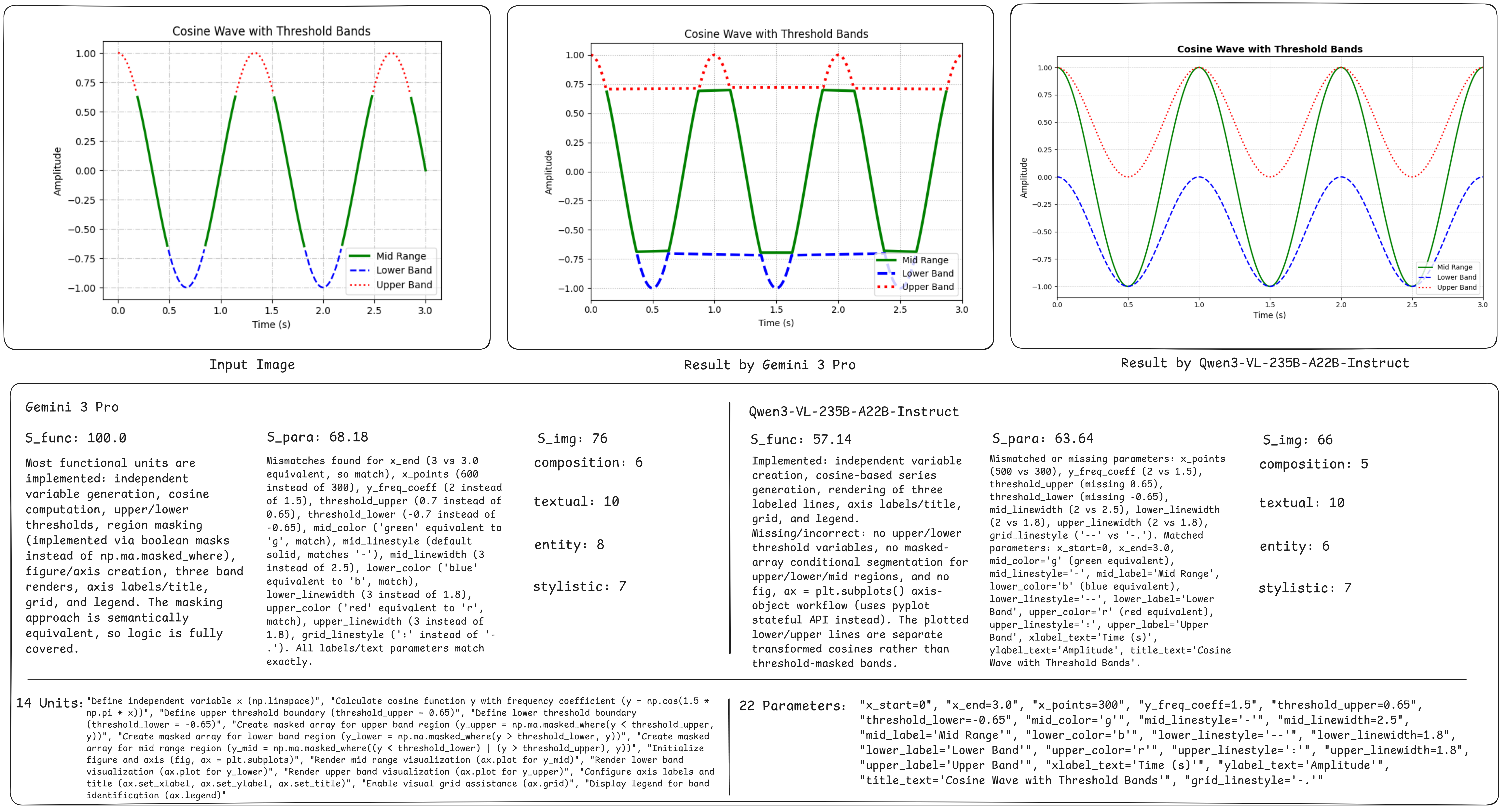}
    \caption{Case of Python}
    \label{case_study_py}
\end{figure*}

\paragraph{Case study in Python} We examine a representative Python code generation case in Fig.~\ref{case_study_py} to detail the evaluation scores. In terms of $S_{img}$ (Visual Fidelity), Gemini 3 Pro achieves a higher overall score than Qwen3-VL-235B-A22B-Instruct (76 vs. 66), specifically demonstrating better alignment in composition and entity rendering. For the $S_{func}$ metric, Gemini 3 Pro attains a full score of 100.0 by correctly implementing boolean masking for conditional threshold segmentation, whereas Qwen3-VL-235B-A22B-Instruct scores 57.14 by incorrectly generating separate, independent curves. Additionally, $S_{para}$ reveals that while both models exhibit numerical mismatches in frequency estimation, Qwen3-VL-235B-A22B-Instruct yields lower parameter precision (63.64 vs. 68.18) due to the complete omission of critical threshold variables.

\begin{figure*}[t]
    \centering
    
    \includegraphics[width=1.0\linewidth]{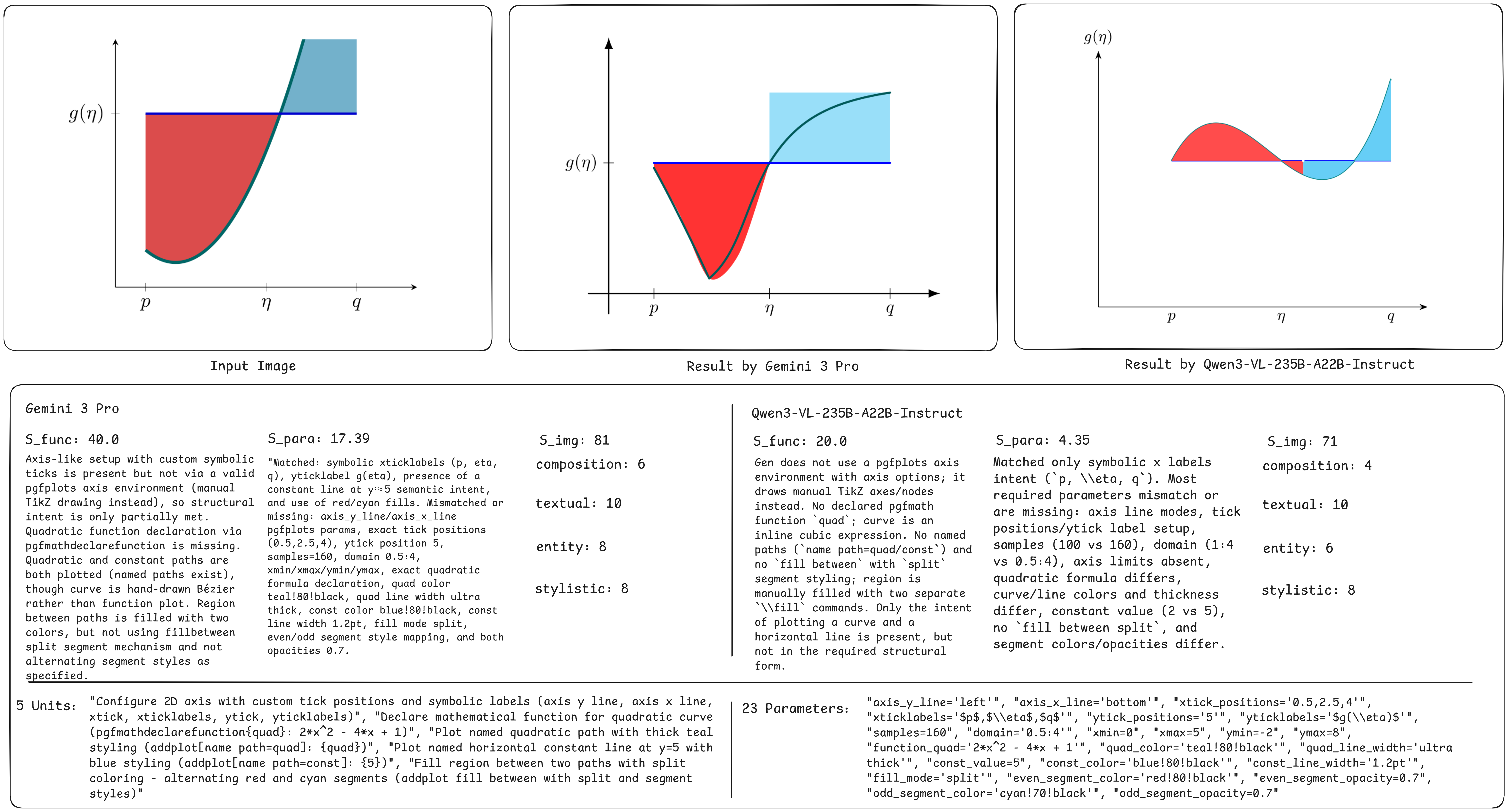}
    \caption{Case of LaTeX}
    \label{case_study_latex}
\end{figure*}

\paragraph{Case study in LaTeX} We examine a representative LaTeX code generation scenario shown in Fig.~\ref{case_study_latex} to analyze the reasoning behind these scores. In the LaTeX-based plotting task, Gemini 3 Pro achieves an $S_{img}$ score of 81, surpassing Qwen3-VL-235B-A22B-Instruct (71) in terms of structural fidelity. The $S_{func}$ evaluation characterizes a shared limitation in semantic LaTeX implementation: both models fail to invoke professional frameworks such as PGFPlots or its associated fillbetween library, defaulting to primitive TikZ operations and manual coordinate systems. Specifically, Gemini 3 Pro employs Bezier curves integrated with region clipping for segmented filling, whereas Qwen3-VL-235B-A22B-Instruct resorts to hard-coded vertices, yielding functionally inconsistent trajectories. Moreover, $S_{para}$ metrics indicate that Qwen3-VL-235B-A22B-Instruct suffers from significant deviations in numerical parameters, including scaling and sampling density. 

\begin{figure*}[t]
    \centering
    
    \includegraphics[width=1.0\linewidth]{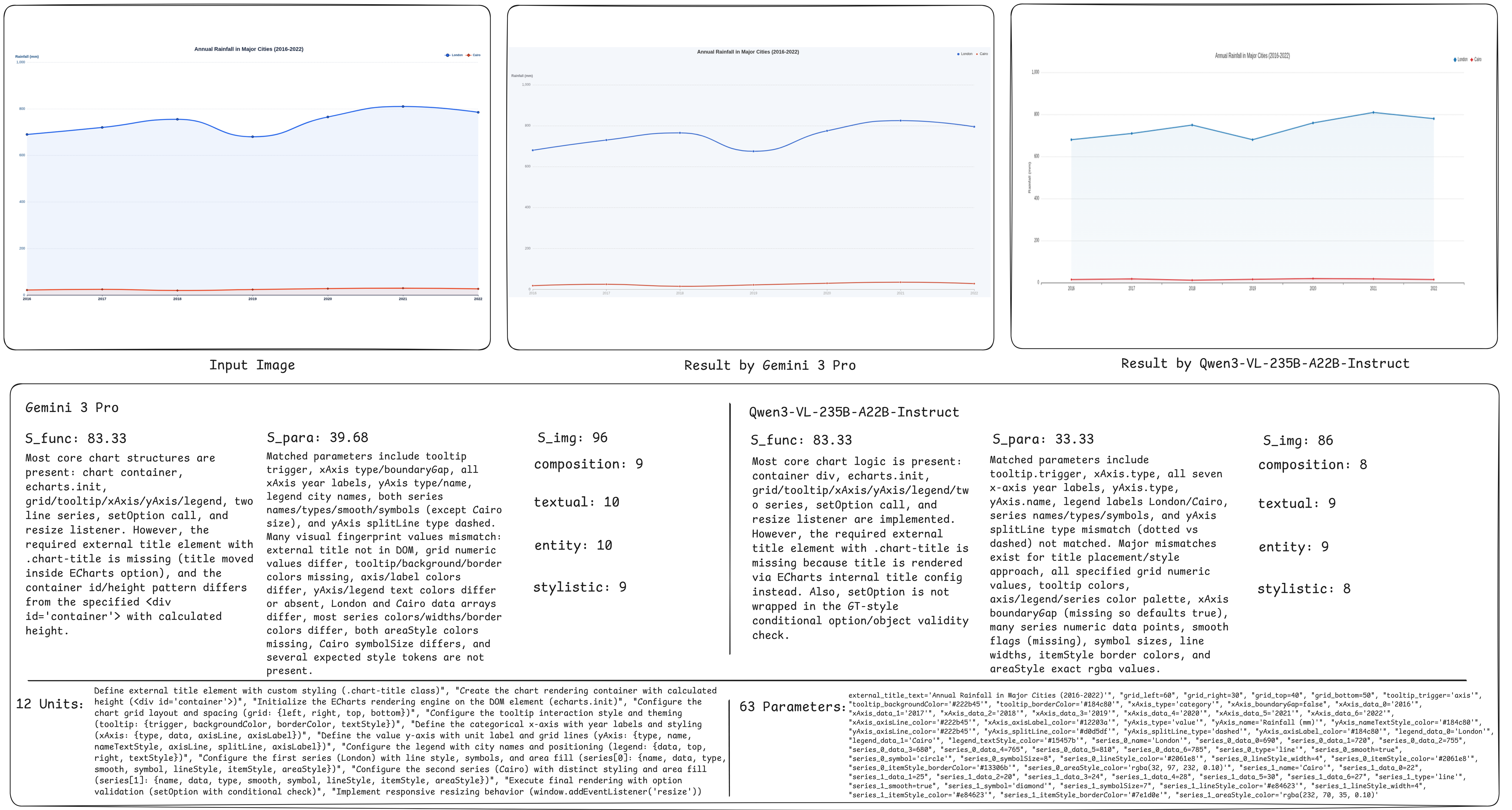}
    \caption{Case of HTML}
    \label{case_study_html}
\end{figure*}

\paragraph{Case study in HTML} We examine a representative HTML code generation scenario presented in Fig.~\ref{case_study_html} to report the evaluation metrics. In terms of $S_{img}$, Gemini 3 Pro achieves a higher overall score than Qwen3-VL-235B-A22B-Instruct (96 vs. 86), consistently demonstrating closer alignment across all fine-grained sub-metrics. For the $S_{func}$ metric, both models achieve an identical score of 83.33. The evaluation indicates that both models successfully implement the core chart-rendering components and basic configurations. However, minor functional deviations exist: Gemini 3 Pro utilizes a differing height calculation pattern, whereas Qwen3-VL-235B-A22B-Instruct omits the target conditional validity check prior to the \texttt{setOption} execution. Furthermore, $S_{para}$ reveals that Gemini 3 Pro yields higher parameter precision than Qwen3-VL-235B-A22B-Instruct (39.68 vs. 33.33). While both models correctly extract the tooltip trigger, axis types, textual labels, and core series definitions, Gemini 3 Pro successfully captures the \texttt{smooth} series property and the dashed \texttt{yAxis} split line, which Qwen3-VL-235B-A22B-Instruct misinterprets as dotted or omits.

\begin{figure*}[t]
    \centering
    
    \includegraphics[width=1.0\linewidth]{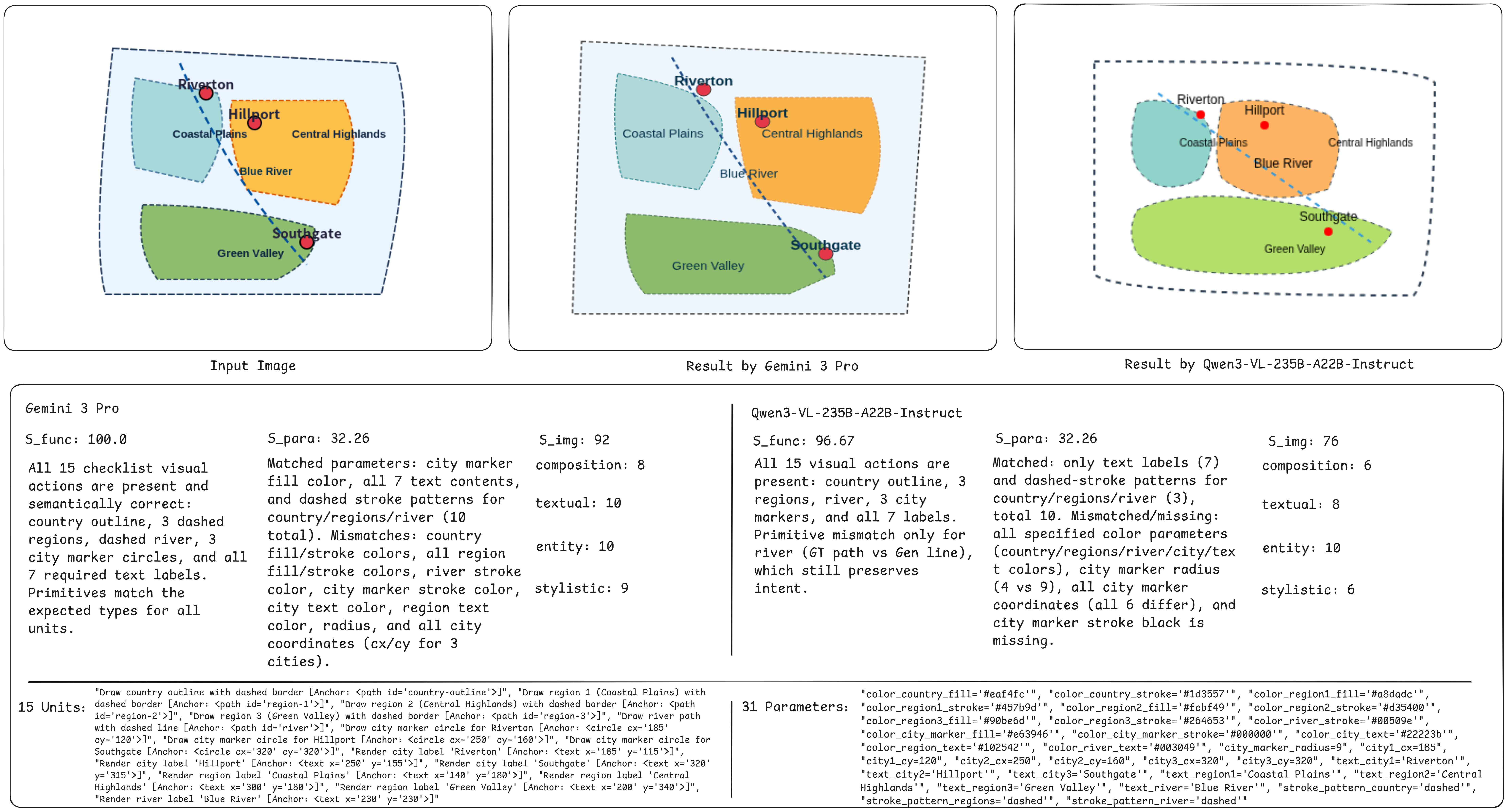}
    \caption{Case of SVG}
    \label{case_study_svg}
\end{figure*}

\paragraph{Case study in SVG}We examine a representative SVG code generation scenario presented in Fig.~\ref{case_study_svg} to report the evaluation metrics. In terms of $S_{img}$, Gemini 3 Pro achieves a higher overall score than Qwen3-VL-235B-A22B-Instruct (92 vs. 76), specifically demonstrating closer alignment in composition (8 vs. 6), textual elements (10 vs. 8), and stylistic rendering (9 vs. 6). For the $S_{func}$ metric, Gemini 3 Pro attains a full score of 100.0, successfully executing all 15 required visual actions with correct primitive types. Qwen3-VL-235B-A22B-Instruct achieves a comparable score of 96.67, rendering all required visual elements but exhibiting a minor primitive mismatch by utilizing a \texttt{line} element instead of a \texttt{path} for the river trajectory. Furthermore, $S_{para}$ reveals identical parameter precision scores for both models (32.26 vs. 32.26). The evaluation shows that both models correctly extract the 7 textual label contents and apply the required dashed stroke patterns for the boundaries and river. However, both systematically fail to reproduce exact spatial and stylistic parameters, exhibiting consistent mismatches in exact fill and stroke colors, city marker radii, and specific spatial coordinates for the city elements.

\begin{figure*}[t]
    \centering
    \includegraphics[width=1.0\linewidth]{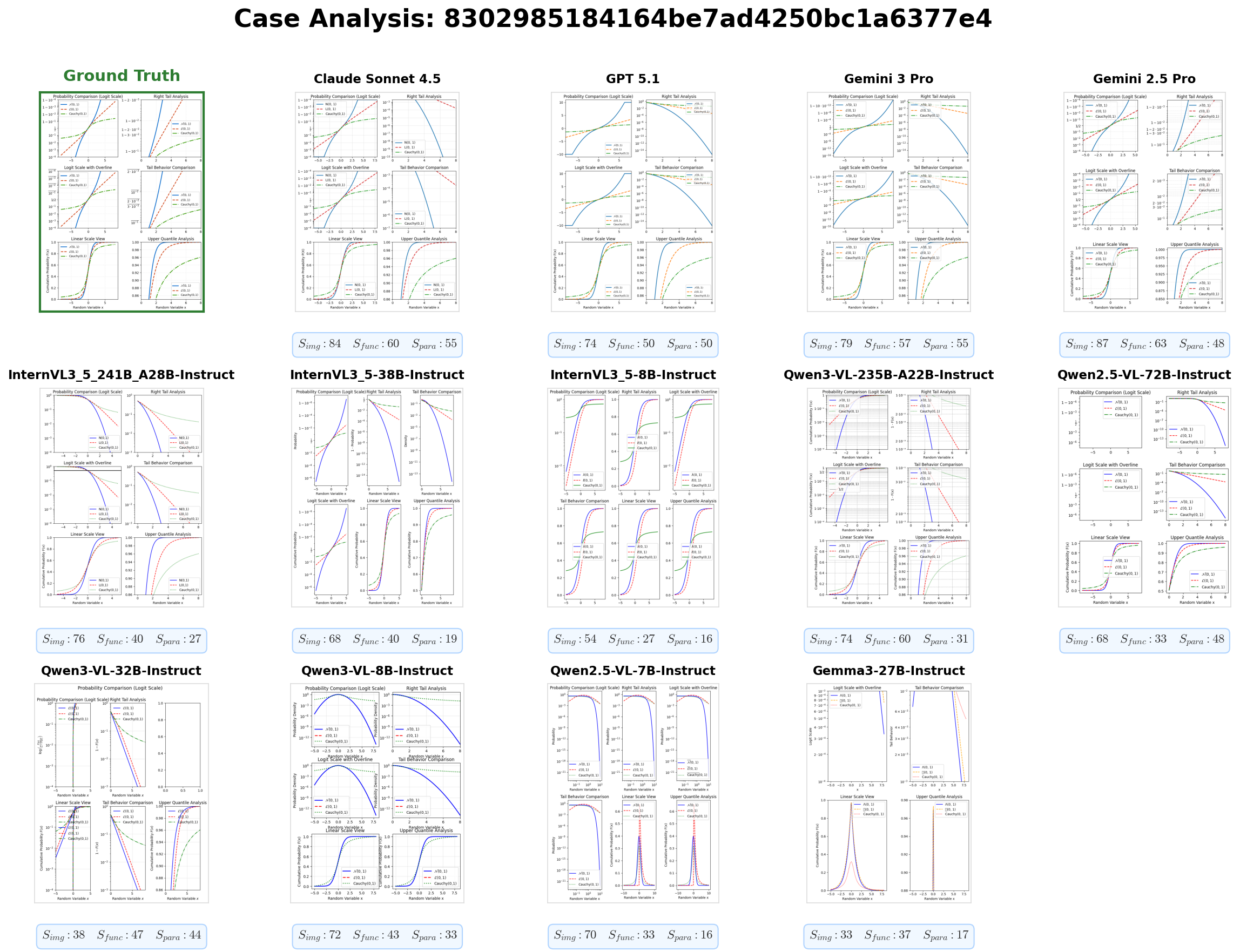}
    \caption{Case 1 of Render samples for all models benchmarked in our Table~\ref{tab:reorganized_breakdown_final_updated}.}
    \label{compare_13_1}
\end{figure*}

\begin{figure*}[t]
    \centering
    \includegraphics[width=1.0\linewidth]{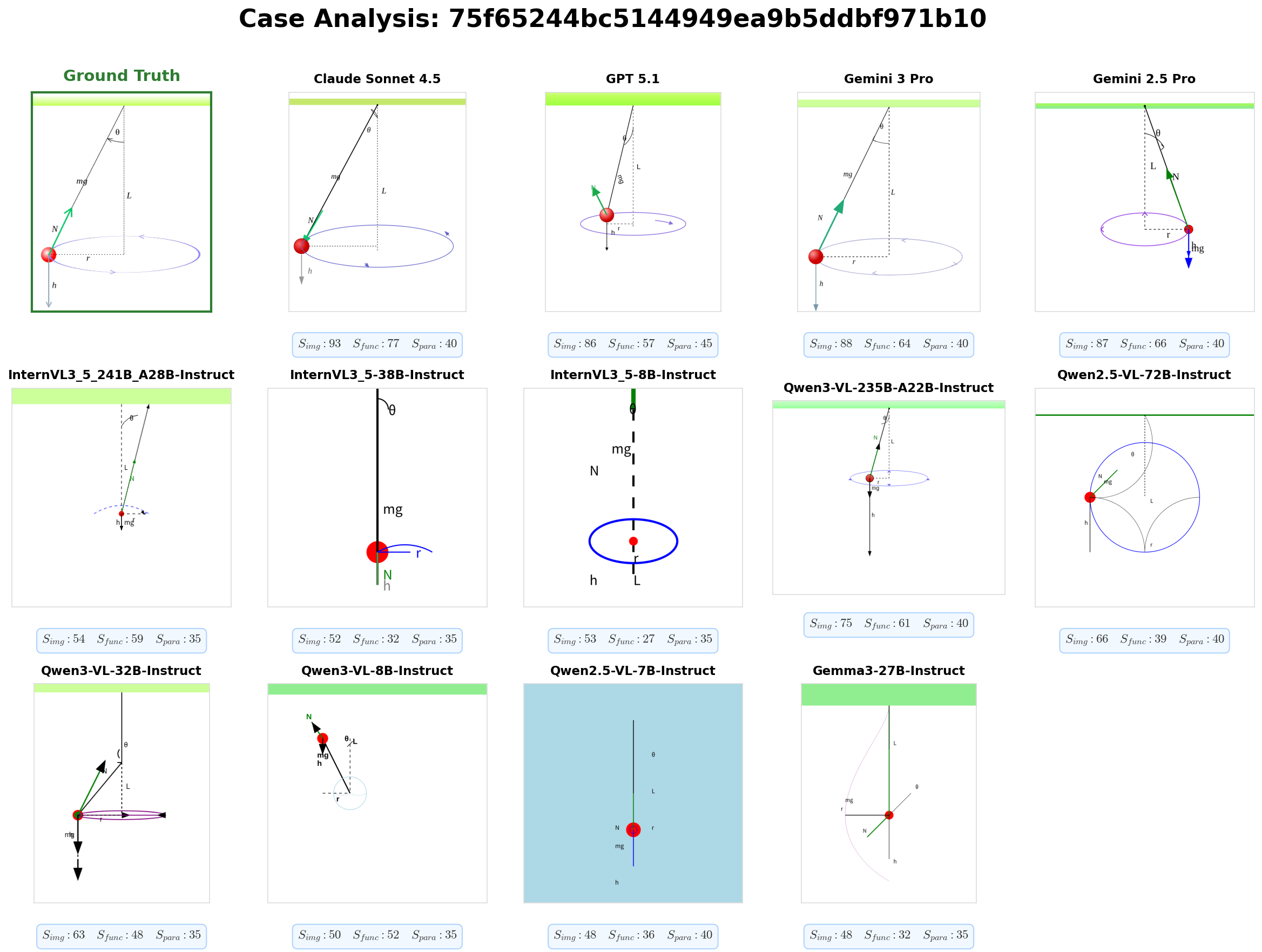}
    \caption{Case 2 of Render samples for all models benchmarked in our Table~\ref{tab:reorganized_breakdown_final_updated}.}
    \label{compare_13_2}
\end{figure*}

Furthermore, we provide qualitative rendering samples for all models benchmarked in our primary performance table. As illustrated in Fig.~\ref{compare_13_1}, Fig.~\ref{compare_13_2}, these examples offer a direct comparison of model outputs on specific data instances.

\section{Judge Alignment and Metric Validation}
\label{sec:appendix_judge}
This section details the methodologies and rationales used to ensure our automated evaluation tracks—both symbolic and visual—align with objective truth and human judgment.

\subsection{Symbolic Track: Code-level Reliability}
For the symbolic track, we evaluate the reliability of our code judges (Claude Sonnet 4.5, Gemini 2.5 Pro, and GPT 5.3 codex) by measuring their alignment with the Majority Vote (MV) ranking. As shown in Table~\ref{tab:judge_code}, we employ a suite of metametrics to quantify both individual judge accuracy and group consensus. Kendall’s $\tau$ and Spearman’s $\rho$: These are non-parametric rank correlation coefficients. $\tau$ measures the proportion of ``concordant pairs'' (pairs of items ranked in the same order by both the judge and MV), making it highly sensitive to even minor swaps in ranking. $\rho$ measures the strength of the monotonic relationship between ranks. MV align: Defined as $\frac{1}{2}(\tau+\rho)$, this composite metric serves as our primary indicator of alignment. We adopt this average to balance the strict ordinal consistency of $\tau$ with the rank-intensity correlation of $\rho$. As indicated in Table~\ref{tab:judge_code}, GPT 5.3 codex achieves the highest MV align ($0.8517$), justifying its selection as the official symbolic evaluator. Closest to MV (count/\%): This represents the frequency with which a specific judge’s ranking is the most similar to the majority consensus on an instance-by-instance basis. This metric exposes judge ``outliers'' and confirms that GPT 5.3 codex is the most consistently representative of the collective ``wisdom of the crowd'' (36.58\% frequency). Kendall’s $W$ (Coefficient of Concordance): This measures the agreement among all judges simultaneously. A value of $0.7618$ indicates a strong consensus across the evaluated frontier models, suggesting that ``quality'' in Image-to-Code tasks is an objective property that these models can identify consistently.

\subsection{Perceptual Track: Image-level Alignment}
To validate the $S_{\text{img}}$ metric, we compare the LMM judge outputs against a ``ground truth'' established by 10 human annotators. The results in Table~\ref{tab:image_level_final} demonstrate the superiority of the Omni-I2C prompting strategy. Ordinal Alignment ($\tau$ and $\rho$): Since $S_{\text{img}}$ is intended to act as a proxy for ``at-a-glance'' human preference, it is critical that the model-generated scores result in the same ranking order as human consensus. Our results show that the GPT 4.1 + Omni-I2C configuration reaches a peak Kendall’s $\tau$ of $0.8304$, significantly outperforming deduction-based alternatives like Chart2Code. Root Mean Square Error (RMSE): Unlike the symbolic track, which is purely rank-based, $S_{\text{img}}$ is a scalar value. We use RMSE to quantify the absolute deviation between the normalized LMM scores and human-assigned preference scores. A lower RMSE ($0.4284$ for Omni-I2C) indicates that the judge is well-calibrated—meaning its ``score magnitude'' matches human perception of quality, rather than just the ``relative order.'' Rationale for Metric Selection: Robustness to Diverse Modalities: By using rank correlation ($\tau$, $\rho$), we ensure the metric remains valid across different image types (e.g., SVG vs. LaTeX) where absolute visual differences might vary in scale. Sensitivity to Structural Failures: The inclusion of RMSE ensures that when a model produces a catastrophic layout failure, the judge penalizes it with a score magnitude that reflects the severity perceived by a human, rather than just ranking it ``last''. The consistent performance of our multi-dimensional prompt across different judges (Qwen-VL-Max and Qwen3-VL-Plus) confirms that a structured rubric focusing on Style, Layout, Elements, and Text is the most reliable way to stabilize perceptual evaluation in LMMs.

To further investigate the reliability of the structured rubric across different visual aspects, we conducted a fine-grained alignment experiment targeting the four specific sub-dimensions: \textit{Composition}, \textit{Text}, \textit{Entity}, and \textit{Style}. 
We recruit 16 professional annotators, each assigned to evaluate and rank 90 items across the four specific dimensions. We then calculate the correlation between the LMM-based scores and human rankings. This analysis is based on the samples generated by Gemini 2.5 Pro, Qwen3-VL-235B-A22B-Instruct, and InternVL3.5-8B-Instruct.

We then calculated the correlation between the LMM-based scores and human rankings using Kendall's $\tau$, Spearman's $\rho$, and RMSE. As summarized in Table~\ref{tab:dimension_alignment}, the LMM judge demonstrates a remarkably strong correlation with human judgment in Composition ($\tau=0.7837$) and Style ($\tau=0.7367$). These results suggest that for macroscopic visual features—such as global spatial arrangements and overall aesthetic coherence—the LMM acts as a highly reliable judge that closely aligns with human visual perception. 

\begin{table}[ht]
\centering
\footnotesize 
\setlength{\tabcolsep}{3.5pt} 
\caption{Human alignment analysis across the four sub-dimensions of $S_{\text{img}}$. Results are based on evaluations by 16 professional annotators.}
\label{tab:dimension_alignment}
\begin{tabular}{lccc}
\toprule
\textbf{Dimension} & \makecell[c]{\textbf{Kendall's} \\ \textbf{$\tau \uparrow$}} & \makecell[c]{\textbf{Spearman's} \\ \textbf{$\rho \uparrow$}} & \makecell[c]{\textbf{RMSE} \\ \textbf{$\downarrow$}} \\ \midrule
Composition & 0.7837 & 0.8222 & 0.4110 \\
Text        & 0.6889 & 0.7225 & 0.5267 \\
Entity      & 0.6531 & 0.6986 & 0.5167 \\
Style       & 0.7367 & 0.7791 & 0.4491 \\ \bottomrule
\end{tabular}
\end{table}

\section{Evaluation Metrics for Molecular Structures}
\label{appendix:smiles_metric}
To ensure a robust evaluation of chemical structures, we convert SMILES strings into molecular objects using the RDKit library. We specifically employ \textit{Morgan Fingerprints} with a radius of 2 and a bit-vector size of 2048. The similarity between the ground-truth ($G$) and predicted ($P$) molecules is then quantified via the Tanimoto coefficient:
\begin{equation}
\text{Tanimoto}(G, P) = \frac{\mathbf{v}_G \cdot \mathbf{v}_P}{\|\mathbf{v}_G\|^2 + \|\mathbf{v}_P\|^2 - \mathbf{v}_G \cdot \mathbf{v}_P}
\end{equation}
where $\mathbf{v}_G$ and $\mathbf{v}_P$ represent the fingerprint vectors.

\begin{figure*}[htbp]
    \centering
    \begin{subfigure}[b]{0.32\textwidth}
        \centering
        \includegraphics[width=\textwidth]{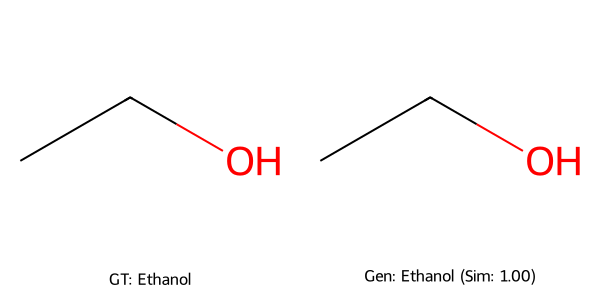}
        \caption{Case 1: Semantic Equivalence (Ethanol)}
        \label{fig:smiles_equiv}
    \end{subfigure}
    \hfill
    \begin{subfigure}[b]{0.32\textwidth}
        \centering
        \includegraphics[width=\textwidth]{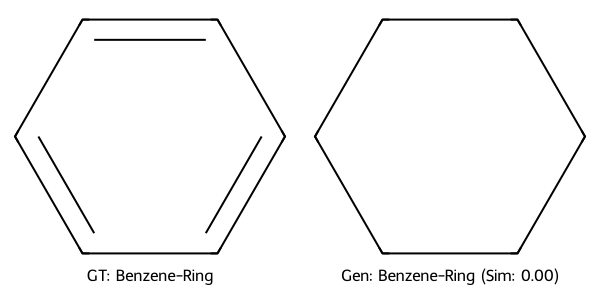}
        \caption{Case 2: Topological Error (Benzene)}
        \label{fig:smiles_topology}
    \end{subfigure}
    \hfill
    \begin{subfigure}[b]{0.32\textwidth}
        \centering
        \includegraphics[width=\textwidth]{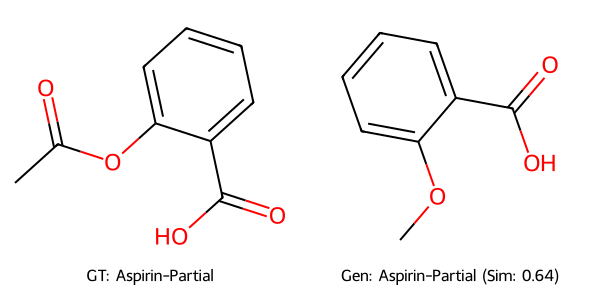}
        \caption{Case 3: Functional Group Fidelity}
        \label{fig:smiles_functional}
    \end{subfigure}
    
    \caption{Visual comparison of SMILES reconstruction cases. (a) Illustrates identical topology despite different string sequences. (b) Highlights critical perception errors in aromaticity captured by Tanimoto similarity. (c) Demonstrates the metric's sensitivity to localized functional group deletions.}
    \label{fig:smiles_all_cases}
\end{figure*}

To further elucidate the superiority of Morgan Fingerprint-based Tanimoto similarity over traditional string-based metrics, we present three representative cases encountered during our evaluation. These examples highlight the necessity of capturing chemical topology rather than mere syntactical sequences. The detailed examples are shown in Fig.~\ref{fig:smiles_all_cases}

\paragraph{Case 1: Semantic Equivalence under Canonicalization Variance} 
The inherent non-uniqueness of SMILES notation often leads to multiple valid strings for a single molecular structure. As shown in our first example (Ethanol, GT: \texttt{CCO}, Gen: \texttt{OCC}), string-based metrics like BLEU or Edit Distance would penalize the model for reversing the atom indexing order. In contrast, our metric yields a perfect Tanimoto score of $1.00$, as both strings map to the identical molecular fingerprint. This confirms the metric's ability to recognize functional identity regardless of the generation sequence.

\paragraph{Case 2: Topological Sensitivity to Atomic Hybridization} 
Small character-level differences in SMILES strings can signify profound changes in chemical properties that are easily overlooked by superficial text metrics. In the second case, we compare a benzene ring (\texttt{c1ccccc1}) with its saturated counterpart, cyclohexane (\texttt{C1CCCCC1}). While the text-level structure appears nearly identical (differing only by character casing), the Tanimoto similarity based on Morgan Fingerprints drops to \textbf{0.00}. This total divergence correctly reflects the fundamental loss of aromaticity and the shift in carbon hybridization from $sp^2$ to $sp^3$—a critical structural shift that renders the reconstructed molecule chemically distinct from the ground truth. String-based metrics, such as Edit Distance or standard token matching, often fail to penalize these ``semantic collapses'' heavily enough, potentially masking the severity of the prediction error.

\paragraph{Case 3: Fidelity of Complex Functional Groups} 
In more complex reconstructions, such as Aspirin (GT: \texttt{CC(=O)Oc1ccccc1C(=O)O}), the model might correctly generate the scaffold but fail on localized functional groups. When the acetoxy group is incorrectly reduced to a methoxy group (Gen: \texttt{COc1ccccc1C(=O)O}), the Tanimoto metric provides a nuanced penalty (Score: $0.58$), reflecting the partial structural loss. This demonstrates that the metric serves as a robust proxy for human expert judgment, as it effectively quantifies the impact of topological errors on the overall molecular integrity.

\clearpage

\clearpage

\clearpage

\clearpage
\section{Datasheet for Omni-I2C}

\subsection{Motivation}

\noindent \textbf{1. For what purpose was the dataset created? Was there a specific task in mind? Was there a specific gap that needed to be filled? Please provide a description.}

\textbf{A1:} Omni-I2C is designed to evaluate the capability of Large Multimodal Models (LMMs) in converting complex, structured digital graphics into executable code. Existing benchmarks are often restricted in image variety and programming languages, failing to capture the structural complexity found in authentic, user-sourced applications. This rigorous benchmark requires a synergy between high-fidelity visual perception and precise generative expression to achieve ``pixel-to-program'' alignment.

\noindent \textbf{2. Who created this dataset (e.g., which team, research group) and on behalf of which entity (e.g., company, institution, organization)?}

\textbf{A2:} This dataset is created by the authors of this paper.

\noindent \textbf{3. Who funded the creation of the dataset? If there is an associated grant, please provide the name of the grantor and the grant name and number.}
  
\textbf{A3:} N/A.

\subsection{Composition}

\noindent \textbf{1. What do the instances that comprise the dataset represent? Please provide a description.}

\textbf{A1:} Omni-I2C consists of 1,080 high-quality Image2Code instances. Each instance comprises an input digital graphic (image) and its corresponding ground-truth implementation in executable code, spanning technical domains such as scientific visualizations, molecular structures, and complex symbolic notations.

\noindent \textbf{2. How many instances are there in total (of each type, if appropriate)?}

\textbf{A2:} There are 1,080 evaluation instances in total, distributed across five programming and markup languages: Python (357), LaTeX (265), HTML (166), SVG (192), and SMILES (100).

\noindent \textbf{3. Does the dataset contain all possible instances or is it a sample? If the dataset is a sample, then what is the larger set?}

\textbf{A3:} It is a curated sample of real-world items. To ensure practical utility, we harvested image-code pairs from official documentation, specialized tutorials, and community galleries, representing a wide spectrum of digital content.

\noindent \textbf{4. What data does each instance consist of? “Raw” data or features?}

\textbf{A4:} Each instance consists of the ``raw'' input image, the corresponding textual instruction, and the ground-truth executable code.

\noindent \textbf{5. Is there a label or target associated with each instance? If so, please provide a description.}

\textbf{A5:} Yes. Each target is the executable code that accurately reconstructs the visual input. The fidelity of the reconstructed image is evaluated across semantic and structural dimensions.

\noindent \textbf{6. Is any information missing from individual instances?}

\textbf{A6:} No.

\noindent \textbf{7. Are relationships between individual instances made explicit?}

\textbf{A7:} Yes. Instances are explicitly categorized by their programming language, academic discipline (8 subjects), and figure archetype (45 types).

\noindent \textbf{8. Are there recommended data splits (e.g., training, development/validation, testing)?}

\textbf{A8:} Omni-I2C is primarily intended as a benchmark for evaluation.

\noindent \textbf{9. Are there any errors, sources of noise, or redundancies in the dataset?}

\textbf{A9:} No. All instances were verified through an isolated sandbox environment to ensure the code compiles and produces the correct visual output.

\noindent \textbf{10. Is the dataset self-contained, or does it link to or otherwise rely on external resources?}

\textbf{A10:} The dataset is self-contained. While raw data was collected from public sources, the final curated benchmark includes all necessary code and images.

\noindent \textbf{11. Does the dataset contain data that might be considered confidential?}

\textbf{A11:} No. A manual audit was performed to prune any Personally Identifiable Information (PII).

\noindent \textbf{12. Does the dataset contain data that might be offensive?}

\textbf{A12:} No.
\subsection{Collection Process}

\noindent \textbf{1. How was the data associated with each instance acquired?}

\textbf{A1:} Data was acquired through a systematic three-stage pipeline: collection, filtering, and decontamination. We targeted official documentation and specialized galleries (e.g., Matplotlib Gallery, TikZ.net, TeXample.net) to capture genuine practical utility.

\noindent \textbf{2. What mechanisms or procedures were used to collect the data?}

\textbf{A2:} We utilized a taxonomy-driven approach, pre-defining a comprehensive schema to guide the search and ensure broad coverage across subjects and languages.

\noindent \textbf{3. If the dataset is a sample from a larger set, what was the sampling strategy?}

\textbf{A3:} Purposeful sampling was used to address gaps in existing benchmarks, with 5 languages and 45 distinct figure types.

\noindent \textbf{4. Who was involved in the data collection process?}

\textbf{A4:} Beyond the primary authors, nine internal annotators from the same research institution were engaged for data collection. These individuals were compensated at a rate of eight times the local minimum hourly wage. To prevent interference with their regular schedules, the tasks were conducted on a flexible, part-time basis without fixed daily hour requirements. The average time commitment per annotator was approximately two days.

\noindent \textbf{5. Over what timeframe was the data collected?}

\textbf{A5}: The collection, manual auditing, and decontamination process took approximately one month.

\subsection{Preprocessing/cleaning/labeling}

\noindent \textbf{1. Was any preprocessing/cleaning/labeling of the data done?}

\textbf{A1:} Yes. Candidates were executed in isolated sandboxes to discard instances with compilation errors. Manual auditing was used to prune social biases and PII. Finally, we used an LLM to refine the categorization of the remaining data.

\noindent \textbf{2. Was a data decontamination strategy employed?}

\textbf{A2:} Yes. We employed an LLM to reformulate code styles and aesthetic parameters (fonts, layouts). For images, we performed ``content erasure'' and generated new synthetic strings to prevent models from relying on linguistic priors or memorization.

\noindent \textbf{3. Is the software used to preprocess/clean/label the instances available?}

\textbf{A3:} The evaluation pipeline and sandbox configurations will be provided upon publication.

\subsection{Uses}

\noindent \textbf{1. Has the dataset been used for any tasks already? If so, please provide a description.}

\textbf{A1:} Yes. We benchmarked 13 state-of-the-art LMMs, including GPT-5.1, Claude Sonnet 4.5, and Gemini 3 Pro, evaluating their ability to reconstruct structural integrity in complex scenarios.

\noindent \textbf{2. Is there a repository that links to any or all papers or systems that use the dataset? If so, please provide a link or other access point.}

\textbf{A2:} Data and code are available at https://github.com/MiliLab/Omni-I2C.

\noindent \textbf{3. What (other) tasks could the dataset be used for?}

\textbf{A3:} Omni-I2C can be used for evaluating multimodal reasoning, precise coordinate modeling in SVG/HTML, and specialized chemical informatics knowledge via SMILES.

\noindent \textbf{4. Is there anything about the composition of the dataset or the way it was collected that might impact future uses? Is there anything a future user could do to mitigate these undesirable harms?}

\textbf{A4:} No. The dataset was designed with counter-intuitive or ``nonsense'' content swaps to force models to rely on visual perception rather than linguistic priors, which actually enhances its robustness as a perceptual benchmark.

\noindent \textbf{5. Are there tasks for which the dataset should not be used? If so, please provide a description.}

\textbf{A5:} N/A.

\subsection{Distribution}

\noindent \textbf{1. Will the dataset be distributed to third parties outside of the entity? If so, please provide a description.}

\textbf{A1:} Yes, the dataset will be publicly available for the research community.

\noindent \textbf{2. How will the dataset will be distributed? Does the dataset have a digital object identifier (DOI)?}

\textbf{A2:} Omni-I2C will be distributed via the official project website and GitHub.

\noindent \textbf{3. When will the dataset be distributed?}

\textbf{A3:} The dataset will be distributed once the paper is accepted after peer review.

\noindent \textbf{4. Will the dataset be distributed under a copyright or other license?}

\textbf{A4:} It will be distributed under the Creative Commons Attribution-NonCommercial-ShareAlike 4.0 International (CC BY-NC-SA 4.0) License.

\noindent \textbf{5. Have any third parties imposed IP-based or other restrictions on the data associated with the instances?}

\textbf{A5:} No.

\noindent \textbf{6. Do any export controls or other regulatory restrictions apply to the dataset or to individual instances?}

\textbf{A6:} No.

\subsection{Maintenance}

\noindent \textbf{1. Who will be supporting/hosting/maintaining the dataset?}

\textbf{A1:} The authors.

\noindent \textbf{2. How can the owner/curator/manager of the dataset be contacted (e.g., email address)?}

\textbf{A2:} Email addresses will be provided on the project homepage post-publication.

\noindent \textbf{3. Is there an erratum? If so, please provide a link or other access point.}

\textbf{A3:} Any errata will be posted on the project GitHub repository.

\noindent \textbf{4. Will the dataset be updated? If so, please describe how often, by whom, and how updates will be communicated to users?}

\textbf{A4:} Yes. The authors will periodically update the benchmark with new instances and categories based on community feedback.

\noindent \textbf{5. Will older versions of the dataset continue to be supported/hosted/maintained? If so, please describe how.}

\textbf{A5:} Older versions will be archived on GitHub to ensure researchers can reproduce results from previous studies.

\noindent \textbf{6. If others want to extend/augment/build on/contribute to the dataset, is there a mechanism for them to do so?}

\textbf{A6:} N/A.

\end{document}